\documentclass[lettersize,journal]{IEEEtran}
\usepackage{amsmath,amsfonts}
\usepackage{algorithmic}
\usepackage{array}
\usepackage{textcomp}
\usepackage{stfloats}
\usepackage{url}
\usepackage{verbatim}
\usepackage{times}
\usepackage{epsfig}
\usepackage{graphicx}
\usepackage{amssymb}
\usepackage{booktabs}
\usepackage{makecell}
\usepackage{xcolor}
\usepackage{makecell}
\usepackage{multirow}
\usepackage{bbding}
\usepackage{pifont}
\usepackage{multicol}
\usepackage{tcolorbox}
\usepackage{fontawesome}
\usepackage{colortbl}
\usepackage{hhline}
\usepackage{float}
\usepackage{soul}
\usepackage{longtable}
\usepackage[linesnumbered,lined,ruled]{algorithm2e}
\usepackage[pagebackref,breaklinks,colorlinks]{hyperref}

\usepackage[misc]{ifsym}
\def\BibTeX{{\rm B\kern-.05em{\sc i\kern-.025em b}\kern-.08em
    T\kern-.1667em\lower.7ex\hbox{E}\kern-.125emX}}
\definecolor{myblue}{RGB}{222,235,247}

\begin{document}

\title{Unleashing the Power of Continual Learning on Non-Centralized Devices: A Survey}

\author{Yichen~Li,
        Haozhao~Wang,
        Wenchao~Xu, 
        Tianzhe~Xiao,
        Hong~Liu,
        Minzhu~Tu,
        Yuying~Wang, 
        Xin~Yang,
        Rui~Zhang,~\IEEEmembership{Senior Member,~IEEE},
        Shui~Yu,~\IEEEmembership{Fellow,~IEEE},
        Song~Guo,~\IEEEmembership{Fellow,~IEEE},
        Ruixuan~Li
\IEEEcompsocitemizethanks{
\IEEEcompsocthanksitem Yichen Li, Haozhao Wang, Tianzhe Xiao, Hong Liu, Rui Zhang, and Ruixuan Li are with the School of Computer Science and Technology, Huazhong University of Science and Technology, Wuhan, China. \protect
E-mail:\{ycli0204, hz\_wang, tz\_\_xiao, lh678\}@hust.edu.cn, rayteam@yeah.net, rxli@hust.edu.cn}
\IEEEcompsocitemizethanks{
\IEEEcompsocthanksitem Wenchao Xu is with the Department of Computing, The Hong Kong Polytechnic University, Hong Kong, China. \protect
E-mail: wenchao.xu@polyu.edu.hk}
\IEEEcompsocitemizethanks{
\IEEEcompsocthanksitem Minzhu Tu is with the School of Computer Science (National Pilot Software Engineering School), Beijing University of Posts and Telecommunications, Beijing, China. \protect
E-mail: cipher0.1104@gmail.com}
\IEEEcompsocitemizethanks{
\IEEEcompsocthanksitem Yuying Wang is with the School of Computer Science and Technology, Soochow University, Suzhou, China. \protect
E-mail: yywang23@stu.suda.edu.cn}
\IEEEcompsocitemizethanks{
\IEEEcompsocthanksitem Xin Yang is with the School of Computing and Artificial Intelligence, Southwestern University of Finance
and Economics, Chengdu, China. \protect
E-mail: yangxin@swufe.edu.cn}
\IEEEcompsocitemizethanks{
\IEEEcompsocthanksitem Shui Yu is with the School of Computer Science, University of Technology Sydney, Sydney, Australia. \protect
E-mail: Shui.Yu@uts.edu.au}
\IEEEcompsocitemizethanks{
\IEEEcompsocthanksitem Song Guo is with the Department of Computer Science and Engineering, The Hong Kong University of Science and Technology, Hong Kong, China. \protect
E-mail: songguo@cse.ust.hk}
}

\markboth{MANUSCRIPT SUBMITTED TO IEEE COMMUNICATIONS SURVEYS \& TUTORIALS
}%
{Shell \MakeLowercase{\textit{et al.}}: A Sample Article Using IEEEtran.cls for IEEE Journals}

\IEEEpubid{0000--0000/00\$00.00~\copyright~2021 IEEE}

\maketitle
\begin{abstract}
Non-Centralized Continual Learning (NCCL) has become an emerging paradigm for enabling distributed devices such as vehicles and servers to handle streaming data from a joint non-stationary environment. To achieve high reliability and scalability in deploying this paradigm in distributed systems, it is essential to {overcome} challenges stemming from both spatial and temporal dimensions, manifesting as distribution shifts, catastrophic forgetting, heterogeneity, and privacy issues. 
%
This survey focuses on a comprehensive examination of the development of the non-centralized continual learning algorithms and the real-world deployment across distributed devices. We begin with an introduction to the background and fundamentals of non-centralized learning and continual learning. Then, we review existing solutions from three levels to represent how existing techniques alleviate the catastrophic forgetting and distribution shift. Additionally, we delve into the various types of heterogeneity issues, security, and privacy attributes, as well as real-world applications across three prevalent scenarios. Furthermore, we establish a large-scale benchmark to revisit this problem and analyze the performance of the state-of-the-art NCCL approaches. Finally, we discuss the important challenges and future research directions in NCCL.
\end{abstract}


\section{Introduction}
\subsection{{Background}}
Artificial intelligence, especially deep learning, stands to benefit immensely from the vast amount of data expected in the coming years \cite{dl}. Many algorithms have proven effective across various industrial, scientific, and engineering applications. However, data generally originates from a massive number of devices and is also commonly stored in a distributed manner for a wide range of scenarios. 
For example, private photos are locally stored on mobile phones, and trajectory data is stored in vehicles.
In such cases, data collection in central entities, referred to as centralized training, is often infeasible or impractical due to limited communication resources, data privacy concerns, or country regulations. Currently, the Non-Centralized Learning (NCL) paradigm emerges as a positive response to the increasing need for distributed learning and other concerns in deep learning applications \cite{ncl1,ncl2}. In NCL, distributed devices process data locally and share knowledge without compromising communication overhead and privacy issues. Among the various NCL algorithms, Federated Learning (FL), introduced by \cite{fedavg} in 2017, has gained wide acknowledgment for its ability to enable entities (also referred to as participants, clients, or parties) to collaboratively train models without sharing sensitive training data. Following years of development, NCL has yielded impressive results, particularly in areas such as edge-cloud collaboration \cite{kang2020reliable,niyato_edgecloud,niyato_edgecloud2}, decentralized learning \cite{decentralized,liu2021consensus,sun2024byzantine}, federated learning \cite{fl,wang2023dafkd,wang2021losp}, edge computing \cite{edge1,edge2,zhong2022flee}, and beyond \cite{zhong2022flee}. {Meanwhile, it has been applied in various areas such as recommendation systems \cite{li2025personalized,li2025survey,zheng2023decentralized} and autonomous driving \cite{kang_drive1,kang_drive2,kang_drive3}.}
\IEEEpubidadjcol

\begin{figure}[t]
    \centering
    \includegraphics[width=\linewidth]{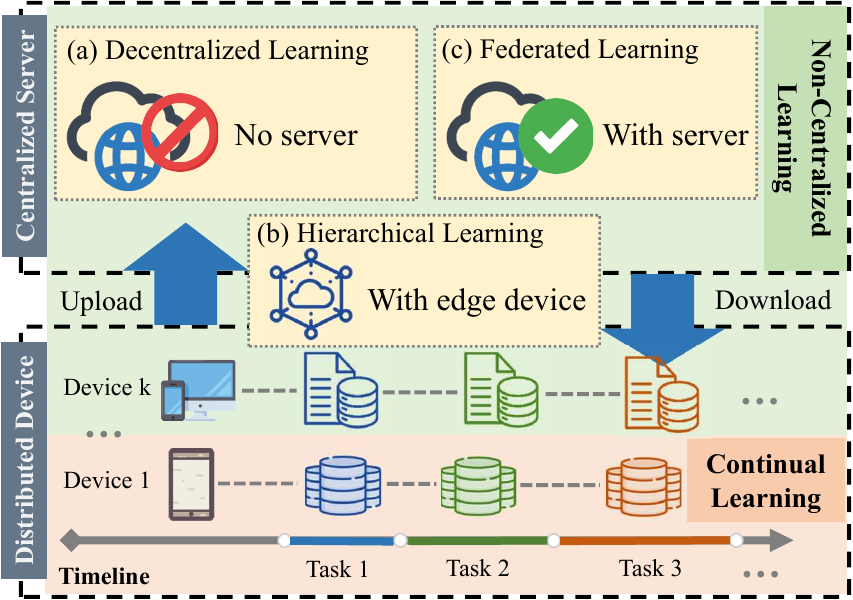}
    \caption{The overview of NCCL paradigms, including the distributed device layer and centralized server layer with the continual learning process and three different NCL learning paradigms. Each device will collect new data during the training process and collaboratively maintain the model on all streaming tasks without breaching data privacy.}
    \label{framework}
\end{figure}
\subsection{{Motivation}}
Despite its impressive advantages, the traditional NCL paradigm makes an unrealistic assumption: all local data on distributed devices should be known beforehand and will remain static and fixed, contradicting the dynamic nature of the real world. In a realistic NCL application, each client may continue collecting new streaming data and train new data on a model that has converged on previous tasks. The lack of dynamism in NCL settings further exacerbates the problem of Catastrophic Forgetting (CF), which is a major challenge in the centralized Continual Learning (CL) paradigm \cite{cf1}. As models adapt to newly collected data, they often struggle to retain the knowledge gained from previous tasks, leading to a significant degradation in model performance, especially on tasks learned earlier in the training process \cite{cf2,cf3}. This challenge is more severe in NCL settings, where the data in each device remains inaccessible to others, and devices often lack enough resources to implement algorithms designed for centralized settings. 
Considering the streaming data in distributed devices over time, the challenge of data distribution shift will deserve more attention. In traditional NCL tasks, each device often encounters heterogeneous data across clients participating in the non-centralized training process \cite{heterogeneity1,heterogeneity2}. Such inconsistencies are primarily due to diverse data sources, distinct data collection methods, and varying data quality. However, when local training data arrives in a streaming manner, each device should focus on the distribution shift between the current task and previous tasks, leading to the problem of knowledge fusion from different tasks and convergence of the local model.

\begin{table*}[t]
    \centering
        \caption{Summary of related works versus our survey. We identify relevant surveys across three categories and summarized them based on four aspects: paradigm, heterogeneity (in data, model, computation, and communication), threats, and application.}
    \renewcommand\arraystretch{1.3}
      \resizebox{\linewidth}{!}{
    \begin{tabular}{c | c c |c c| c c c c| c c| c| m{0.35\textwidth}}
    \toprule[1pt]
    \multirow{2}{*}{Category} & \multirow{2}{*}{Year} & \multirow{2}{*}{Ref.} & \multicolumn{2}{c|}{Paradigm} & \multicolumn{4}{c|}{Heterogeneity} & \multicolumn{2}{c|}{Threat} & \multirow{2}{*}{Appli.} & \multirow{2}{*}{Contributions} \\ \cline{4-11}
    & & & NCL & CL & Data & Model & Compu. & Commu. & Privacy & Security &  & \\ \hline
   \rowcolor{blue!8}
    & 2021 & \cite{survey1_ncl} & \textcolor{olive}{\large{\faCheckCircle}} & \textcolor{red}{\large{\faTimesCircle}} & \textcolor{olive}{\large{\faCheckCircle}} & \textcolor{red}{\large{\faTimesCircle}} & \textcolor{red}{\large{\faTimesCircle}} & \textcolor{red}{\large{\faTimesCircle}} & \textcolor{olive}{\large{\faCheckCircle}} & \textcolor{red}{\large{\faTimesCircle}} & \textcolor{olive}{\large{\faCheckCircle}}&  Highlight various FL architectures and algorithms, emphasizing the need for non-centralized approaches to address privacy and scalability issues. \\ \hhline{|~|------------}
    \rowcolor{blue!8}
    & 2020 & \cite{survey2_ncl} & \textcolor{olive}{\large{\faCheckCircle}} & \textcolor{red}{\large{\faTimesCircle}} & \textcolor{red}{\large{\faTimesCircle}} & \textcolor{red}{\large{\faTimesCircle}} & \textcolor{olive}{\large{\faCheckCircle}} & \textcolor{olive}{\large{\faCheckCircle}} & \textcolor{olive}{\large{\faCheckCircle}} & \textcolor{olive}{\large{\faCheckCircle}} & \textcolor{olive}{\large{\faCheckCircle}} & Survey FL in mobile edge networks with resource allocation and communication efficiency and identify research directions and challenges in applications. \\ \hhline{|~|------------}
    \rowcolor{blue!8}
    & 2021 & \cite{survey3_ncl} & \textcolor{olive}{\large{\faCheckCircle}} & \textcolor{red}{\large{\faTimesCircle}} & \textcolor{red}{\large{\faTimesCircle}} & \textcolor{red}{\large{\faTimesCircle}} & \textcolor{red}{\large{\faTimesCircle}} & \textcolor{olive}{\large{\faCheckCircle}} & \textcolor{olive}{\large{\faCheckCircle}} & \textcolor{olive}{\large{\faCheckCircle}} & \textcolor{olive}{\large{\faCheckCircle}} & Review FL for the IoT and discuss the unique challenges like heterogeneity and resources. \\ \hhline{|~|------------}
    \rowcolor{blue!8}
    & 2021 & \cite{survey4_ncl} & \textcolor{olive}{\large{\faCheckCircle}} & \textcolor{red}{\large{\faTimesCircle}} & \textcolor{olive}{\large{\faCheckCircle}} & \textcolor{red}{\large{\faTimesCircle}} & \textcolor{red}{\large{\faTimesCircle}} & \textcolor{red}{\large{\faTimesCircle}} & \textcolor{olive}{\large{\faCheckCircle}} & \textcolor{olive}{\large{\faCheckCircle}} & \textcolor{red}{\large{\faTimesCircle}} & Provide a taxonomy and comprehensive survey of privacy-preserving FL techniques and contrast different privacy-preserving mechanisms. \\ \hhline{|~|------------}
    \rowcolor{blue!8}
    \multirow{-5}{*}[9ex]{\makecell[c]{Non-Centralized\\ Learning}} & 2023 & \cite{survey5_ncl} & \textcolor{olive}{\large{\faCheckCircle}} & \textcolor{red}{\large{\faTimesCircle}} & \textcolor{olive}{\large{\faCheckCircle}} & \textcolor{red}{\large{\faTimesCircle}} & \textcolor{red}{\large{\faTimesCircle}} & \textcolor{olive}{\large{\faCheckCircle}} & \textcolor{olive}{\large{\faCheckCircle}} & \textcolor{olive}{\large{\faCheckCircle}} & \textcolor{olive}{\large{\faCheckCircle}} &  Introduce decentralized learning and emphasize the benefits of decentralized approaches in improving scalability, robustness, and fault tolerance.\\ 
    \hline \hline

    \rowcolor{green!12}
     & 2021 & \cite{survey1_cl} & \textcolor{red}{\large{\faTimesCircle}} & \textcolor{olive}{\large{\faCheckCircle}} & \textcolor{olive}{\large{\faCheckCircle}} & \textcolor{red}{\large{\faTimesCircle}} & \textcolor{red}{\large{\faTimesCircle}} & \textcolor{red}{\large{\faTimesCircle}} & \textcolor{red}{\large{\faTimesCircle}} & \textcolor{red}{\large{\faTimesCircle}} & \textcolor{red}{\large{\faTimesCircle}} &  Survey CL methods for classification tasks with various CL strategies, including regularization-based, rehearsal-based, and dynamic architectures. \\ \hhline{|~|------------}
    \rowcolor{green!12}
    & 2022 & \cite{survey2_cl} & \textcolor{red}{\large{\faTimesCircle}} & \textcolor{olive}{\large{\faCheckCircle}} & \textcolor{olive}{\large{\faCheckCircle}} & \textcolor{red}{\large{\faTimesCircle}} & \textcolor{red}{\large{\faTimesCircle}} & \textcolor{red}{\large{\faTimesCircle}} & \textcolor{red}{\large{\faTimesCircle}} & \textcolor{red}{\large{\faTimesCircle}} & \textcolor{red}{\large{\faTimesCircle}} & Classify incremental learning into three types: CIL, TIL, and DIL and discuss the unique challenges and research directions for each type of incremental learning.\\ \hhline{|~|------------}
    \rowcolor{green!12}
    \multirow{-3}{*}[4ex]{Continual Learning} & 2022 & \cite{survey3_cl} & \textcolor{red}{\large{\faTimesCircle}} & \textcolor{olive}{\large{\faCheckCircle}} & \textcolor{olive}{\large{\faCheckCircle}} & \textcolor{red}{\large{\faTimesCircle}} & \textcolor{red}{\large{\faTimesCircle}} & \textcolor{red}{\large{\faTimesCircle}} & \textcolor{red}{\large{\faTimesCircle}} & \textcolor{red}{\large{\faTimesCircle}} & \textcolor{red}{\large{\faTimesCircle}} &  Provide a survey and benchmark evaluation of CIL methods for image classification.\\ 
    
    \hline \hline
    \rowcolor{orange!15}
     & 2023 & \cite{survey1_nccl}& \textcolor{olive}{\large{\faCheckCircle}} & \textcolor{olive}{\large{\faCheckCircle}} & \textcolor{olive}{\large{\faCheckCircle}} & \textcolor{red}{\large{\faTimesCircle}} & \textcolor{red}{\large{\faTimesCircle}} & \textcolor{red}{\large{\faTimesCircle}} & \textcolor{red}{\large{\faTimesCircle}} & \textcolor{red}{\large{\faTimesCircle}} & \textcolor{red}{\large{\faTimesCircle}} & Surveys incremental transfer learning, combining P2P FL and DIL for multi-center collaboration and propose a novel framework for NCCL. \\ \hhline{|~|------------}
    \rowcolor{orange!15}
    & 2022 & \cite{survey2_nccl} & \textcolor{olive}{\large{\faCheckCircle}} & \textcolor{olive}{\large{\faCheckCircle}} & \textcolor{olive}{\large{\faCheckCircle}} & \textcolor{red}{\large{\faTimesCircle}} & \textcolor{red}{\large{\faTimesCircle}} & \textcolor{red}{\large{\faTimesCircle}} & \textcolor{red}{\large{\faTimesCircle}} & \textcolor{red}{\large{\faTimesCircle}} & \textcolor{red}{\large{\faTimesCircle}} & Discuss the non-IID data and CL processes in FL with potential solutions and research direction.\\ \hhline{|~|------------}
    \rowcolor{orange!15}
    \multirow{-3}{*}[3ex]{\makecell[c]{Non-Centralized \\ Continual Learning}} & 2024 & \cite{survey3_nccl} & \textcolor{olive}{\large{\faCheckCircle}} & \textcolor{olive}{\large{\faCheckCircle}} & \textcolor{olive}{\large{\faCheckCircle}} & \textcolor{red}{\large{\faTimesCircle}} & \textcolor{red}{\large{\faTimesCircle}} & \textcolor{olive}{\large{\faCheckCircle}} & \textcolor{red}{\large{\faTimesCircle}} & \textcolor{red}{\large{\faTimesCircle}} & \textcolor{red}{\large{\faTimesCircle}} & Integrate CL with FL and highlight the benefits of knowledge fusion in improving FCL systems.  \\ 
    \hline \hline
    \rowcolor{gray!10}
    \textbf{\underline{Ours}} & 2024 & - & \textcolor{olive}{\large{\faCheckCircle}} & \textcolor{olive}{\large{\faCheckCircle}} & \textcolor{olive}{\large{\faCheckCircle}} & \textcolor{olive}{\large{\faCheckCircle}} & \textcolor{olive}{\large{\faCheckCircle}} & \textcolor{olive}{\large{\faCheckCircle}} & \textcolor{olive}{\large{\faCheckCircle}} & \textcolor{olive}{\large{\faCheckCircle}} & \textcolor{olive}{\large{\faCheckCircle}} & A comprehensive and advanced tutorial on the NCCL with all aspects.\\ 
    \bottomrule[1pt]
    \end{tabular}}
    \label{survey}
\end{table*}

\begin{figure*}[t]
    \centering
    \includegraphics[width=\linewidth]{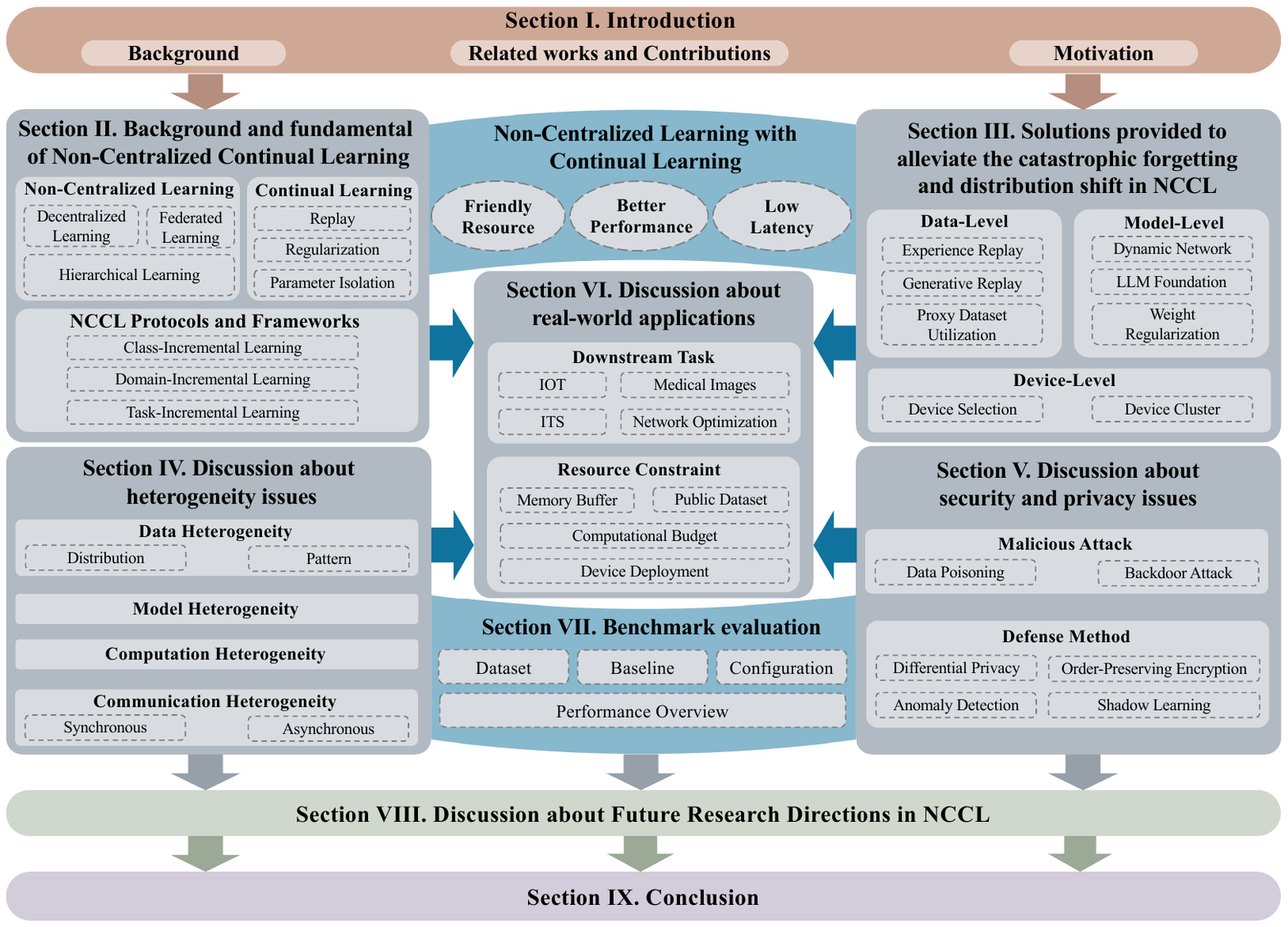}
    \caption{The outline of this survey, where we introduce the provisioning of continual learning in the non-centralized learning paradigm and highlight some essential issues and challenges.}
    \label{structure}
\end{figure*}

To tackle these challenges, Non-Centralized Continual Learning (NCCL) breaks the static limitations of traditional non-centralized learning by learning a task sequence on each distributed device. {We illustrate this new paradigm in Fig. \ref{framework}}
Compared to traditional NCL training approaches, implementing CL for NCL training at distributed devices features the following advantages.
\begin{itemize}
    \item \textit{{Low Resource Consumption}}: Less computational resources and storage are required to learn the new streaming data. For example, instead of training the new data from scratch, devices only continually train the new data with the model converged on previous tasks. As a result, this significantly reduces the computational resources and storage to cache all data of previous tasks.
    \item \textit{Better Performance}: Continual learning algorithms are designed for streaming tasks with effective skills like data rehearsal and regularization methods. Especially on distributed devices, sometimes the data of new tasks is limited, making it difficult for the model to learn the new features alone. In such a scenario, the CL algorithm can leverage similar knowledge from previous tasks to enhance the performance of the new task.
    \item \textit{Low Latency}: With CL, distributed devices can be consistently trained and updated. Meanwhile, in the NCL paradigm, real-time decisions, e.g., event decisions, can be made locally at the edge devices. Therefore, the latency is much lower than that when decisions are made after the model trains from scratch. This is vital for time-critical applications such as recommendation systems, in which delays can potentially cause users to have a poor experience, thereby reducing economic benefits.
\end{itemize}

Given the aforementioned advantages, NCCL has seen recent successes in many applications. For instance, federated continual Learning is introduced to tackle the challenge of learning from streaming tasks in individual clients \cite{survey3_nccl,wang2024traceable}. \cite{yoon2021federated} makes the pioneering effort in this area. Their work focuses on task-incremental learning, requiring distinct task IDs during inference and employing separate task-specific masks to enhance personalized performance. Other studies, such as \cite{ma2022continual}, employ knowledge distillation with a surrogate dataset to retrain the model. Meanwhile, \cite{qi2023better,zhang2023target,wuerkaixi2024accurate} utilizes the generator to synthesize samples to enhance the training process. \cite{li_tpds,li2024towards,li2025tpami} proposed calculating the importance of samples separately for the local and global distributions, selectively saving necessary samples for retraining to mitigate catastrophic forgetting in federated incremental learning scenarios. Additionally, research like \cite{Dong_2023_CVPR,dong2023no} extends NCCL algorithms to medical systems to predict epidemic diseases. In addition, several studies have also explored the use of IoT decentralized learning in some scenarios where a reliable server is absent, e.g., to foster collaboration across government agencies \cite{iot1,iot2}. 

Although many research works investigate the integration of CL algorithms with the NCCL paradigm, several challenges must be solved before NCCL can be scaled. Firstly, the NCCL paradigm is relatively complex and diverse, encompassing various sophisticated paradigms such as federated learning, decentralized learning, hierarchical Learning, and more \cite{distributedlearning,dinh2013survey}. However, the mainstream of existing research focuses on traditional FL scenarios, which rely on the capabilities of a server, among other factors. Secondly, the distribution of each local streaming data is complex in real-world scenarios. Each time a new task arrives, the distribution of the global data changes. Distributed devices should be able to handle dynamic data distributions, especially when such distributions can be caused by combinations of factors such as sample numbers, label changes, and feature skews \cite{li2024towards}. Thirdly, distributed devices like mobile phones are constrained by resources such as storage, computational budget, and sparsely labeled data. The transfer of traditional CL algorithms, which aim to address catastrophic forgetting in the centralized setting, may not work due to insufficient resources \cite{wang2024feddse}. Last but not least, recent research works have clearly shown that a malicious participant may exist in NCL and can infer the information of other participants just from the shared parameters alone \cite{gansun}. In particular, centralized CL algorithms do not guarantee privacy in the presence of distributed devices. As such, privacy and security issues in NCCL still need to be considered.

\subsection{{Related Works and Contributions}}
Although there are surveys on NCL algorithms and CL methods, the existing studies usually treat the two topics separately. For existing surveys on NCL, the authors in \cite{survey1_ncl} place more emphasis on discussing the architecture and categorization of different FL settings to be used for the varying distributions of training data. The authors in \cite{survey2_ncl} highlight the applications of FL in mobile edge networks with resource allocation and communication efficiency. On the other hand, for surveys in CL that focus on implementing CL model training in the centralized setting. For example, the authors in \cite{survey1_cl,survey2_cl} discuss strategies for alleviating catastrophic forgetting and systematically summarize the latest advances in continual learning. Similarly, \cite{survey3_cl} further provides a benchmark evaluation of existing methods for image classification. 

In addition, there are a few existing surveys on federated continual learning. {\cite{survey1_nccl} explores the integration of peer-to-peer federated learning and domain incremental learning within the context of incremental transfer learning. This survey analyzes existing methodologies, highlights technical challenges such as domain drift detection and cross-institutional trust mechanisms, and outlines promising directions for applications in healthcare, finance, and IoT ecosystems.} The focus of \cite{survey2_nccl} is on non-IID data, briefly mentions the process of continual learning in federated learning, and discusses several common scenarios where continuous learning and federated learning can be simply combined. In contrast, the authors in \cite{survey3_nccl} provide a brief tutorial on the knowledge fusion technique in federated continual learning and the challenges related to its implementation but do not consider the issue of complex heterogeneity, privacy \& security, and real-world applications at distributing devices. These surveys also face some common problems, such as incomplete literature coverage (lacking many recent articles published in prominent conferences and journals), relatively simple classification methods that merely migrate the categories of traditional CL in a centralized setting, only focusing on the federated learning scenario among the non-centralized learning paradigm, and neglecting the real-world application of NCCL. 

%
In summary, most existing surveys on NCL do not consider the dynamism of the distributed data collection, whereas surveys on CL do not consider the challenges, including privacy, communication, and heterogeneity. Regarding the few surveys that combine federated learning with continual learning, the analysis of NCCL is not comprehensive and leaves many issues that need to be addressed urgently. We summarize some typical related surveys in comparison to our survey in Table \ref{survey}. {Our survey comprehensively reviews NCCL-related research up to March 2025 (over 200 relevant publications retrieved via Google Scholar) and incorporates mainstream taxonomy frameworks from NCL/CL research along with typical real-world applications.} In this manuscript, we observe that the scenarios and experiments in NCCL research are diverse, and there is a lack of a comprehensive survey with benchmark evaluation for NCCL research, which hinders future research in this field. This motivates us to produce a comprehensive survey with the following contributions:
\begin{itemize}
    \item We motivate the importance of NCCL as an important paradigm shift towards enabling collaborative CL model training on distributed devices. Then, we provide a concise tutorial on NCCL implementation and present a list of useful open-source frameworks for readers that pave the way for future research on NCCL and its applications.
    \item We discuss the unique features of NCCL relative to a centralized deep learning approach and the resulting implementation challenges. We further raise the challenges of heterogeneity, privacy, and application issues in NCCL. For each of these challenges, we present a comprehensive discussion of existing solutions and approaches explored in existing works.
    \item We formulate the Federated Continual Learning (FCL), which is the most extensively studied scenario within the NCCL paradigm, and revisit this problem with a large-scale benchmark and analyze the performance of the state-of-the-art FCL approaches. 
    \item We discuss the challenges and future research directions of NCCL.
\end{itemize}

For the reader’s convenience, we classify the related research to be discussed in this survey in Fig. \ref{structure}. Section \ref{II} introduces the background and fundamentals of NCCL. Section \ref{III} reviews solutions provided to alleviate catastrophic forgetting. Section \ref{IV} discusses heterogeneity across multiple devices in NCCL. Section \ref{V} discusses privacy and security issues. Section \ref{VI} discusses applications of NCCL in real-world scenarios. Section \ref{VII} evaluates some typical existing methods with benchmark experiments. Section \ref{VIII} discusses the challenges and future research directions in NCCL. Section \ref{IX} concludes the paper. We also present a list of common abbreviations for reference in Table \ref{abbr}.

\section{Background and fundamental of Non-centralized Continual Learning}\label{II}
The main objective of this section is to establish a solid foundation for the reader by providing the background and fundamentals of NCCL. We will first analyze three different NCL paradigms and their challenges, then followed by three distinct CL methods and their respective challenges. Finally, we will present three NCCL protocols and discuss the unique challenges and issues associated with NCCL.
\begin{table}[h]
    \centering
        \caption{List of common abbreviations in our survey.}
    \renewcommand\arraystretch{1.2}
      \resizebox{0.8\linewidth}{!}{
    \begin{tabular}{ll}
    \toprule[1pt]
    \textbf{Abbreviation} & \textbf{Description} \\ \hline
    NCL & Non-Centralized Learning \\ 
    NCCL & Non-Centralized Continual Learning \\ 
    IID & Independent and Identically Distributed \\ 
    SGD & Stochastic Gradient Descent \\ 
    DL & Decentralized Learning \\ 
    FL & Federated Learning\\ 
    HL & Hierarchical Learning\\ 
    CF & Catastrophic Forgetting \\ 
    IoT & Internet of Things \\ 
    IoV & Internet of Vehicles \\
    UAV & Unmanned Aerial Vehicle \\
    IL & Incremental Learning \\ 
    LLM & Large Language Model \\
    RAN & Radio Access Networks \\
    ViT & Vision Transformer \\
    EWC & Elastic Weight Consolidation \\
    LoRA & Low-Rank Adaption \\
    KNN & K-Nearest Neighbors \\
    MMFL & Multi-Modal Federated Learning \\
    GAN & Generative Adversarial Network \\
    LDP & Local Differential Privacy \\
    \bottomrule[1pt]
    \end{tabular}}
    \label{abbr}
\end{table}

\subsection{Non-Centralized Learning}
\begin{figure*}
\centering
\includegraphics[width=0.9\textwidth]{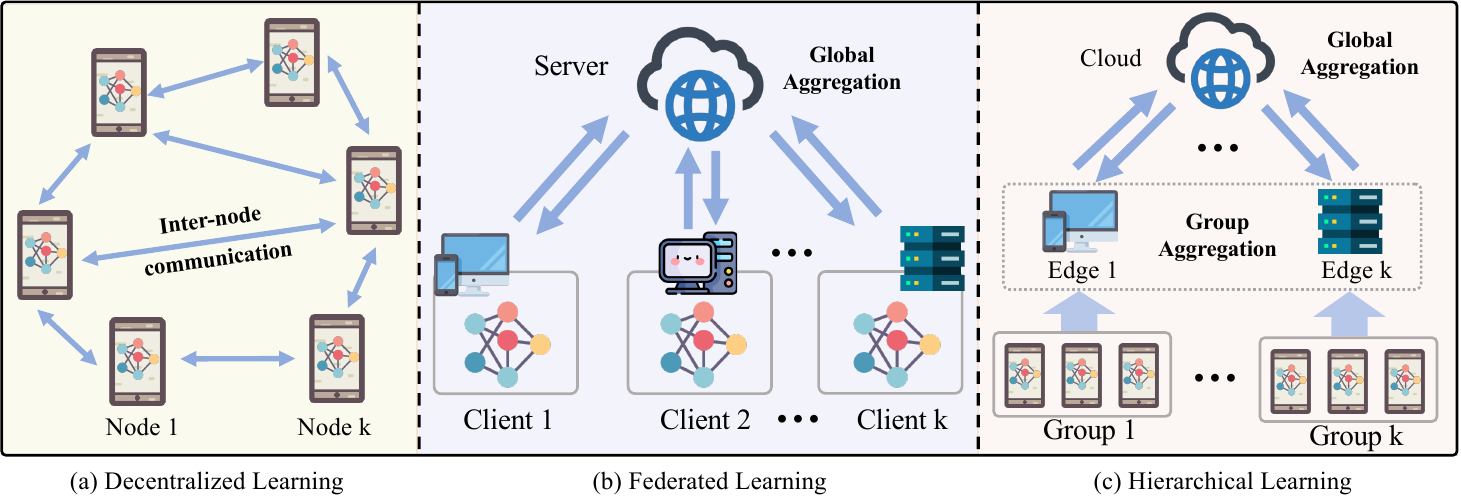}
\caption{Three typical Non-Centralized Learning paradigms include (a) Decentralized learning without the central server; (b) Federated learning with a central server aggregating clients; (c) Hierarchical learning with the client-edge-cloud model.}
\label{ncl}
\end{figure*}

We have selected typical popular deep learning-based paradigms related to the NCL: decentralized learning, federated learning, and hierarchical learning. We provide the basic processes and common challenges associated with these paradigms. We illustrate three NCL paradigms in Fig. \ref{ncl}.

\subsubsection{Decentralized Learning}
Decentralized learning (DL) is a paradigm where central servers are entirely abandoned, leaving only several independent distributed devices \cite{liu2022decentralized,giannakis2017decentralized,hsieh2020non,li2022learning}. Each device trains the local model with a direct exchange of model information between different devices. This approach brings multiple advantages. Firstly, the high scalability avoids the burden of the central server with a large number of devices. Furthermore, data privacy is also ensured since there is no necessity to send data to the central server. In addition, robustness is preserved when dealing with malfunctions or power outages of devices \cite{elgabli2020gadmm}. At the same time, communication overhead presents a substantial challenge, especially in the involvement of numerous devices.

The DL paradigm typically involves a two-step process: local update and inter-node communication (i.e., communication with neighboring nodes. We usually refer to a distributed device as a node in DL). Initially, during the local update stage, each node calculates the gradient of its local loss function and updates the model parameters by performing Stochastic Gradient Descent (SGD) multiple times in parallel. When it comes to the inter-node communication stage, each node exchanges its local model parameters with its connected neighbors and then updates its local model $w_k^{t+1}$ for the next ($t$+1)-th round by averaging the received parameters:
\begin{equation}
    w_{k}^{t+1}={\sum_{i\in S_n}}c_{i}\,w_{i}^t.
\end{equation}
where $S_n$ is the set of neighbors of node $k$, $c_{i}$ is the contribution weight assigned to the model of neighboring node $i$, $w_i^t$ denotes the $t$-th round model parameters of node $i$.

\subsubsection{Federated Learning}
In recent years, with the rising concern of privacy leakage, researchers have focused on finding a trade-off between preserving data privacy and optimizing model training performance. Federated Learning (FL) is proposed for machine learning over distributed local clients, allowing multiple users to contribute to a global model collaboratively by exchanging and aggregating model parameters \cite{kairouz2021advances,wang2024fednlr,wang2023fedcda,wang2024feddse} (we usually refer to a distributed device as a client in FL). Since it involves only the transfer of parameters, the federated learning method significantly reduces communication overhead and effectively preserves data privacy. The foundational federated learning algorithm, FedAvg \cite{mcmahan2017communication}, was introduced by Google in 2017. Since then, numerous studies have been conducted aiming to optimize it further. FedProx \cite{li2020federated} significantly alleviates the heterogeneity issue by adding a regularization term to penalize the divergence of the local models from the global model. MOON \cite{li2021model} applies contrastive learning at the model level by comparing model representations to improve the local training of different clients. SCAFFOLD \cite{karimireddy2020scaffold} handles the client drift issue caused by non-IID data by introducing control variates to adjust and correct the local updates of each client model. 

Then, we will provide a detailed discussion of the federated learning process with three main steps: 1) After determining the training tasks and specifying the hyper-parameters, the server transmits the initialized global model $w^0$ to all clients; 2) In each communication round $t$, a subset of clients is selected randomly. Those participants receive the current global model $w^t$ and perform SGD on local data. After local training, participating clients will upload the updated parameters to the server; 3) In the aggregation phase, the server aggregates the local models and distributes the updated global model $w^{t+1}$ to the clients participating in the next round. Particularly, FedAvg uses a weighted mean to balance the contributions based on the number of local data samples, ensuring that clients with more data have a greater influence on the global model:
\begin{equation}
w^{t+1} = \sum_{k \in S_t} \frac{|D_k|}{|D|} w^{t+1}_k.
\end{equation}
where $|D|$ is the total number of data, $|D_k|$ is the number of data on the client $k$ and $S_t$ is the set of clients participating in the $t$-th round.

\subsubsection{Hierarchical Learning}
Hierarchical Learning (HL) is an advanced paradigm of NCL that enhances both scalability and robustness for edge computing systems \cite{liu2023group,lim2021dynamic,luo2020hfel,chen2023enhanced}. It is structured in a client-edge-cloud framework with improved techniques at the edge level to optimize communication and computation costs (we usually refer to a distributed device as a client and construct edge devices and a cloud server in HL).

The main paradigms of HL leverage hierarchical architectures to improve data privacy and system efficiency by forming groups of clients. Specifically, each edge device manages clients and strategically groups them. These groups are then sent to the cloud server, which employs probabilistic group sampling to select some for training. Clients in the participating groups download a global model for local training and send their updated local models back to the edge server for aggregation within the group:
\begin{equation}
    w^{t,c+1}_g = {\sum_{k\in g}{\frac{|D_k|}{|D_g|}}w^{t,c,e}_k}.
\end{equation}
Where $|D_k|$ is the number of data on client $k$, $|D_g|$ is the number of data for all clients in group $g$, $w^{t,c,e}_k$ is the local model of client $k$ after the $e$-th local round during the $c$-th edge aggregation epoch of the $t$-th cloud round. $w^{t,c+1}_g$ is the group model for group $g$ in ($c$+1)-th edge aggregation round.

After several training rounds, edge servers send the aggregated group models to the cloud for global aggregation. At the end of each global round, the aggregation function is defined as follows, where $S_t$ denotes the participating client groups, $w_{g}^{t,k-1}$ represents the group model at ($k$-1)-th group round within $t$-th global round, and $w^{t+1}_g$ is the aggregated global model at the end of global round $t$.
\begin{equation}
    w^{t+1}_g = {\sum_{g\in S_{t}}}{\frac{n_g}{m_t}}w_{g}^{t,k-1}.
\end{equation}

\noindent\textbf{Challenges of Non-Centralized Learning.} Despite the aforementioned three NCL paradigms serving as a powerful approach that tremendously boosts distributed machine learning, it faces several complex challenges that can hinder general performance and efficiency. Here, we will examine these challenges across three critical aspects: data heterogeneity, data privacy, and resource limitation.
\begin{itemize}
    \item \textit{Data Heterogeneity:} It is characterized by variances in data types, formats, and distributions on different devices, posing a critical threat to distributed learning and degrading performance in real-world scenarios. {Many research studies \cite{li2022towards,hu2021novel,li2022federated,wang2022peer} have contributed to alleviating the non-IID problem through personalization techniques.}
    \item \textit{Data Privacy:} Since transmitting data from edge devices to edge servers still threatens data privacy, one potential solution is to exchange only the parameters of ML models. However, the privacy leakage risk still exists where model parameters may be reversely attacked. {A variety of solutions \cite{xu2022non,huang2022retracted,park2022privacy,gholami2022trusted} have been proposed to address this issue.} For instance, Local Differential Privacy (LDP) is a natural choice, which adds noise without significantly hurting the outcome of any analysis \cite{dwork2014algorithmic}.
    \item \textit{Resource Limitation:} This challenge arises from the constrained computational, communication, or other resources on distributed devices. Limited computational power can slow down training processes, while restricted communication bandwidth can hinder the timely exchange of information between devices and servers. {Researchers have explored various strategies \cite{imteaj2022federated,li2022fairness} to mitigate these resource limitations, such as using lightweight models and reducing communication.}
\end{itemize}

\subsection{Continual Learning}
{Continual Learning (CL), also known as lifelong learning \cite{chen2018lifelong,parisi2019continual,aljundi2017expert} or incremental learning \cite{gepperth2016bio,rosenfeld2018incremental}, refers to the process of acquiring, updating, and applying knowledge over time, much like how humans and other organisms adapt to dynamic environments. It involves learning from a sequential stream of data, where new tasks, domains, or contexts are introduced gradually, while the system aims to retain previously learned knowledge without significant performance degradation, a challenge commonly referred to as catastrophic forgetting. The primary goal of continual learning is to enable AI systems to adapt continuously to changing data and tasks, integrate new knowledge, and improve or maintain performance on earlier tasks without losing previous skills. This is achieved through balancing plasticity and stability, a balance that remains a key challenge in the field. The ultimate aim is to develop systems that can accumulate knowledge over time, transferring what has been learned to better tackle future tasks.}



\subsubsection{Rehearsal-based method:} It involves storing a selection of previous data or generating synthetic data for rehearsal. Through the fusion of overall knowledge from both previous and new data, the model performance can be ensured \cite{rebuffi2017icarl,lopez2017gradient}. {Given the approach, the loss function is the combination of the new $t$-th task loss $\mathcal{L}_{new}(w^t)$ and a rehearsal loss $\mathcal{L}_{rehearsal}(w^t)$ which retains information from previous tasks. Here, $w^t$ denotes the model weights for the $t$-th task. $\alpha$ is a weighting factor that balances the importance.}
\begin{equation}
\mathcal{L}(w)=\mathcal{L}_{new}(w^t)+\alpha\cdot\mathcal{L}_{rehearsal}(w^t).
\end{equation}

\subsubsection{Regularization-based method:} It prevents catastrophic forgetting by adding constraints or regularization terms which penalizes significant changes to the model parameters that are important for earlier tasks \cite{kirkpatrick2017overcoming, zenke2017continual,li2017learning}. The total loss function combines the new task loss and a regularization term $R$ weighted by $\lambda$ that penalizes deviations of the model $w^t$ of the $t$-th task from their optimal values $\hat{w}$ of previous tasks. {Here, $i$ represents the index of the previously learned tasks, and $T-1$ denotes the total number of these prior tasks.
{\begin{equation}
\mathcal{L}(w)=\mathcal{L}_{new}(w^t)+\lambda\sum_{i=1}^{T-1} R(w^t,\hat{w}).
\end{equation}}}

\subsubsection{Parameter isolation-based method:} It isolates distinct parameters to specific tasks, which ensures the learning of new tasks does not interfere with the weights and structures learned in the previous tasks \cite{serra2018overcoming,rosenfeld2018incremental}. Given the approach, the loss function is:
\begin{equation}
\mathcal{L}(w^t)=\mathcal{L}_{new}(w^t)+\alpha\cdot\mathcal{L}_{isolation}(w^t,\tilde{w}).
\end{equation}
where $\mathcal{L}_{isolation}(w^t,\tilde{w})$ denotes the loss measuring differences between new parameters $w^t$ and isolated parameters $\tilde{w}$.

\noindent\textbf{Challenges of Continual Learning.} Despite its ability to handle dynamic streaming data, continual learning (CL) faces several significant challenges: catastrophic forgetting, concept drift, and computational constraints.
\begin{itemize}
    \item \textit{Catastrophic Forgetting:} A primary issue in CL, where models forget previously learned information upon acquiring new data. This occurs during task training as parameter updates adapt to new data, potentially overwriting crucial parameters for prior tasks, leading to performance degradation. The problem intensifies when new and previous tasks have minimal overlap, as learning the new task often conflicts with retained knowledge. {Balancing new and previous knowledge remains a persistent challenge \cite{li2022overcoming,varshney2024prompt,scialom2022fine}.}
    \item \textit{Concept Drift:} Occurs when the statistical properties of the target variable change unexpectedly over time, particularly in the conditional distribution given the input, even if the input distribution remains stable. For instance, consumers' purchasing intentions may evolve while product variety and quantity stay constant. This can prompt the development of drift-aware adaptive learning algorithms to update models online.
    \item \textit{Computational Constraint:} Frequently encountered in CL, posing a dilemma between high processing demands and limited computing capabilities. {To mitigate latency, \cite{prabhu2023computationally} suggests offloading computations to edge servers with GPUs, shifting the workload from server to edges}.
\end{itemize}

\begin{figure}[t]
    \centering
    \includegraphics[width=\linewidth]{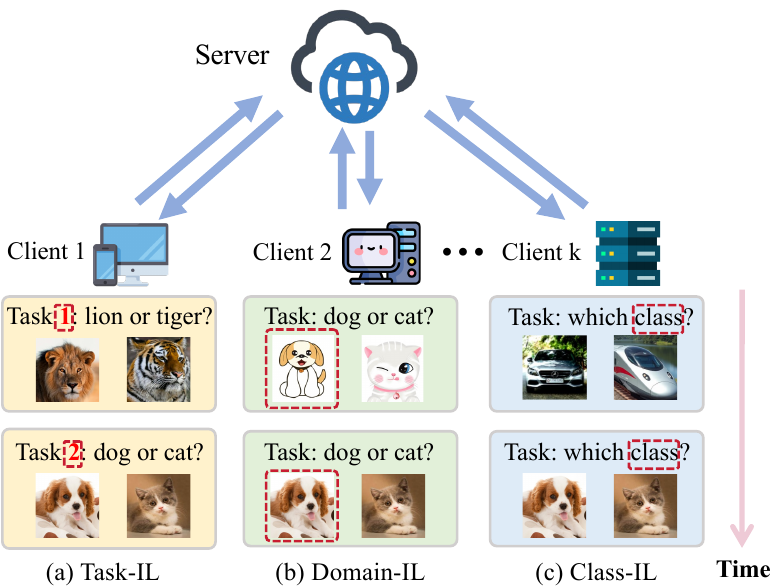}
    \caption{Three Non-Centralized CL protocols. (a) represents the Task-IL scenario, where the boundaries of different tasks (e.g., task-id) are well-defined; (b) is the Domain-IL scenario, where there exists a feature shift among different tasks, but generally no new classes are introduced; (c) is the Class-IL scenario, where the class types of different tasks do not overlap.}
    \label{protocol}
\end{figure}
\subsection{Non-Centralized CL Protocols and Frameworks}
NCCL integrates the task scenarios of NCL and CL, aiming to enable each distributed device to dynamically collect and train streaming tasks. Based on the differences among the streaming tasks of the devices, we categorize such scenarios into the following three types \cite{van2018generative}. These scenarios are designated as Task-Incremental Learning, Domain-Incremental Learning, and Class-Incremental Learning, and we illustrate it in Fig. \ref{protocol}.




\subsubsection{Task-Incremental Learning}
In Task-Incremental Learning (Task-IL), each device progressively acquires knowledge of streaming tasks, with clear boundaries defined for each task. During the inference, the device will Know in advance that the sample belongs to the $t$-th task. Then, device $k$ can narrow down the search range of models based on this prior knowledge, thus improving the performance accuracy. However, in the remaining two scenarios, the boundary of each task is ambiguous and task-id is not available during the reference.

\subsubsection{Domain-Incremental Learning}
Domain-Incremental Learning (Domain-IL) primarily focuses on changes in input distribution or features over time where the structure of the tasks remains the same. For each task's training dataset $T_{k}^{t}$ at device $k$, all tasks in domain incremental learning share the same number of categories. Each task contains all categories, i.e., $C_k^t = C_k^1$, but the data distribution will shift between different tasks, i.e., $X_{k}^{t}\neq X_{k}^{t+1}$.

\subsubsection{Class-Incremental Learning}
Class-Incremental Learning (Class-IL) involves the scenario where the device progressively learns to recognize new classes over time. Unlike Task-IL, where the device knows which task a sample belongs to during inference, in Class-IL, the device will classify samples into any of the classes it has encountered so far. Initially, the device is trained on a subset of classes. Subsequently, as new classes are introduced, i.e., $C_k^t \neq C_k^1$, the device updates its model to incorporate this new knowledge without forgetting the previously learned classes.

\subsection{Unique Challenges and Issues of Non-Centralized CL}
Although Non-centralized continual learning can be viewed as integrating CL algorithms into the NCL paradigm, it will introduce the following unique challenges.
\begin{itemize}
    \item \textit{Local-Global Knowledge Forgetting}: On the one hand, distributed devices locally collect new data, and when models are trained on this new data, they may overwrite the knowledge learned from old data. On the other hand, data collection efficiency varies between different devices and there are issues with asynchronous communication. Devices that collect data quickly and communicate frequently tend to dominate the knowledge in the global model during global aggregation, potentially overwriting the knowledge from other devices. We summarize this challenge as local-global knowledge forgetting that will significantly impact the performance of NCCL systems.   
    \item \textit{Spatial-Temporal Distribution Shift}: In NCL, different distributed devices often possess different data distributions. This situation evolves further in NCCL, where each distributed device collects new local data during the training process, potentially leading to a shift from its original data distribution. We refer to this phenomenon as the spatial-temporal distribution shift and consider it a significant yet unique challenge in NCCL.
    \item \textit{Deteriorating Privacy Threats}: Preserving data privacy for distributed devices has always been a paramount challenge in NCL and this issue becomes even more prominent in NCCL due to the need for privacy protection for each task. Furthermore, there may be certain correlations among different tasks, potentially facilitating attackers in obtaining a series of data privacy more conveniently. Meanwhile, to address the aforementioned two challenges, existing methods often employ additional techniques that involve a trade-off between performance and privacy. Therefore, we emphasize here that the issue of privacy and security in NCCL is a challenge that cannot be ignored.
    \item \textit{Ongoing Resource Overhead}: Resource overhead has been a popular research topic in NCL, as many IoT devices have limited hardware resources, preventing the realization of idealized local training and global communication. However, CL is often challenging and requires sophisticated algorithms and computational resources. Reducing ongoing resource overhead is a necessary challenge for the deployment of NCCL in practical scenarios.
\end{itemize}

In Section \ref{III}, we will first delve into existing methods with advanced techniques to alleviate catastrophic forgetting and distribution shift at the data, method, and device levels. Then, the following sections will review related work and address heterogeneity, privacy \& security, and real-world application issues separately.

\section{Solutions provided for alleviating catastrophic forgetting and distribution shift in Non-Centralized Continual Learning}\label{III}
This section outlines a series of targeted solutions designed to tackle catastrophic forgetting and distribution shifts within NCCL. Considering the research methods of both NCL and CL, we categorize the existing studies into data-level, model-level, and device-level, with 2-3 major techniques.
\begin{itemize}
    \item \textit{Data-Level Method:} This method primarily centers on the various processing techniques that devices use to manage their local data. A key aspect of this approach involves the innovative reuse of existing data, which helps to strengthen the model's understanding and retention of previously learned tasks. To achieve this, the method employs strategies such as experience replay, where cached samples from past tasks are revisited \cite{li2025tpami}, or the generation of synthetic data for replay purposes \cite{qi2023better}. These techniques ensure that the model remains capable of continuously learning and adapting from an ongoing sequence of tasks. Additionally, the method can incorporate knowledge distillation using proxy datasets \cite{ma2022continual}. This process further refines the learning experience by extracting and distilling the core knowledge and essential features of the learned tasks into a more compact and efficient model. 
    \item \textit{Model-Level Method:} This category represents the advancement and evolution of learning algorithms in NCCL. These techniques are primarily concerned with refining and enhancing the mechanisms of the model, enabling it to resist forgetting previously learned information naturally. One such technique is weight regularization, which promotes continuity in the learning process by imposing penalties on updates that could disrupt or overwrite the model's memory of prior tasks \cite{cox2023parameterizing,wu2025personalized}. Meanwhile, the dynamic network allows the model to remain adaptable to new challenges and tasks while preserving its foundational knowledge. This approach ensures that the model can integrate new information without compromising its existing understanding \cite{li2024pfeddil,chen2023flexibility}. Additionally, incorporating Large Language Model (LLM) foundations marks a significant breakthrough in this method. LLMs provide a robust and comprehensive framework for knowledge retention, enabling the model to store and access vast amounts of information in a structured and efficient manner \cite{he2024fppl}. 
    \item \textit{Device-Level Method:} This approach focuses on harnessing the collective intelligence of NCCL through optimizing device participation. Devices possessing complementary datasets work together to foster a comprehensive understanding. By employing meticulous selection criteria and clustering techniques for distributed devices, these strategies augment the capacity for learning and retention \cite{li_tpds,yoon2025pick}.
\end{itemize}

\subsection{Data-Level}
\begin{figure}
    \centering
    \includegraphics[width=\linewidth]{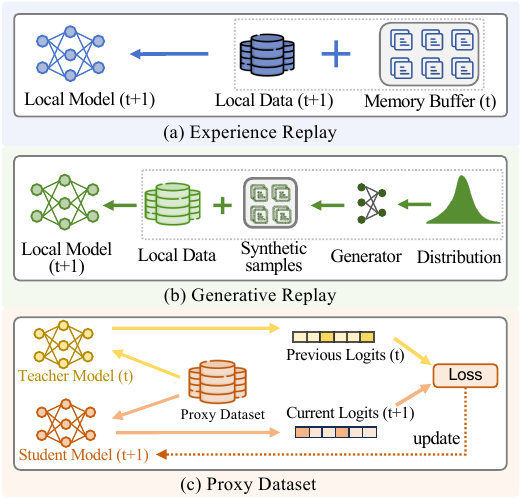}
    \caption{Data-level methods, including experience replay, generative replay, and proxy dataset. Experience replay focuses on reusing data and features stored in a local buffer; generative replay often relies on generative models to synthesize pseudo-data for replay; and a proxy dataset serves as a medium to facilitate knowledge transfer between different tasks.}
    \label{data-level}
\end{figure}

The data-level methods focus on the strategic use of data to mitigate forgetting. We will explore three different techniques: using cached data from past tasks for replay, synthesizing additional data to assist training, and leveraging external proxy datasets to transfer knowledge between tasks through knowledge distillation. We depict these data-level techniques in Fig. \ref{data-level} and summarize current main methods in Table \ref{Data-Level Methods}.

\begin{table*}[h]
\centering
\caption{A Summary of Main Data-Level Methods in NCCL.}
\label{Data-Level Methods}
\renewcommand{\arraystretch}{1.1}
\begin{tabular}{|m{0.05\textwidth}|m{0.45\textwidth}|m{0.20\textwidth}|m{0.20\textwidth}|} 
\hline 
\textbf{Ref} & \textbf{Key Ideas} & {\textbf{Advantages}} & {\textbf{Limitations}} \\ 
    \hline
    \multicolumn{4}{|c|}{\raisebox{-1.8ex}[0pt][0pt]{\centering \textbf{Data-Level: Experience Replay}}} \\[3ex]\hline
    \cite{gong2024delta} & Employing a soft data matching strategy to align the on-device data distribution with the cloud-side directory dataset. & Enhanced accuracy, privacy, and reduced costs.& Dependent on data quality and limited adaptability. \\ \hline
    \cite{alhamoud2024fedmedicl} & Utilizing experience replay to evaluate distribution shifts in federated medical imaging. & Effective class balancing and pandemic simulation.& Limited attribute diversity and dataset-specific performance.\\ \hline
    \cite{dahdal2024roamml} & Using replay buffers to cache selected samples from the agent’s journey across the network for disaster relief operations. & Significant bandwidth savings and adaptability to disaster scenarios. & Reliant on the data gravity concept, sensitive to node mobility and network conditions.\\ \hline
    \cite{revathi2024exploring} & Enabling nodes to dynamically update their models with new phishing data streams without data accumulation with replay buffers. & Attention mechanism captures phishing patterns effectively. & Depending on data streams and the replay buffer may cause latency.\\ \hline
    \cite{zhuang2024coala} & Using data manager component for data loading, partitioning, and distribution across clients, enabling dynamic updates and continual learning in federated continual learning. & Comprehensive vision-centric FL platform that supports diverse tasks and data patterns. & Complexity in customization, potential scalability issues with large-scale deployment. \\ \hline
    \cite{serra2024federated} & Utilizing an uncertainty-aware memory management strategy, which selectively retains and replays critical samples in an online data stream. &Uncertainty-aware memory reduces forgetting in online FCL. & Requires test-time augmentation for Bregman Information estimation. \\ \hline
    \cite{le2024distributed} & Using receiver-initiated data sharing approach, where an agent requests and receives the most relevant data instances from its neighbors to address local knowledge gaps, thereby facilitating incremental learning and adaptation. &Modular parameter sharing enhances performance with low communication cost.&Focuses on supervised learning; lacks reinforcement learning extension.\\ \hline
    \multicolumn{4}{|c|}{\raisebox{-1.8ex}[0pt][0pt]{\centering \textbf{Data-Level: Generative Replay}}} \\[3ex]\hline
    \cite{mei2024using} & Leveraging conditional diffusion models to generate synthetic historical data at local devices. &Diffusion models reduce forgetting effectively.&Computationally intensive diffusion training limits scalability. \\ \hline
    \cite{zhang1federated} & Using memory generator model on the central server, which creates pseudo data to mimic the distribution of historical tasks, thereby facilitating knowledge retention across sequential learning tasks. &Central memory rehearsal mitigates catastrophic forgetting efficiently.&Depends heavily on server-side computation. \\ \hline
    \cite{li2024facing} & Utilizing a self-challenge diffusion model to generate synthetic data by masking and recovering information within samples. &Self-challenge replay and gradient balance address spatiotemporal heterogeneity effectively. &Computational overhead from diffusion models on original data scale. \\ \hline
    \cite{liu2024adaptive} & Using a feature generator to synthesize features for previous tasks, leveraging knowledge distillation and dynamic adaptive weight allocation & Deconfounded graph transfer learning enhances cross-city scheduling adaptability. &Continual learning's dynamic nature complicates cross-city transfer. \\ \hline
    \cite{yoo2024federated} & Generating pseudo features from prototypes to replace the need for rehearsal memory, thus reducing communication costs and enhancing privacy. &Pseudo feature generation reduces communication costs without rehearsal memory. &Fixed feature extractor and reliance on PCA may limit adaptability and flexibility. \\ \hline
    \cite{churamani2024feature} & Employing a local generator model at each client to efficiently pseudo-rehearsal latent features for replay. &Efficient feature aggregation and pseudo-rehearsal. & Generator struggles with complex features. \\ \hline
    \multicolumn{4}{|c|}{\raisebox{-1.8ex}[0pt][0pt]{\centering \textbf{Data-Level: Proxy Dataset}}} \\[3ex]\hline
    \cite{wu2024federated} & Employing new-class augmented self-distillation in federated class-incremental learning to mitigate catastrophic forgetting by enriching historical model class scores with new class scores predicted by current models. &Distillation harmonizes old and new class scores for effective knowledge transfer. &Distillation relies on accurate historical model outputs. \\ \hline
    \cite{9821057} & Utilizing distillation techniques to mitigate catastrophic forgetting in pervasive computing environments, enabling models to retain the knowledge of previously learned tasks while adapting to new tasks. & Distillation effectively preserves knowledge from past and server models. & Distillation underperforms on balanced tasks and increases complexity.\\ \hline
    \cite{babendererde2025federated} & Leveraging a proxy dataset to evaluate model updates based on their continuity, thus ensuring robustness against spatiotemporal data shifts. & Enhanced performance with promising privacy. & Limiting its applicability in scenarios with scarce public data. \\ \hline
\end{tabular}
\end{table*}

\subsubsection{Experience Replay} Its main idea is to use cached data with limited memory buffers to help models retain the knowledge of previous tasks as they learn new ones, which reduces catastrophic forgetting and alleviates distribution shifts. In \cite{xie2024fedmes}, the authors propose a framework named federated memory strengthening (FedMeS) to address the challenges of personalized federated continual learning. FedMeS leverages local memory at each client to store samples from previous tasks, which are then used to both calibrate gradient updates during training and enhance inference via KNN-based Gaussian inference. During the training process, FedMeS uses a novel regularization term that is dynamically adjusted based on the current loss. 
A dynamic adjustment is employed to help draw useful information from other clients through the global model, facilitating the learning of local tasks. During inference, FedMeS utilizes the local memory to perform KNN-based Gaussian inference. For a test sample, the K nearest neighbors from the memory are found based on the Euclidean distance in the feature space. The local prediction is then computed using a Gaussian kernel.
The final prediction of FedMeS is obtained by a convex combination of the outputs from the local, personalized model and the KNN inference.
Experimental results show that FedMeS significantly outperforms existing baselines in terms of average accuracy and forgetting rate across various datasets and task distributions. For example, on the Split CIFAR-100 dataset with 10 clients, FedMeS achieves an average accuracy of 53.0\% and a forgetting rate of 0.06\%, compared to the state-of-the-art baselines.

Inspired by \cite{sebaa2024life,rolnick2019experience}, the authors in \cite{li2024towards} propose Re-Fed to enable each client to cache samples based on their calculated importance scores, which are determined by both global and local understanding. This is particularly important given the limited storage capacity at edge clients. Upon the arrival of a new task, each client selects samples for caching using a personalized, informative model. The importance score for each sample is quantified by the gradient norm concerning the personalized informative model parameters.
The clients then train their local models with both the cached samples from previous tasks and the samples from the new task, which theoretically helps to alleviate catastrophic forgetting.

Similarly, in \cite{li_tpds}, the authors aim to address the federated domain-incremental learning with the proposed method SR-FDIL. SR-FDIL is designed to mitigate the effects of catastrophic forgetting by employing a synergistic replay mechanism, allowing clients to cache and replay samples crucial for learning from incremental data with domain shifts. A distinctive feature of SR-FDIL is the use of generative adversarial networks (GAN) to assign a cross-client collaborative score, which is pivotal for determining the global importance of samples. GAN has two main components: a generator and a discriminator. The generator's role is to produce samples that mimic the data distribution, while the discriminator's role is to distinguish between real samples from the data distribution and fake samples generated by the generator. 
In the context of the SR-FDIL method, the discriminator with parameters is repurposed to assess the global importance of samples rather than distinguishing between real and generated samples. 
This score is used to determine the significance of the sample in the context of the entire dataset, which is crucial for the sample selection process in SR-FDIL. The local importance is quantified through a domain-representative score, which measures the similarity of a sample to the local prototype using cosine similarity.
The final importance score for each sample is a fusion of both local and global scores, weighted by a hyperparameter.
Extensive experiments on various datasets (Digit-10, Office-31, and DomainNet) have been done to illustrate that SR-FDIL can outperform the state-of-the-art methods by up to 4.05\% in average accuracy on all domains.  

\subsubsection{Generative Replay} It allows each device to synthesize data for replay when the data from previous tasks involves privacy concerns or local storage is insufficient to cache data. In \cite{qi2023better}, the authors apply ACGAN \cite{odena2017conditional} as the base model and introduce FedCIL, which employs the model consolidation and consistency enforcement of two modules to stabilize the training process and enhance the performance of generative replay in the continual federated learning. On the server side, the model consolidation phase integrates parameters from participating clients to initialize the global model. Subsequently, balanced synthetic data is generated using data provided by the clients. 
On the client side, consistency enforcement is achieved by aligning the output logits of the classification module with those from the global model and previous local models. This alignment is facilitated through a set of consistency loss functions, which are integrated into the overall loss function.
This comprehensive approach ensures that the global model remains robust and capable of classifying all classes learned by the clients thus far without suffering from forgetting.

Based on this, the authors in \cite{zhang2023target} propose to address the challenge of federated class-continual learning by leveraging knowledge from the global model without requiring real data from previous tasks. On the server side, it uses a generator to synthesize data that conforms to the global model distribution. 
On the client side, it trains the local model on the real data of the current task and the synthetic data generated for the previous tasks. The training objective combines a cross-entropy loss for the current task and a KL divergence loss for the previous tasks.
To further improve the quality of the synthetic data, it employs a student model in addition to the generator. The student model assists in training the generator by providing a distillation loss.
This distillation loss helps the generator produce more diverse and valuable synthetic data, enhancing the knowledge transfer from the global model to the local models. Extensive experiments demonstrate the method achieves an accuracy of 36.31\%, which is about 6\% higher than the best baseline method on CIFAR-100.

{Recently, the authors in \cite{wuerkaixi2024accurate} further propose AF-FCL, which utilizes a normalizing flow (NF) model trained globally to model the feature distribution of previous tasks. The NF model is trained to maximize the likelihood of the input features from client data, as well as sampled features from the previous NF model. During local training, the classifier is optimized using a multi-loss objective consisting of cross-entropy losses on real data, generated data, and a knowledge distillation loss in the feature space. The generated data loss is re-weighted based on the correlation probability of the generated features with the current task distribution.}

{Similar to \cite{khademi2025federated,sun2024exemplar,MFCL}, authors in \cite{salami2024federated} proposed hierarchical generative prototypes that are a combination of prompt-based fine-tuning and generative replay. Specifically, this method constrains the biases in the last layer of the model by using learnable prompts, ensuring that the clients' representations remain close to the pre-training optimum and minimizing federated bias. The server then re-balances the global classifier using generative prototypes. Each client approximates the feature distribution of each class using a multivariate Gaussian distribution.
The server identifies the global distribution for each class that minimizes the Jensen-Shannon divergence among all client distributions.
The server samples synthetic feature vectors to re-balance the global classifier, effectively mitigating biases in the classification layer. By leveraging prompt-based fine-tuning and hierarchical generative prototypes, the method achieves state-of-the-art performance while maintaining minimal communication costs.}

{To address catastrophic forgetting in spatiotemporal prediction on streaming data, \cite{miao2025spatio} integrates historical data synthesis, local model training with spatiotemporal mix-up, and federated training to preserve historical knowledge while protecting data privacy. The framework synthesizes historical data using a generator that produces synthetic spatiotemporal data, which is stored in a global replay buffer to simulate the global data distribution without compromising privacy. Each client then fuses current training data with this synthetic data using a spatiotemporal mix-up mechanism, which interpolates between current observations and historical observations sampled from the replay buffer.
This interpolation encourages the model to behave linearly across streaming spatiotemporal data sequences, effectively minimizing catastrophic forgetting.}

\subsubsection{Proxy Dataset} It assumes that the device or server can acquire other proxy datasets and use them for knowledge distillation to facilitate the transfer of previous and new knowledge. The authors in \cite{wu2024federated} propose FedCLASS mitigate catastrophic forgetting in federated class-incremental learning scenarios where data volume and class diversity expand over time. The core idea of FedCLASS is to use new-class augmented self-distillation, where class scores of new classes predicted by current models are utilized to enrich the class scores of historical models. This augmented knowledge is then used for self-distillation, enabling more precise knowledge transfer from historical to current models. During the training process, FedCLASS maintains a historical model on each client that represents the model's state before the current incremental task. For each sample, FedCLASS reconstructs the historical model's output by scaling the old class scores and appending new class scores predicted by the current model. This reconstructed output serves as a soft label to guide the training of the current model through self-distillation. Experimental results show that FedCLASS significantly outperforms baseline methods in terms of reducing the average forgetting rate and improving global accuracy. Across four datasets with different class-incremental learning settings, FedCLASS consistently achieves higher global accuracy and lower average forgetting rates compared to state-of-the-art methods. For instance, on the EMNIST dataset with three incremental tasks, FedCLASS achieves a global accuracy of 74.44\% and an average forgetting rate of 22.05\%, while the best baseline method only reaches 61.79\% global accuracy with a 42.43\% average forgetting rate.

In \cite{9821057}, the authors introduce two approaches, FLwF-1 and FLwF-2, both of which leverage knowledge distillation to transfer knowledge between models. In FLwF-1, knowledge distillation is employed to transfer knowledge from the past model of a client (teacher) to its current model (student). The distillation loss is computed as the difference between the temperature-scaled logits of the teacher and student models.
FLwF-2 extends FLwF-1 by introducing a second teacher model: the current server model. This allows the client model to benefit from the general knowledge encapsulated in the server model and its own past knowledge. The distillation loss for the server model is similarly defined, and the overall loss function for FLwF-2 combines the classification loss, the distillation loss from the past client model, and the distillation loss from the server model.
By incorporating knowledge distillation with one or two teacher models, the proposed FLwF approaches aim to mitigate catastrophic forgetting, enabling clients to retain knowledge of previously learned tasks while adapting to new tasks in a continual learning setting.
{Similarly, the authors in \cite{gai2025mufti} employ a dual-domain knowledge distillation approach. For the $t$-th task, the framework uses two teacher models to guide the student model training. The inter-domain knowledge distillation loss.
This loss ensures that the model retains knowledge from previous collaborative training. Additionally, the intra-domain knowledge distillation loss is defined similarly to ensure that the model retains knowledge specific to the local domain. Combining these losses can effectively balance knowledge across different tasks.}


\subsection{Model-Level}
\begin{figure}[t]
    \centering
    \includegraphics[width=\linewidth]{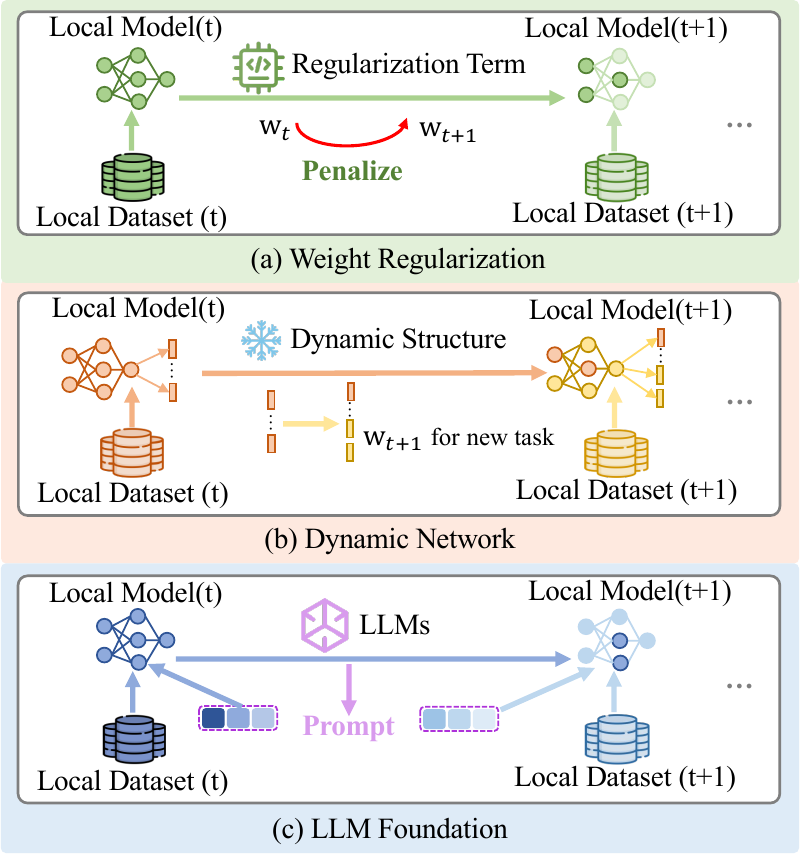}
    \caption{Model-level methods, including weight regularization, dynamic networks, and LLM-based methods. Regularization-based methods penalize parameter updates through regularization terms, dynamic network-based approaches retain past knowledge by modifying the network structure, and LLM-based methods guide knowledge retention and parameter updates by integrating knowledge from LLM.}
    \label{model-level figure}
\end{figure}

These solutions modify the learning process to improve knowledge retention and three effective techniques will be explored in this section. Weight regularization methods, like prior task focus, penalize updates harmful to past tasks, promoting stable task representations. Dynamic Network strategies, such as parameter sharing and freezing, alter model architecture to retain knowledge. Parameter Sharing finds task commonalities, while freezing preserves learned features by restricting updates. Large language models use prompt-based and fine-tuning approaches. Prompt-based methods guide models to integrate cross-task knowledge via designed prompts, enhancing transfer and reducing forgetting. Fine-tuning optimizes retention and adaptation by adjusting key parameters, mitigating forgetting, and protecting privacy. We depict these three model-level methods in Fig. \ref{model-level figure} and summarize the currently available main methods in detail in Table \ref{Model-Level Methods}.

\begin{table*}[h]
\centering
\caption{A Summary of main Model-Level methods in NCCL.}
\label{Model-Level Methods}
\renewcommand{\arraystretch}{1.1}
\begin{tabular}{|m{0.05\textwidth}|m{0.45\textwidth}|m{0.20\textwidth}|m{0.20\textwidth}|} 
\hline 
\textbf{Ref} & \textbf{Key Ideas} & {\textbf{Advantages}} & {\textbf{Limitations}} \\ 
    \hline
    \multicolumn{4}{|c|}{\raisebox{-1.8ex}[0pt][0pt]{\centering \textbf{Method-Level: Weight Regularization}}} \\[3ex]\hline
    \cite{yu2024pi}  & PI-Fed implements parameter-level importance aggregation to regularize gradients for continual federated learning, mitigating catastrophic forgetting without the need for experience replay. & No Experience Replay Required and Enhancing Efficiency. & Slight increase in communication overhead compared to vanilla FL.  \\ 
    \hline
    \cite{yu2024overcoming} & Leveraging class-specific binary classifiers based on variational auto-encoders to overcome catastrophic forgetting. & Improved efficiency with selective knowledge sharing. & High storage requirement for prototype rehearsal.\\ 
    \hline 
    \cite{tan2024fl} & Integrating plasticity and stability in two training stages, employing a task-weighted auxiliary loss for class imbalance and a lightweight distillation method with pre-trained federated models. & Lightweight distillation to reduce communication. & Dependence on pre-trained model architecture. \\ \hline 
    \cite{wang2025vertical} & Evolving prototypes to transfer knowledge across tasks and constrain local model updates to mitigate catastrophic forgetting. & Effective knowledge retention via evolving prototypes and parameter constraints. & Increased communication and computational overhead. \\ \hline 
    \cite{yu2024addressing} & Leveraging frozen pre-trained ViT and trainable tail anchors to stabilize feature positions in the spatial-temporal feature space. & Reduced parameter-forgetting and privacy-preserving. & Increased computational cost due to synthetic data generation. \\ \hline
    \cite{xu2024federated} & Combining automatic client weight optimization and dual-domain knowledge distillation to mitigate data heterogeneity and catastrophic forgetting in federated continual learning. & Effective on Non-IID data and improved performance. & Bringing additional potential privacy risks. \\ \hline 
    \cite{zhao2024towards} & Accumulating global and federated Fisher information matrices to balance parameter importance across tasks and clients, mitigating catastrophic forgetting in federated continual learning. & Robust privacy-preserving with enhanced performance. & Dependence on server validation data. \\
    \hline
    \multicolumn{4}{|c|}{\raisebox{-1.8ex}[0pt][0pt]{\centering \textbf{Method-Level: Dynamic Network}}} \\[3ex]\hline  
    \cite{wang2024fedfrr} & Adjusting primary network units and applying weight truncation to enhance model memory and reduce interference. & Multi-PNU architecture for enhanced memory capacity. & High computational complexity due to multiple PNUs. \\  
    \hline
    \cite{yuan2024gcpn} &  Employing group connected and scaling layers to minimize additional parameters for new tasks in VFL. & Reduced parameters compared to initial ones for new tasks. & Complexity from group and scale Layers.  \\  
    \hline
    \cite{yang2024resource} & Decomposing the global model into an adapter for new task learning and a retainer for preserving previous knowledge. & Low storage overhead via knowledge retention. & Dependency on feature similarity for specific datasets.\\  
    \hline    
    \multicolumn{4}{|c|}{\raisebox{-1.8ex}[0pt][0pt]{\centering \textbf{Method-Level: LLM Foundation}}} \\[3ex]\hline 
    \cite{he2024fppl} & Integrating prompt tuning with a fusion function on the client side and prototype-based de-biasing on the server side to address catastrophic forgetting and non-IID data distribution. & Resource-efficient model decomposition. & Complexity of dynamic feature aggregation. \\  
    \hline
    \cite{he2024masked} & Utilizing client-side reconstructive prompts and server-side aggregated restoration information to fine-tune classifiers, mitigating forgetting and non-IID effects. & Parameter-efficient learning via prompt tuning. & Dependence on the pre-trained model and large computational overhead.  \\  
    \hline
    \cite{salami2024closed} & Achieving closed-form merging of LoRA modules by alternating between optimizing A and B low-rank matrices, ensuring model response alignment across tasks and clients. & Fast convergence with privacy-preserving. & Dependent on low-rank assumptions and task-specific module storage.\\  
    \hline
    \cite{salami2024reducing} & Employing prompt tuning and hierarchical generative prototypes to constrain and re-balance biases in the classification layer, enhancing model adaptability and communication efficiency. & Efficient communication with privacy-preserving. & Additional computational overhead and Dependence on prototype quality.\\  
    \hline
    \cite{peng2024learning} & Leveraging cross-task and cross-client knowledge transfer to generate task-specific initializations for efficient adapter fine-tuning through learning attentive weights. & Parameter-efficient and faster learning through informed initialization. & Higher computational overhead and dependence on the historical model.\\ 
    \hline
\end{tabular}
\end{table*}

\subsubsection{Weight Regularization} It penalizes weight updates that degrade performance on previously learned tasks \cite{ritter2018online,schwarz2018progress}, anchoring the model to essential knowledge and mitigating catastrophic forgetting. This strategy stabilizes the model's representation by preventing significant changes to weights important for previous tasks.
{In \cite{feng2025cgofed}, the authors propose CGoFed, a constrained gradient optimization strategy for federated class incremental learning, which addresses catastrophic forgetting and non-IID label distribution challenges. The core innovation lies in two synergistic modules: relax-constrained gradient update and cross-task gradient regularization. The relax-constrained gradient update mitigates catastrophic forgetting by dynamically restricting gradient updates to subspaces orthogonal to historical task gradients. Unlike strict orthogonal projection, this module introduces a variable constraint strength coefficient that decays exponentially with task progression and adaptively tightens when forgetting exceeds a threshold.
This ensures stability for old tasks while preserving plasticity for new tasks.
To address non-IID label distributions across clients, the cross-task gradient regularization module leverages historical models from other clients. The server computes a similarity matrix using L2-norm distances between representation matrices and selects relevant historical tasks to construct a regularization loss.}

In \cite{bakman2023federated}, the authors propose a method called federated orthogonal training that alleviates the catastrophic forgetting problem by ensuring that the updates for the new task are orthogonal to the global principal component space of activations from the previous tasks at each network layer. The methodology involves maintaining an orthogonal set for each layer, which captures the global principal subspace for all tasks. Similar to \cite{saha2021gradient}, when training on a new task, the server projects the aggregated updates onto the orthogonal, effectively minimizing interference and preserving the performance on older tasks.
The orthogonal vector set is updated through the global principal subspace extraction step, carried out in an additional communication round after each task. During this round, the server broadcasts the current global model parameters and global principal subspace to all clients. Each client then processes its local data and projects their layer inputs onto the subspace orthogonal.
The server then applies singular value decomposition from the matrix formed from these projected inputs to determine the global principal subspace. Empirically, in benchmarks such as Permuted MNIST and Split CIFAR-100, the method has shown a 15\% increase in average accuracy and a 27\% reduction in forgetfulness with minimal increase in computational and communication costs.

In \cite{shoham2019overcoming}, the authors propose FedCurv, an elastic weight merging-based scheme \cite{kirkpatrick2017overcoming} that mitigates the catastrophic forgetting problem in joint learning of non-IID data. EWC prevents catastrophic forgetting by penalizing changes to parameters important for previous task A when learning a new task B. FedCurv applies this idea to federated continual learning by adding a regularization term that encourages all local models to converge to a shared optimum. 
FedCurv can effectively prevent the model drift of each node and alleviate the catastrophic forgetting problem. Experiments are conducted on the MNIST dataset, simulating a federated continual learning scenario with 96 devices and non-IID data distribution. 

\subsubsection{Dynamic Network}
The core idea of the dynamic network strategy involves adjusting the model structuring to preserve the acquired knowledge from previous tasks. By implementing parameter sharing for specific model components, the model can distinguish and retain consistent features across various tasks \cite{hadsell2020embracing}, enhancing the generalization of the learned representations. Conversely, parameter freezing strategically keeps partial parameters fixed \cite{rusu2016progressive,xu2018reinforced}, thereby retaining task-specific features that have been previously learned. This approach to adapting the model's parameter structure ensures proficiency in previously trained tasks while allowing for continuous updates to incorporate new information, effectively mitigating the issue of catastrophic forgetting.

Similarly, in \cite{le2021federated}, a strategy based on broad learning is developed as a local-independent training solution, where the local training can be performed independently without relying on the global model's knowledge. The core innovation is introducing a weighted processing strategy that effectively mitigates catastrophic forgetting. This strategy computes a weighted average of the global model weight and the client's local model weight upon receipt from the cloud server.
This approach ensures that the updated global model retains previously learned knowledge while adapting to new data. Furthermore, the broad learning technique can effectively support incremental learning by updating the parameters when new data is introduced without retraining the deep architecture. When new data arrives, it enables the model to be incrementally updated without full retraining. The updated feature and enhancement nodes are calculated using the same random weights and biases from the initial setup.
Moreover, the broad and flat structure of broad learning networks, with feature and enhancement nodes, enables them to effectively capture the information from new data without relying on deep architectures. This further contributes to the efficiency of the learning process. Empirical evaluations conducted on MNIST and other datasets demonstrate that it not only achieves superior prediction accuracy compared to existing FL schemes but also effectively handles imbalanced and non-IID datasets, outperforming other incremental learning schemes.

In \cite{li2025personalized}, the authors propose a parameter-sharing approach called pFedDIL, which enables each client to adaptively select an appropriate incremental task learning strategy based on the knowledge matching strength and transfer knowledge from previous tasks. To identify the correlations between the new task and previous tasks, pFedDIL employs an auxiliary classifier for each personalized model. The knowledge matching intensity at each client is calculated for each new task by averaging the outputs of the auxiliary classifier across all samples in the new task.
Based on this metric, clients can either continue training with a previous model that shares similar knowledge characteristics or start with a newly initialized model for the new task. During training, inspired by ensemble learning \cite{dong2020survey}, each client migrates knowledge from other personalized models using the knowledge matching intensity.
Furthermore, pFedDIL shares partial parameters between the target classification model and the auxiliary classifier to condense model parameters. During inference, the final result is obtained by aggregating predictions from all personalized models weighted by the normalized output of the auxiliary classifiers. Experiments on Digit-10, Office-31, and DomainNet datasets demonstrate that pFedDIL outperforms state-of-the-art methods by up to 14.35\% in terms of average accuracy across all tasks.

\subsubsection{LLM Foundation}
Recently, approaches to LLMs have been introduced in NCCL, which leverage the generalization capabilities of pre-trained models to optimize model adaptation and knowledge retention for previous and new tasks. LLMs such as Vision Transformers (ViT) typically have learned rich feature representations on large-scale datasets during the pre-training phase \cite{yuan2021tokens,zhai2022scaling}. In NCCL, these pre-trained models are used as a basis for fine-tuning a small number of key parameters or directing the model's attention via prompt \cite{lester2021power,he2024fppl} so that it can recall and utilize the knowledge learned from previous tasks while learning a new task. This strategy not only helps the model remain robust in the face of data distribution shifts \cite{paul2022vision}, but also facilitates knowledge transfer and integration across tasks.

\cite{yu2024personalized} proposes a method using a pre-trained visual transformer ViT combined with the federated multi-granularity prompt to overcome spatial-temporal catastrophic forgetting in personalized federated continual learning. The main idea is to construct a multi-granularity knowledge space by utilizing prompts of different granularities, namely coarse-grained global prompts and fine-grained local prompts, with ViT. It employs ViT as the shared public cognition and trains coarse-grained prompts operating at the input without altering internal parameters to represent temporal-spatial invariant knowledge. 
The local prompts are built upon the frozen global prompts and directly interact with the model's multi-head self-attention layer. They are designed to capture class-wise fine-grained knowledge for personalization and overcoming temporal forgetting. The local prompt selection is based on the global prompt using a similar key-value pair strategy. Moreover, a selective prompt fusion mechanism is designed to aggregate global prompts from different clients on the server side without spatial forgetting. 

Also utilizing ViT and prompt-based approaches, \cite{piaofederated} proposed a dual knowledge transfer method to mitigate forgetting, which refers to spatial transfer across different clients and temporal transfer across time-sequential tasks. Prompts are short, trainable sequences concatenated to the input or attention layers of a neural network, allowing the model to adapt to new tasks with minimal parameter changes. The key to preventing forgetting is the selective aggregation of these prompts based on estimated task correlations. The method proposed by authors PKT-FCL introduces a two-step aggregation process for prompt generation: (1) First-step aggregation involves estimating a dual task correlation matrix that captures the similarity between tasks. This matrix is used to selectively aggregate prompts from the global prompt pool based on their relevance to the current task.
(2) Second-step aggregation refines this process by selecting only the top-k most relevant tasks for each task, reducing the communication overhead while retaining the essence of knowledge transfer.
The dual distillation loss, a cornerstone of the method, ensures that the transferred knowledge is not overwritten while learning new tasks. It adjusts the knowledge distillation process based on the dual-class correlation matrix, which reflects the relevance of classes across tasks.
The effectiveness of this method lies in its ability to selectively transfer knowledge that is most relevant to the current task while avoiding the transmission of unnecessary or potentially interfering information. The experiments conducted on the ImageNet-R and DomainNet datasets demonstrate that PKT-FCL significantly enhances accuracy compared to existing methods, showcasing its effectiveness in overcoming catastrophic forgetting in federated continual learning scenarios.

The authors in \cite{guo2024federated} propose a method called prototype and low-rank adaptation (PLoRA) to effectively mitigate the challenges of catastrophic forgetting and data heterogeneity with low communication costs. PLoRA utilizes LoRA to fine-tune the pre-trained transformer model for sequential tasks. Unlike other transformer-based approaches, PLoRA does not rely on module similarity selection mechanisms and is optimized end-to-end. The LoRA module is inserted in the first five blocks of the pre-trained ViT model, and only the parameters of LoRA are trainable and uploaded to the global server during adaptation to new tasks. To address classifier bias caused by data heterogeneity, PLoRA employs prototype learning and designs a prototype re-weight module. In particular, the authors set a prototype for each class.
The effectiveness of PLoRA is demonstrated through extensive experiments on CIFAR-100 and Tiny-ImageNet. Moreover, PLoRA exhibits strong robustness and superiority under various non-IID settings and degrees of data heterogeneity.

\subsection{Device-Level}
\begin{figure}
    \centering
    \includegraphics[width=\linewidth]{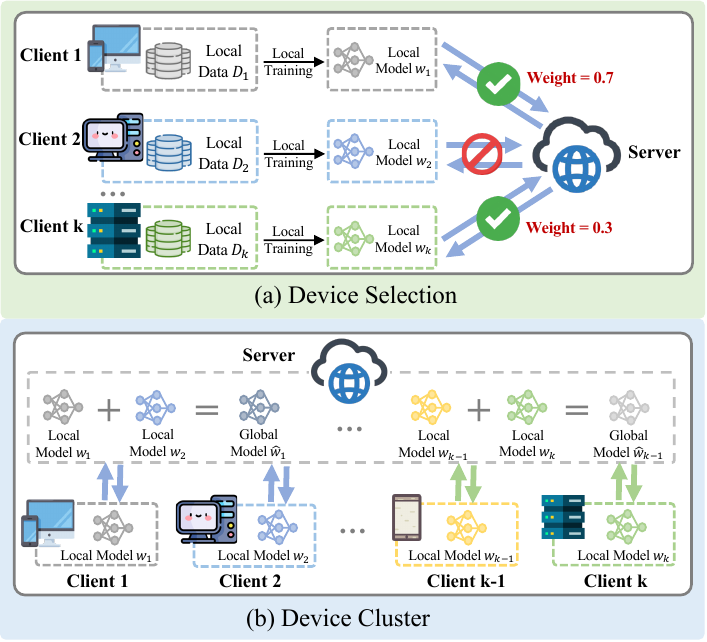}
    \caption{Device-level methods, including device selection and device cluster. Device selection focuses on assigning different weights to participating devices based on the local data, and the device cluster often aggregates the devices that may share similar local data.}
    \label{device-level}
\end{figure}

These methods focus on how to adjust the parameter aggregation strategy based on the data distribution of each device after local training. The device selection approach primarily involves adjusting the weights of different local models within the global model, while the device clustering approach prioritizes aggregating device data with similar data distribution characteristics to mitigate distribution shifts. {We illustrate these methods in Table \ref{Device-Level Methods} and Fig. \ref{device-level}.}
\begin{table*}[h]
\centering
\caption{A Summary of Main Device-Level Methods in NCCL.}
\label{Device-Level Methods}
\renewcommand{\arraystretch}{1.1}
\begin{tabular}{|m{0.05\textwidth}|m{0.45\textwidth}|m{0.20\textwidth}|m{0.20\textwidth}|} 
\hline 
\textbf{Ref} & \textbf{Key Ideas} & {\textbf{Advantages}} & {\textbf{Limitations}} \\ 
    \hline
    \multicolumn{4}{|c|}{\raisebox{-1.8ex}[0pt][0pt]{\centering \textbf{Device-Level: Device Selection}}} \\[3ex]\hline
\cite{li_tpds} & Utilizing a balanced client selection mechanism that prioritizes devices based on cache sample importance. & Addressing heterogeneity to improve performance & Data scarcity issues for adversarial training. \\ 
\hline
\cite{yoon2025pick} & Employing selective device-to-device knowledge transfer to enhance task performance through relevant external knowledge. & Improved accuracy and communication efficiency & Hard to train the model and high training costs.\\
\hline
\multicolumn{4}{|c|}{\raisebox{-1.8ex}[0pt][0pt]{\centering \textbf{Device-Level: Device Cluster}}} \\[3ex]\hline
\cite{jiang2023concept} & Utilizing a clustering-based framework with concept matching for federated continual learning that groups client models into concept clusters and collaboratively trains global concept models. & Alleviating catastrophic forgetting and reducing client interference & Increasing client overhead and dependence on concept estimation. \\ 
\hline
\end{tabular}
\end{table*}

\subsubsection{Device Selection} It refers to the process of adjusting the weight of participating devices in each training round based on their data and system characteristics \cite{fu2023client}. In \cite{li_tpds}, the authors introduce a balanced client selection mechanism to accelerate convergence and improve model performance during federated domain-incremental learning. This mechanism considers the importance of both old and new data by computing an average importance score for each client. 
During each communication round, each client uploads its score to the server. The server then selects a subset of clients for aggregation through a weighted random sampling method, prioritizing clients with higher scores. This balanced client selection mechanism ensures that clients with more essential data contributions are more likely to be selected, thereby accelerating the learning of new domain samples while maintaining the memory of previous domains.

{In \cite{yoon2024pick}, each edge device selectively retrieves external knowledge from a neighboring device based on the similarity of their decision patterns. More specifically, after training on a sequence of tasks, a device computes a decision distance vector for its local model.
The similarity between two models (the local model and a neighboring model) is then measured using the cosine similarity between their decision distance vectors. Based on this similarity, the device selects the neighboring model with the highest score as the backbone for transferring knowledge to its current task. This approach adjusts the contribution of each device’s knowledge by effectively weighting the devices based on the relevance of their learned representations to the target task. As a result, it enables the system to leverage the most beneficial external knowledge while mitigating the risk of negative transfer, leading to improved model accuracy and more efficient learning in heterogeneous federated continual learning environments.}

\subsubsection{Device Cluster} It involves grouping devices based on their model parameters to form multiple centers, each associated with a global model that captures the shared knowledge within that cluster \cite{long2023multi}. In \cite{jiang2023concept}, concept matching (CM) is proposed to tackle the challenges of catastrophic forgetting and interference among clients in federated continual learning. The core idea is to group client models into concept model clusters and build different global models to capture different concepts over time. At each round, the server sends the global concept models to the clients. To avoid catastrophic forgetting, each client selects the concept model that best matches the concept of the current data for further fine-tuning. This selection is done through a client concept matching algorithm, where the client tests the global concept models on its local data and selects the one with the lowest loss.
After receiving the updated client models, the server clusters the models representing the same concept and aggregates them to form cluster models. Since the server does not know the concepts captured by the aggregated cluster models, it employs a novel server concept matching algorithm to update the global concept models. 
The server concept matching algorithm uses a distance-based approach to match cluster models with global concept models. For each cluster model, it finds the global concept model with the smallest distance and updates it only if this distance is smaller than the recorded distance from the previous round. 

\vspace{3pt}
\noindent {\textbf{Conclusion.} In this section, we explore existing approaches addressing two primary challenges in NCCL - catastrophic forgetting and distribution shift - through three key levels: data, model, and device, along with their corresponding techniques. Our analysis reveals that data-level methods have attracted the most research attention. Enhancing models with additional datasets proves to be a relatively effective approach for performance improvement. However, this typically raises privacy leakage concerns and introduces additional training overhead. Model-level methods also demonstrate substantial research activity. While regularization techniques and dynamic network architectures represent extensions of traditional continual learning approaches, recent advancements in LLMs have spurred rapidly emerging research directions that may dominate future studies. Device-level methods remain in their nascent stage, with numerous academic opportunities existing for adjusting client aggregation processes according to streaming tasks in CL scenarios. However, these investigations may pose significant challenges. In practical applications, NCCL itself constitutes an inherently challenging learning paradigm, leading current research to prioritize model performance enhancement over optimizations like lightweight implementations or privacy preservation. Future research directions are recommended as follows: 1) Investigate LLM foundation-based NCCL approaches, focusing on CL optimization methods for LLM base models; 2) Develop context-aware algorithms considering practical constraints such as privacy protection and training environments to improve model applicability; 3) Conduct theoretical analyses of existing NCCL methods to examine convergence properties and interpretability, thereby breaking through current performance limitations.}

\section{Discussion about heterogeneity issues}\label{IV}
\begin{figure}[t]
    \centering
    \includegraphics[width=0.8\linewidth]{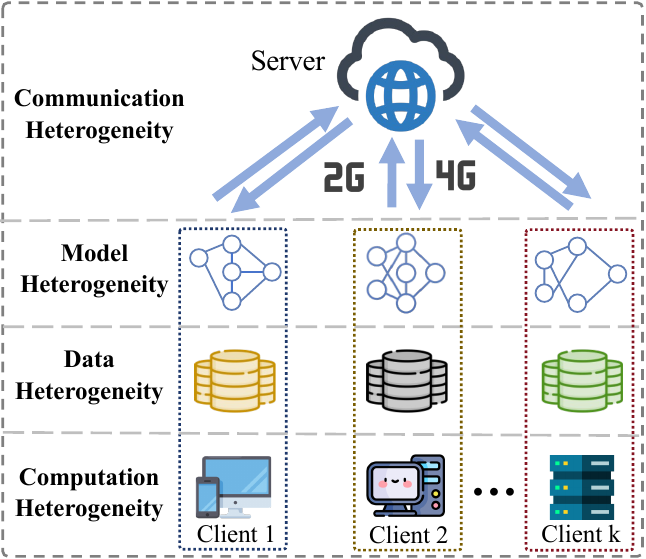}
    \caption{Heterogeneity issues encompass four types: data, model, computation, and communication. Data heterogeneity focuses on the distribution shift among local datasets. Model heterogeneity refers to the different backbone model architectures on various devices. Computation heterogeneity arises primarily from differences in computing resources due to varying hardware environments. Communication heterogeneity pertains to the differing communication conditions when devices communicate knowledge.}
    \label{heterogeneity}
\end{figure}

In the aforementioned section, we have analyzed various techniques to address catastrophic forgetting and distribution drift in NCCL. Then, we will delve into the heterogeneity issue, encompassing four common types of heterogeneity in distributed systems: data heterogeneity, model heterogeneity, computation heterogeneity, and communication heterogeneity. These are also the primary issues that traditional NCL aims to resolve. Furthermore, in NCCL, these challenges may be exacerbated because the CL paradigm introduces more complex tasks and higher hardware requirements for training. {We illustrate four heterogeneity issues in Fig. \ref{heterogeneity} and Table \ref{Heterogeneity methods}.}
\begin{itemize}
    \item \emph{Data Heterogeneity:} It is the most prevalent issue in NCL because the simplistic assumption that data on different devices are independent and identically distributed often does not align with reality. This problem persists in NCCL, and as new tasks arrive, it exacerbates the disparity in data distribution among devices. On the one hand, more research uses real-world datasets to conduct experiments \cite{palazzo2024fedrewind}. On the other hand, considering the value and privacy concerns of real-world data, most research studies employ mathematical algorithms to partition benchmark datasets to simulate different data distributions \cite{casado2020federated}. Furthermore, the data pattern is also a crucial aspect, where the label rate and modality on different devices may vary, which can significantly impact the model training paradigm \cite{song2024systematic}. 
    %
    \item \emph{Model Heterogeneity:} It arises when different devices in the distributed system use varying model architectures or parameters. In NCCL, this complexity is heightened due to the continual learning paradigm, where models need to adapt to new tasks without forgetting previous ones \cite{FCCL_CVPR22,FCCLPlus_TPAMI23}. Addressing model heterogeneity involves developing robust model adaptation techniques to align model parameters across devices while maintaining task-specific knowledge. Ensuring consistent model performance across diverse hardware and software configurations is another challenge that researchers must tackle.
    \item \emph{Computation Heterogeneity:} It refers to the varying computational capabilities among different devices in a distributed system. This type of heterogeneity is particularly relevant in domains such as IoT, airborne systems, and satellite networks. In NCCL, this heterogeneity can lead to uneven workload distribution, where some devices may become bottlenecks due to their limited computational power \cite{luopan2023fedknow,wu2023lifelong}. To mitigate this issue, researchers must design efficient task allocation and load-balancing strategies. Additionally, leveraging hardware acceleration can help alleviate computational bottlenecks. Ensuring that the computational workload is evenly distributed and that all nodes can contribute effectively to the learning process is crucial for the success of NCCL systems.
    \item \emph{Communication Heterogeneity:} It stems from the differing communication capabilities and network latencies among devices in a distributed system. In NCCL, this heterogeneity can significantly impact the synchronization of model updates and the overall efficiency of the learning process. High network latency or bandwidth limitations can lead to stale gradients and inconsistent model updates \cite{gao2024fedprok,shenaj2023asynchronous}. To address this issue, researchers often employ communication-efficient strategies, such as gradient compression and sparse updates, to reduce the amount of data transmitted over the network. Furthermore, developing robust synchronization protocols and leveraging advanced network technologies can help minimize the impact of communication heterogeneity on NCCL systems.
\end{itemize}
\begin{table*}[h]
    \centering
    \caption{A summary of heterogeneity issues in NCCL.}
    \label{Heterogeneity methods}
    \renewcommand{\arraystretch}{1.1}
    \begin{tabular}{|m{0.05\textwidth}|m{0.45\textwidth}|m{0.20\textwidth}|m{0.20\textwidth}|} 
\hline 
\textbf{Ref} & \textbf{Key Ideas} & {\textbf{Advantages}} & {\textbf{Limitations}} \\ 
    \hline
    \multicolumn{4}{|c|}{\raisebox{-1.8ex}[0pt][0pt]{\centering \textbf{Heterogeneity Issue: Data Heterogeneity}}} \\[3ex]\hline
    \cite{guo2021towards} & Improving time-evolving federated learning by approximating past objective functions to handle data heterogeneity. & High convergence rate & Information loss \\
    \hline
    \cite{yu2024personalized} & Utilizing coarse-grained global and fine-grained local prompts to prevent forgetting and achieve effective knowledge personalization. & Efficient knowledge fusion and personalization & Increased computational complexity \\ \hline  
    \cite{ma2022continual} & Mitigating catastrophic forgetting by using client-server distillation and surrogate datasets for task review. & Better utilization of computational resources & Relies on surrogate datasets \\ 
    \hline
    \cite{mei2024using} & Using diffusion models for generating synthetic historical data to mitigate catastrophic forgetting and non-IID challenges. & No need for data storage & Computational overhead \\ 
    \hline
    \cite{zhang1federated} & Utilizing a central server to train a memory generator that creates pseudo data for old tasks, combined with new data to mitigate catastrophic forgetting. & Reduced catastrophic forgetting and privacy preservation & Dependence on synthetic data quality \\ 
    \hline
    \cite{yuan2023peer} & Utilizing a gossip protocol and continual learning paradigm to propagate knowledge among clients for naturalistic driving action recognition. & Resource efficiency & Limited generalization\\
    \hline
    \multicolumn{4}{|c|}{\raisebox{-1.8ex}[0pt][0pt]{\centering \textbf{Heterogeneity Issue: Model Heterogeneity}}} \\[3ex]\hline
    \cite{pennisi2023experience} & Employing experience replay and privacy-preserving generative adversarial networks for decentralized continual learning while maintaining data privacy. & Improved generalization & Increased communication costs\\
    \hline
    \cite{liaoswiss} & A federated multi-task learning framework utilizing tensor trace norm regularization to address data, model, and task heterogeneity, enabling efficient knowledge transfer across clients. & Flexibility in handling heterogeneity in data and model & Computational complexity \\
    \hline
    \cite{FCCL_CVPR22} & Leveraging unlabeled public data to address heterogeneity through cross-correlation learning and alleviate catastrophic forgetting using dual-domain knowledge distillation. & Effective communication across heterogeneous models & Task consistency limitation\\
    \hline
    \cite{FCCLPlus_TPAMI23} & Using federated correlation and similarity learning with non-target distillation to improve both intra-domain discriminability and inter-domain generalization. & Effective cross-domain knowledge transfer & Optimization conflicts in local updating\\
    \hline
    \multicolumn{4}{|c|}{\raisebox{-1.8ex}[0pt][0pt]{\centering \textbf{Heterogeneity Issue: Computational Heterogeneity}}} \\[3ex]\hline \cite{luopan2023fedknow} & Integrating signature task knowledge at the client side to mitigate catastrophic forgetting and negative knowledge transfer.& Efficient local computation & Increased complexity with many tasks\\
    \hline
    \cite{wu2023lifelong} & Optimizing computing resource allocation, balancing age-of-information and energy efficiency in dynamic satellite-edge networks.& Dynamic adaptation to heterogeneous environments & Increased complexity in knowledge management \\
    \hline
    \multicolumn{4}{|c|}{\raisebox{-1.8ex}[0pt][0pt]{\centering \textbf{Heterogeneity Issue: Communication Heterogeneity}}} \\[3ex]\hline \cite{dong2022federated} & Mitigating catastrophic forgetting through class-aware gradient compensation, class-semantic relation distillation, and a proxy server for model selection. & Dual forgetting compensation & Memory overhead \\ 
    \hline 
    \cite{shenaj2023asynchronous} & Using fractal pre-training, prototype-based learning, contrastive loss, and modified aggregation to handle asynchronous and continual learning across clients. & Independent learning schedules & Scalability challenges \\ 
    \hline
    \cite{gao2024fedprok} & Utilizing client-side feature translation and server-side prototypical knowledge fusion for spatial-temporal knowledge transfer. & Efficient communication via prototypical knowledge fusion & Computational complexity \\
    \hline
    \end{tabular}
\end{table*}

\subsection{Data Heterogeneity}
Data heterogeneity (i.e., data with different statistical characteristics) has long been a significant issue in model training. Based on existing research, we will summarize here from the perspectives of data distribution and patterns.

\subsubsection{Distribution} We will examine the data partitioning methods previously used in NCCL research to explore heterogeneous data distribution. Dirichlet Distribution is the most commonly used data partitioning method, especially in image classification tasks, where it shows significant advantages \cite{palazzo2024fedrewind,yoon2025pick}. We assume a distribution over $N$ classes parameterized by a vector $q (q_i \geq 0, \, i \in [1, N] \, \text{and} \, \|q\|_1 = 1)$ , then $q \sim \text{Dir}(\alpha p)$ is drawn from a Dirichlet distribution, where $p$ is a prior class distribution over $N$ classes and $\alpha >0$ is a concentration parameter controlling the identicalness among clients. A smaller parameter $\alpha$ indicates a more skewed distribution. In \cite{guo2021towards}, the authors divided all data sets into 210 subsets for 7 different clients (30 subsets per client) using the Dirichlet distribution. Specifically, they also discussed the issue of overlapping in partitioning. For example, using the CIFAR-100 dataset, with $s$ clients and $m$ subsets per client, when $s=285$ and $m=30$, and the local dataset size is 285, the local datasets do not overlap, i.e., 0\% overlap. Conversely, when $s = 213$, the overlap is 25\%, and when $s = 142$, the overlap is 50\%. Although the overlap degree did not affect the performance of different algorithms in the experiments, the authors believe that the overlap parameter and new arriving data will impact the final performance, and they look forward to resolving these issues in future research.

Based on this, the Dirichlet distribution can be combined with other algorithms. In \cite{shenaj2023asynchronous}, the number of data points per class for each client is computed using a power-law distribution, simulating the imbalance often found in real-world data. For each client, the partitioning is further refined based on a Dirichlet distribution, where the proportions of samples from each class are randomly sampled. The parameter $\alpha$ of the Dirichlet distribution controls the degree of concentration

Similar to \cite{ma2022continual, zhang1federated}, the authors in \cite{mei2024using} sort data by categories and then partition them into shards. The data is first sorted according to the labels and then divided into equal-sized shards, where the number of shards is twice the number of clients. After that, each client is randomly given 2 shards in a non-replicative manner. In the end, each client has data from at most two classes. Furthermore, within each local client, the data is further partitioned into real data and synthetic data generated by the diffusion model. Synthetic data is generated to mitigate the issue of catastrophic forgetting by recovering data distributions.
This partitioning allows the local model to train on a mixture of real and synthetic data, improving its robustness to data distribution shifts.

Unlike other research with data partition algorithms, the authors in \cite{yuan2023peer} utilize two real-world datasets for naturalistic driving action recognition: the State Farm Distracted Driver Detection dataset \cite{statefarm} and the Track3NDAR dataset from the 2023 AICity Challenge. Statistical heterogeneity depends on factors such as the driver’s physical characteristics, behaviors, and postures. Within the same cluster, StateFarm clients' data distribution overlaps more, while AICity clients' data distribution is dispersed. For both datasets, a double-splitting approach was employed to divide the data into training and test sets. First, the clients were partitioned into training and test clients with an 80-20 ratio. Secondly, for each client, the local dataset was randomly split into training and test datasets with the same 80-20 ratio. This method ensured that the test data remained unseen during the iterative training process, allowing for an unbiased evaluation of the model's performance on new data. This data partitioning strategy facilitated evaluating the proposed FedPC framework's ability to generalize to new clients and unseen data, which is crucial for assessing its real-world applicability.

\subsubsection{Pattern}
Apart from differences in data distribution, the heterogeneity of data patterns also poses significant challenges to the NCCL system.
In centralized learning, tasks can be categorized into supervised learning, semi-supervised learning, and unsupervised learning based on the label rate. This issue is inherited and further complicated in NCL and NCCL, as the label rate may vary across each client. According to \cite{song2024systematic}, there are also three different paradigms in federated semi-supervised learning: label-at-all-client-sides, label-at-partial-clients, and label-at-server cases. Therefore, in NCCL, with tasks arriving at different timestamps and exhibiting distinct characteristics, label rates will be even more diverse. In \cite{casado2020federated}, the authors propose a semi-supervised labeling approach to address the issue of partial label data. It utilizes the global model obtained from the cloud to label the unlabeled data in each local device. The confidence of the prediction made by the global model for each unlabeled instance is calculated.
This approach allows the local devices to leverage the global knowledge to enrich their local datasets, thus improving the local models over time. To further enhance the robustness against noisy labels that may arise from the semi-supervised labeling process, the paper employs an ensemble-based method for local learning. Each local device maintains an ensemble of base classifiers, and the final prediction is made based on the median of the posterior probabilities provided by these classifiers. The median rule is chosen due to its robustness against outliers.
This ensemble-based approach helps to mitigate the impact of mislabeled data and improves the overall performance of the local models.

In addition, the global model selection process is designed to filter out local models that perform poorly or provide noisy data. By applying the distributed effective voting technique \cite{chondros2016d}, the server evaluates the significance of each incoming local model based on its performance on a subset of randomly selected local datasets. Only the top-performing models are included in the global ensemble.
This selection mechanism ensures that the global model comprises the most reliable and informative local models, enhancing its robustness against partial and noisy label data.


\subsection{Model Heterogeneity}
Model heterogeneity pertains to the variations in network architectures and learning paradigms that exist across a wide range of devices \cite{liaoswiss,zhong2023semi}. In \cite{pennisi2023experience}, the author introduces the concept of model heterogeneity, suggesting its potential applicability. While the methodology employed in the study was evaluated within the confines of a uniform network architecture, it is noteworthy that the proposed strategy does not necessitate the adoption of an identical model architecture across all nodes. This is particularly relevant given the diverse nature of tasks encountered in real-world applications, where local devices may not necessarily utilize the same model architecture due to varying requirements and constraints. 


{Additionally, to simulate and handle model heterogeneity, the authors in \cite{FCCL_CVPR22} assign different models, including ResNet, EfficientNet, MobileNet, and GoogLeNet, to each domain in the classification tasks. They then introduce the Federated Cross-Correlation Learning method, which computes the logits output from these diverse models. This approach enhances collaboration by maximizing similarity within the same dimensions of the output while minimizing the correlation between different dimensions. This approach enables effective collaboration among participants with varying model architectures using shared public data to learn a generalized representation.}

{In \cite{FCCLPlus_TPAMI23}, the author extends the method proposed in \cite{FCCL_CVPR22} by ensuring that each participant in the federated learning environment employs a distinct model, thus creating a heterogeneous federated environment. The novel method utilizes the embedding features produced by these models to align the instance similarity distribution across participants. This facilitates communication at the feature level, thus improving collaboration while preserving the each unique model.}

\subsection{Computation Heterogeneity}
Computation heterogeneity refers to the differences in computing capabilities among devices. These differences can arise from variations in hardware specifications, processing power, memory capacity, battery life, and other device-specific constraints. In \cite{luopan2023fedknow}, the impact of heterogeneous edge devices is highlighted, specifically extending training time and reducing performance. The experiment included adding 10 CPU-based devices to a cluster with 20 Jetson devices. The CPU-based devices were Raspberry Pi devices with limited computing power, consisting of one with 2 GB memory, five with 4 GB memory, and four with 8 GB memory. In contrast, Jetson devices had strong parallel computing capabilities. The authors propose to address this challenge through its gradient integration method, which optimizes the model update process to be accurate and efficient. The method utilizes a knowledge extractor to retain critical model weights as task knowledge, reducing the computational footprint. During training, the gradient integrator combines the gradient of the current task with gradients of previously learned signature tasks, ensuring that the updated model weights do not degrade performance on previous tasks.
The results showed that training on resource-constrained Raspberry Pi devices delayed the training time of all techniques by an average of 12 times. Moreover, computational heterogeneity reduced the accuracy of all tested algorithms by 3\% to 5\%.

Similarly, in \cite{wu2023lifelong}, the authors study the satellite-airborne-terrestrial edge computing networks, devices such as IoT devices, unmanned aerial vehicles, and satellites, which possess diverse computing capabilities due to their different hardware specifications and deployment environments. This heterogeneity in computing power leads to significant variations in the amount of CPU resources available for data processing. For instance, an IoT device may have a CPU frequency ranging from 315MHz to 916MHz, whereas a satellite can have CPU frequencies as high as 1GHz to 10GHz. This variance in computational capabilities poses challenges in devising efficient resource allocation policies that optimize both the age of information and energy consumption across these heterogeneous devices. To tackle this problem, a lifelong learning algorithm is used to optimize the trade-off between the age of information and energy consumption. It models the problem as a sequence of tasks where each device in the network experiences changes in its operating environment. These tasks are formulated as Markov decision processes. The lifelong learning approach allows the network to learn policies for new tasks more efficiently by reusing knowledge from previously learned tasks. A high-altitude platform acts as a knowledge base, collecting data from devices, performing lifelong learning, and transmitting updated policies back to the devices. Experimental results show that the algorithm significantly accelerates the learning process compared to traditional reinforcement learning algorithms.

\subsection{Communication Heterogeneity}
Communication heterogeneity refers to the differences in communication capabilities and network conditions among devices participating in NCL. These differences can arise from variations in network bandwidth, latency, connectivity stability, and data transfer protocols, impacting how efficiently devices can exchange information. Based on whether the task orders are the same among clients, communication heterogeneity can be categorized into two main types: synchronous and asynchronous communication. 

\subsubsection{Synchronous} In \cite{gao2024fedprok}, the issue of synchronous communication arises when clients are required to proceed through the same sequence of class-incremental tasks in a lockstep manner. This synchronization can lead to inefficiencies and bottlenecks, especially when clients have varying computational capabilities and data availability. The challenge can be formalized as ensuring that all clients complete the task in the same incremental state before proceeding to the next task. This is often hampered by data heterogeneity. To address this synchronous communication issue, the authors propose to incorporate a prototypical knowledge fusion mechanism on the server side. This mechanism allows for spatial knowledge transfer among clients by constructing a global knowledge base that aggregates prototypical information from all participating clients. The fusion process horizontally aggregates heterogeneous prototypical knowledge from different clients and vertically fuses previous and new knowledge along the timeline. By leveraging prototypical knowledge fusion, the method enables clients to proceed asynchronously while still maintaining a global consensus on the learned knowledge. The server periodically distributes the updated global knowledge base to clients, ensuring that each client's local model can incorporate the latest global prototypical information. It's a typical setting implemented in the previous works \cite{dong2022federated}. For instance, after apportioning ten classes across five tasks and assigning them to three clients, when a new task arrives, each client will acquire an additional two classes consistent across clients, yet the samples within these classes are distinct.

\subsubsection{Asynchronous} In \cite{shenaj2023asynchronous}, the authors first formulate the asynchronous federated continual learning scenario; the primary challenge arises from the asynchronous nature of communication among clients. Each client follows its own task stream and learns from new classes at its own pace, leading to out-of-sync local data streams. This asynchrony can cause catastrophic forgetting, where the global model loses knowledge of previously learned classes as new classes are introduced asynchronously across clients. This is represented by the class set for each client at each round, which can differ significantly from client to client. 

To tackle this asynchronous communication issue, the FedSpace framework employs a prototype aggregation mechanism. At each communication round, each client computes the prototype and radius for each class in its current task set. These prototypes are then aggregated on the server side into global prototypes, which capture common knowledge across clients. By aggregating prototypes instead of raw data, FedSpace preserves privacy while enabling knowledge transfer across asynchronously updating clients. Furthermore, FedSpace introduces a contrastive representation loss to align the old aggregated prototypes with the new locally learned representations. This loss function encourages feature vectors of the same class to be close together while pushing apart feature vectors of different classes. This helps mitigate catastrophic forgetting by ensuring the new local representations remain consistent with the global knowledge base, even as clients learn new classes asynchronously.

\vspace{3pt}
\noindent {\textbf{Conclusion.} In this section, we analyze four common heterogeneity issues in distributed learning: data, model, computation, and communication heterogeneity. As an inevitable challenge in practical distributed learning systems, addressing heterogeneity can enhance algorithm robustness. Overall, research on heterogeneity in NCCL predominantly focuses on data heterogeneity, while studies on other heterogeneity issues remain relatively scarce, mostly confined to preliminary explorations without in-depth investigation of their unique impacts. In contrast, NCL has achieved more mature investigations across all four heterogeneity issues. Integrating these insights with CL, future research directions could prioritize: 1) Communication heterogeneity under sequential task arrival scenarios: The asynchronous arrival of new tasks across clients and the incompatibility of direct model aggregation require systematic solutions; 2) Computation heterogeneity for synchronization efficiency: Addressing varying task update speeds across clients through computation-aware methods could improve model robustness in NCCL; 3) Model heterogeneity for practical deployment: Developing algorithms compatible with diverse model architectures (driven by policy or environmental constraints) could facilitate NCCL implementation across heterogeneous devices.}

\section{Discussion about security and privacy}\label{V}
In the previous section, we focus on analyzing the performance of the NCCL method across different techniques and heterogeneity issues. Given the nature of NCCL, where learning occurs on distributed devices, it is essential to address the security and privacy issues during knowledge transmission. In this section, we explore the critical aspects of privacy and security within the framework of NCCL. Specifically, we will analyze several types of malicious attacks and the defense methods that can be employed to safeguard the integrity of local private data{, which is shown in Table \ref{aspect of security and privacy}}.
\begin{itemize}
    %
    \item \emph{Malicious Attack:} Several malicious attacks in NCCL systems will be introduced that aim to disrupt the learning process or extract valuable information. We will detail data poisoning attacks \cite{trinh2023tailoring}, where adversaries introduce harmful data into the training set, and backdoor attacks \cite{wang2022towards}, which involve embedding a hidden, manipulatable feature within the model.
    \item \emph{Defense Method:} To mitigate the threats posed by privacy leakage and malicious attacks, it is essential to explore the defensive strategies implemented in NCCL. We will discuss differential privacy techniques that add controlled noise to protect individual data points \cite{chathoth2022differentially}, the use of order-preserving encryption to maintain data order after encryption \cite{han2022lightweight}, and the incorporation of anomaly detection \cite{chetouane2025new}, adversarial training \cite{chen2024multi}, and shadow learning to prevent potential attacks \cite{wang2022towards}.
\end{itemize}

\begin{table*}[h]
\centering
\caption{A Summary of Security and Privacy Issues in NCCL.}
\label{aspect of security and privacy}
\renewcommand{\arraystretch}{1.1}
\begin{tabular}{|m{0.05\textwidth}|m{0.45\textwidth}|m{0.2\textwidth}|m{0.2\textwidth}|}
\hline 
    \textbf{Ref} & \textbf{Key Ideas}  & \textbf{Advantages} & \textbf{Limitations}\\
    \hline
    \multicolumn{4}{|c|}{\raisebox{-1.8ex}[0pt][0pt]{\centering \textbf{Security and Privacy Issue: Malicious Attack}}} \\[3ex]\hline
    \cite{trinh2023tailoring} & Utilizing Gaussian noise and label flipping methods to analyze data poisoning attacks in federated continual learning. & Gaussian noise reveals basic attack impact. & Label flipping struggles with task-specificity in FCL. \\ 
    \hline 
    \cite{liu2020reflection} & Investigating tailoring byzantine attacks and a novel incremental forgetting attack to disrupt federated continual learning models. & Introduces novel attacks tailored to NCCL vulnerabilities. & Effectiveness varies with defense mechanisms and settings. \\  
    \hline 
    \cite{wang2022towards} & Discussing backdoor attacks in federated continual learning, along with pixel attack and anomaly detection evasion approaches. & Exploits backdoor leakage to degrade model robustness. & Requires sustained malicious influence over training rounds. \\  
    \hline 
    \multicolumn{4}{|c|}{\raisebox{-1.8ex}[0pt][0pt]{\centering \textbf{Security and Privacy Issue: Defense Method}}} \\[3ex]\hline
    \cite{chathoth2022differentially} & Using cohort-based differential privacy to ensure varying privacy requirements and data distributions among different clients. & Introduces cohort-level DP to protect heterogeneous privacy budgets. & Relies on cohort structure and may degrade with dynamic data. \\  
    \hline
    \cite{xiao2023privacy} & Using Bayesian differential privacy to provide more balanced privacy protection for different streaming tasks. & Dual-model structure and Bayesian differential privacy enhance accuracy and privacy. & Model compression and noise addition may introduce computational overhead. \\  
    \hline
    \cite{zizzo2022federated} & Calculating differentially private means of local data per class for each device. & DP noise enables shared replay buffers to enhance performance. & High client participation reduces gains from the global buffer. \\  
    \hline
    \cite{han2022lightweight} & Utilizing a practical order-preserving encryption mechanism designed for vertical federated continual learning. & Efficient privacy protection with OPE maintains model accuracy. & Potential distribution leakage, overheads with sparse data.\\  
    \hline
    \cite{talpur2022gfcl} & Introducing a GRU-based anomaly detection framework to counteract data poisoning attacks in IoV environments. & GRU-based anomaly detection in NCCL for dynamic IoV. & Relying on the delay feature, threshold selection is critical with specific datasets.\\  
    \hline
    \cite{wang2022towards} & Employing a robust defense against backdoor attacks using data filtering, early stopping, and periodic retraining. & Uses shadow learning to filter and early-stop backdoor attacks. & May struggle with high client heterogeneity or dynamic data distributions. \\  
    \hline  
\end{tabular}
\end{table*}

\subsection{Malicious Attack} 
We focus on two typical attack methods (data poisoning and backdoor attack) aimed at disrupting the learning process or stealing valuable information. 

\subsubsection{Data Poisoning} It often uses an adversarial strategy where the attacker introduces malicious data into the training set of the distributed device \cite{chen2017targeted}. The goal of data poisoning is to degrade the performance of the trained model or to manipulate its behavior. The authors in \cite{trinh2023tailoring} analyze data poisoning attacks in NCCL in detail. The Gaussian noise attack, where an adversary introduces noise drawn from a zero-mean Gaussian distribution into the model parameters, is employed.
The effect of this attack is to disrupt the model's ability to learn from the data, leading to increased variance in the model's predictions and a potential decrease in accuracy. Additionally, the label-flipping method is discussed, which involves the alteration of training sample labels to disrupt the learning process. 
The impact of label flipping is to mislead the model, causing it to learn incorrect associations between inputs and outputs, which can significantly impair the model's ability to generalize to unseen data.

\subsubsection{Backdoor Attack} It involves inserting a backdoor into a model during training, which can later be activated to manipulate the model's behavior \cite{liu2020reflection}. In \cite{wang2022towards}, the authors discuss a backdoor attack in NCCL, where an adversary manipulates the training process to embed a hidden behavior into the model. One common approach is the pixel attack, where a malicious client adds a specific pattern or trigger to a subset of training samples and alters their labels to a target label. 
The goal is to ensure that when the model encounters a sample with the trigger, it will classify it as other labels, regardless of the actual content of the sample. Another variant discussed is the backdoor attack with evasion of anomaly detection, where the adversary modifies their objective function to include an anomaly detection term, aiming to evade defenses.

\subsection{Defense Methods}
Existing NCCL research has primarily focused on defense methods to ensure the security of distributed data during transmission. We have identified four relevant studies: differential privacy, encryption, anomaly detection, and shadow learning.

\subsubsection{Differential Privacy} It is characterized by the introduction of controlled noise into data or its derivatives, which guarantees that the addition or removal of a single record from a dataset will not significantly affect the prediction \cite{dwork2006differential}. 


Similar to \cite{chathoth2021federated}, authors in \cite{chathoth2022differentially} introduce a novel approach to address privacy challenges and heterogeneous privacy budgets in federated learning. The method leverages differential privacy to ensure strong privacy guarantees while allowing clients grouped into cohorts to define their own privacy requirements. To train a differentially private model, the paper adapts the DP-stochastic gradient descent algorithm to incorporate heterogeneous privacy budgets. At each communication round, the algorithm tracks the privacy loss for each cohort using a cohort-based privacy accountant. Additionally, it introduces two novel continual learning-based DP training methods, DP-synaptic intelligence and DP-rehearsal, to mitigate the effects of forgetting previously learned experiences when the privacy budget of a cohort is spent. These methods regularize the learning process, reducing the performance drop in stricter privacy cohorts.

{The method proposed in \cite{chen2024evaluating} employs the Laplace mechanism to add noise to local model updates, ensuring that individual contributions from edge devices remain obscured. This approach not only protects user data privacy but also maintains model performance by balancing the trade-off between privacy and utility. To incorporate DP, each device adds Laplace noise to its local gradient updates.
The server then aggregates these noise-added updates to form the global model.
This method ensures that the global model retains the ability to adapt to new tasks while preserving previously learned knowledge, thereby effectively mitigating catastrophic forgetting. Extensive experiments on the MNIST dataset demonstrate that the proposed method achieves a favorable trade-off between privacy protection and model accuracy, particularly under non-IID data distributions.}

Furthermore, \cite{chen2022differential} introduces a blockchain-based differential optimization federated incremental learning algorithm to enhance privacy preservation in NCCL. The algorithm employs differential privacy techniques, specifically the addition of Laplace noise, to safeguard sensitive data during the model training phase. The core idea is to ensure that the model parameters do not leak sensitive information, which is achieved by adding noise to the model updates. The Laplace mechanism is used to add noise to the model's output.
This approach helps in reducing the impact on the model's accuracy due to the addition of differential privacy. The paper discusses optimizing parameters within a weighted random forest to mitigate the accuracy loss associated with differential privacy. The algorithm also incorporates an ensemble learning technique to integrate local models, improving the global model's accuracy. The model parameters are uploaded to the blockchain, ensuring secure and efficient parameter management. This method not only bolsters the model's robustness against adversarial attacks but also leverages blockchain technology to ensure data integrity and non-repudiation.

\subsubsection{Order-Preserving Encryption} It is a cryptographic technique that allows encrypted data to be sorted or compared while maintaining the original data order \cite{agrawal2004order}. The core property of OPE is that for any two values $i, j$ in a range $R$, if $i < j$, then the encrypted values $f(i) < f(j)$, where $f$ is the OPE function. This property is crucial for decision trees, which compare feature values to make splitting decisions.

In \cite{han2022lightweight}, order-preserving encryption (OPE) is employed to protect data privacy while enabling the training of decision trees in federated incremental learning. Traditional OPE algorithms may provide weaker security than common encryption methods, as they preserve the size relationships of the ciphertext. To mitigate this, a practically order-preserving encoding mechanism is proposed to add random noise to the encrypted data to enhance privacy. 
This noise helps to achieve differential privacy, which is a formal framework for protecting the privacy of individual data points in a dataset. To further improve the efficiency of the solution and reduce storage overhead, a regional counting method is introduced. This method divides the values of a feature into different groups and uses the group mean to denote the feature value of the group. The encryption function is then applied to these group means rather than to individual data points. This strategy not only maintains model accuracy but also enhances privacy by reducing the granularity of the data that is exposed to potential attackers. The use of OPE in regional counting allows for the training of federated incremental decision trees with strong privacy guarantees while maintaining high efficiency.

\subsubsection{Anomaly Detection} It is defined as the process of identifying clients or data points that significantly deviate from the expected behavior or model updates, which may indicate malicious activities or data corruption \cite{chandola2009anomaly,nogami2025federated}. 

For anomaly detection in Kubernetes environments, the holistic security and privacy framework \cite{tomas2024novel} employs NCCL with unsupervised machine learning, notably auto-encoders, to detect anomalies directly from raw network traffic, sidestepping the need for labeled data and preserving privacy. This framework comprises three main components: the collector, which gathers network traffic data; the agent, responsible for local analysis and model training; and the aggregator, acting as the central server to coordinate federated training and model updates. This architecture enables the framework to process and classify the network traffic in real-time, leveraging unsupervised learning to refine models based on unlabeled data and to adapt to new patterns without forgetting previously learned knowledge. The evaluation demonstrates its effectiveness in a micro-services-oriented application environment, where it was tested against various types of attacks, including denial of service, port scan, brute force, and SQL injection \cite{halfond2006classification}.

To detect accounting anomalies in financial auditing, \cite{schreyer2022federated} is designed to address the challenge of learning highly adaptive audit models in decentralized and dynamic settings, where data distribution shifts over multiple clients and periods are common. The anomaly detection is performed using auto-encoder networks, which are trained to learn a comprehensive model of the data distribution. This architecture consists of an encoder and a decoder. The encoder learns a representation of a given input sample, and the decoder learns a reconstruction of the original input. The model is optimized to minimize the reconstruction error.
In the audit setting, the reconstruction error magnitude is interpreted as the deviation from regular posting patterns, and journal entries with high reconstruction errors are selected for detailed procedures.

\subsubsection{Shadow Learning} It provides a robust defense against backdoor attacks by integrating data filtering, early stopping, and periodic retraining of the shadow model \cite{beane2019shadow}. Two models will be trained in parallel: a backbone model is trained on all client data and aims to achieve high performance on the main task without specific defense mechanisms while a shadow model is central to the defense approach. The authors in \cite{wang2022towards} further optimize shadow learning for federated continual learning. The backbone model undergoes continual training with data contributions from all clients, unburdened by any defensive mechanisms, thereby focusing solely on executing the primary classification task. However, given the omnipresent risk of backdoor attacks, where adversarial clients may introduce malicious training data, the backbone model may gradually become compromised. To counteract this peril, shadow Learning introduces a filtering mechanism rooted in robust covariance estimation and data whitening \cite{kocher2020spectre}.

The filtering process hinges on computing the quantum entropy score of sample representations, a metric derived from the spectral properties of the data. This score serves as a discriminant tool, amplifying the spectral signature of corrupted data, thus facilitating the identification and removal of malicious updates.
After the filtering process, the resultant clean updates are employed to nurture the shadow model. This model undergoes training until an early stopping criterion is met, thereby averting overfitting to potential backdoor triggers. The shadow model, thus, functions as a pristine version of the backbone model, specifically tasked with detecting and accurately classifying targets potentially tainted by backdoor attacks. During the inference phase, input samples are initially processed by the backbone model. If the predicted label falls within a predefined set of suspicious target labels (determined through the filtering phase), the sample is subsequently passed through the shadow model for definitive classification. This dual-model approach ensures the system maintains high accuracy on the primary task while effectively fortifying its defenses against backdoor attacks.

\vspace{3pt}
\noindent {\textbf{Conclusion.} This section examines security and privacy concerns through the lens of malicious attacks and defense methods, along with their associated techniques. Current research predominantly focuses on defense strategies, while investigations into attack methodologies remain limited, primarily combining conventional NCL attack approaches with continual learning scenarios. Future studies should prioritize exploring unique challenges inherent to the NCCL paradigm, such as inter-task correlations within streaming tasks and their implications for designing attack-resilient defense algorithms.}


\section{Discussion about real-world applications}\label{VI}
In the preceding sections of this survey, we have discussed the foundational issues and methodologies of NCCL in enabling collaborative continual learning among distributed devices. This emerging paradigm enables distributed systems to adapt continuously to new data while preserving previously acquired knowledge with privacy preservation. In this section, we turn our focus to practical applications of how NCCL can be effectively utilized in the real world to solve downstream tasks and cope with resource constraints.

\subsection{Downstream Tasks}
\begin{figure}
\centering
\includegraphics[width=\linewidth]{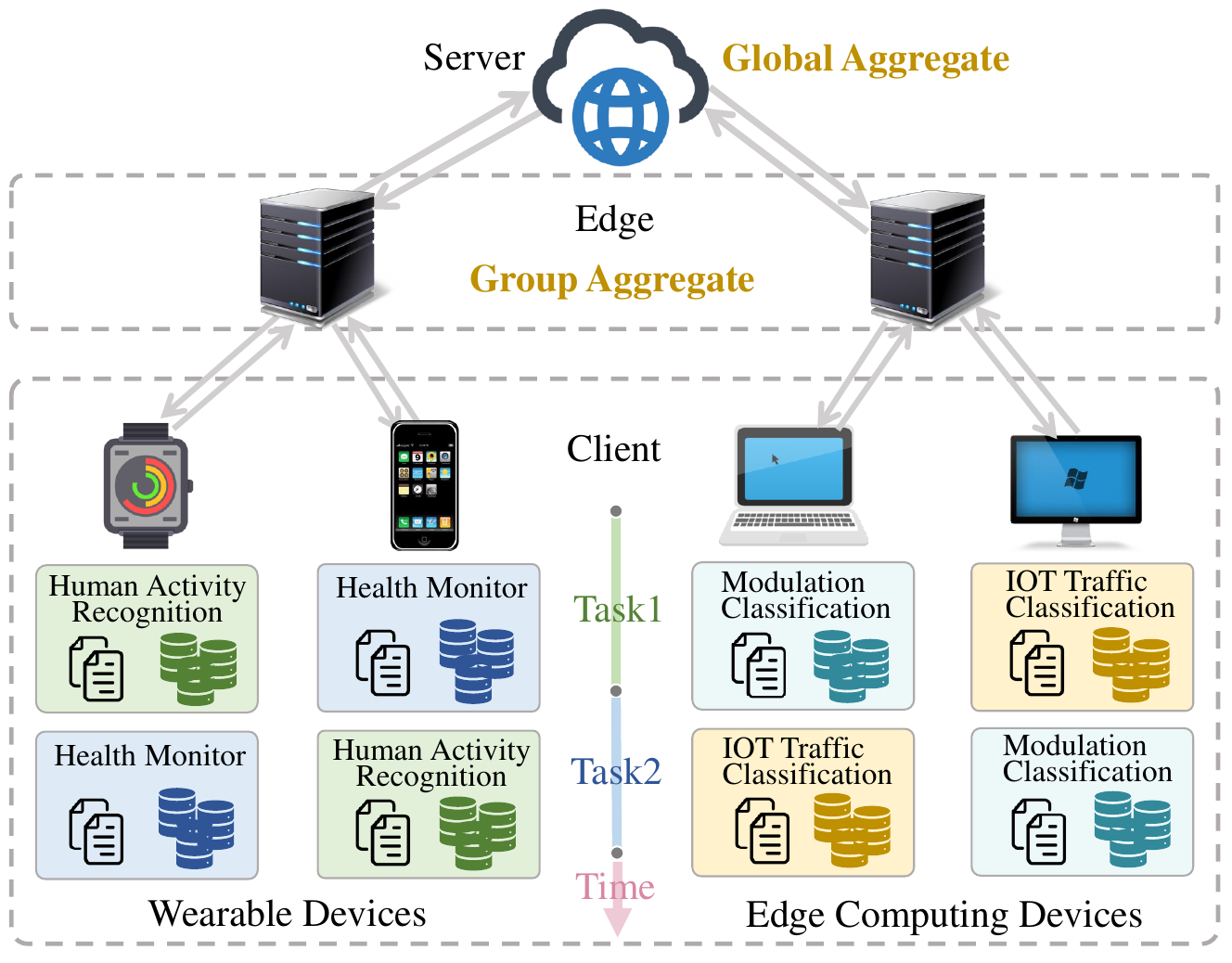}
\caption{Hierarchical learning scenario with continual learning of IoT devices scenario. The IoT devices are divided into multiple groups to continuously learn new tasks, and each group will rely on the edge aggregation model at the side end and, ultimately, on the server aggregation on the cloud.}
\label{figureiot}
\end{figure}

\begin{figure}
\centering
\includegraphics[width=\linewidth]{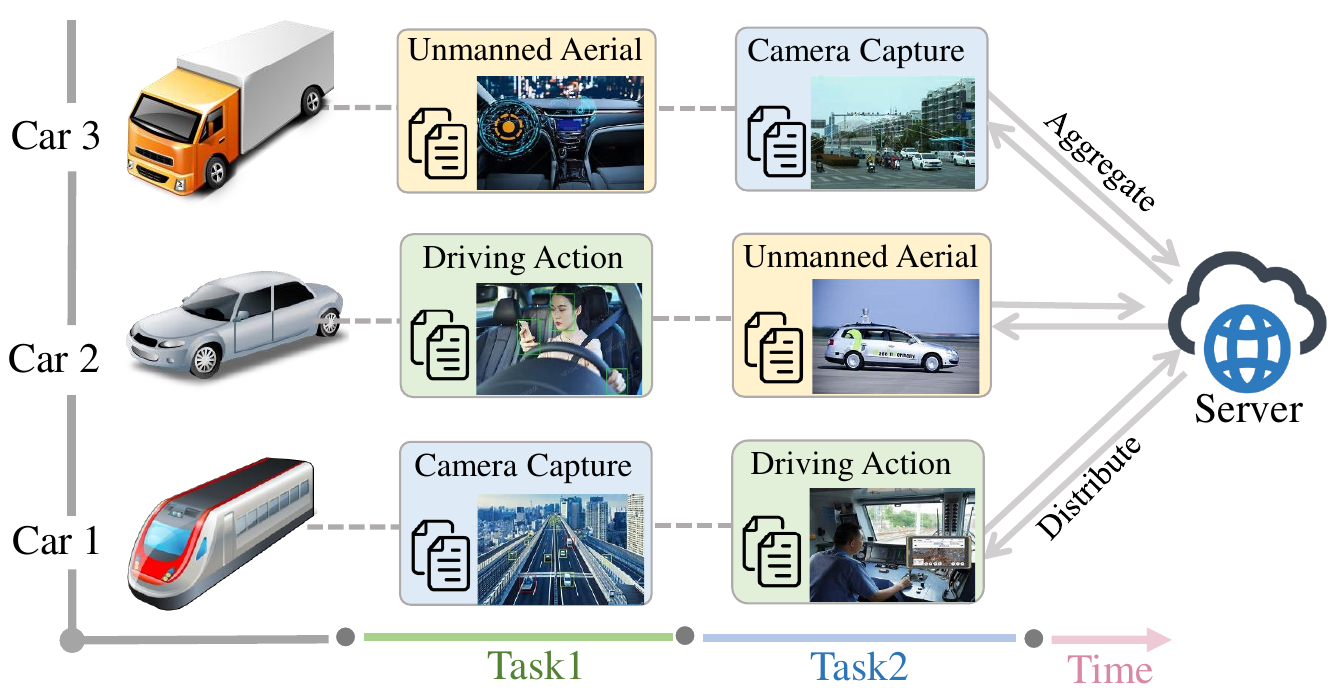}
\caption{Federated learning for transportation systems with the central server. Each transportation system end client needs to send the model to the server for aggregation after learning the task flow.}
\label{figuretransport}
\end{figure}

\begin{figure}
\centering
\includegraphics[width=0.5\textwidth]{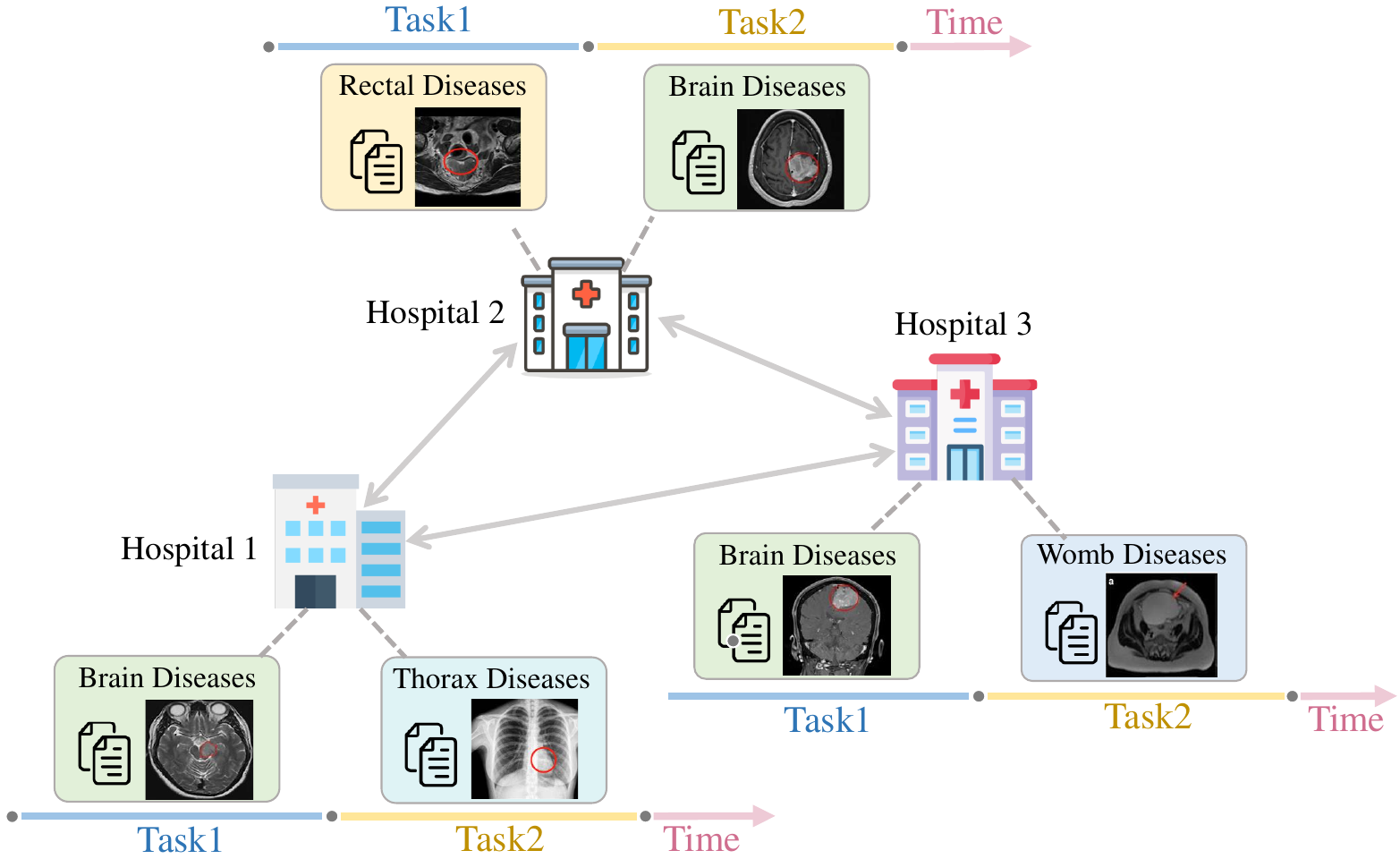}
\caption{Decentralized learning for multiple healthcare providers to jointly learn from streaming medical image analysis tasks, where each client exchanges learned knowledge peer-to-peer without the need for a centralized server to help aggregate the model.}
\label{figuremedical}
\end{figure}

The growing complexity and diversity of tasks in real-world applications \cite{wang2024federated} highlight the importance of NCCL. Despite extensive academic research conducted on NCCL, practical applications require greater attention to challenges arising from privacy concerns, communication overhead, and scalability issues, especially amidst rapid data growth and real-time demands from various devices. This section focuses on four key application areas where NCCL can have a substantial impact, summarized in Table \ref{tableapplications1} and \ref{tableapplications2}.
\begin{itemize}
    \item \emph{Internet of Things (IoT):} In the era of pervasive computing, the Internet of Things has become a critical infrastructure, with countless IoT devices continuously generating and processing data, leading to an explosive growth of distributed data \cite{kumar2020trustworthy,guerdan2023federated}. NCCL enables IoT devices to continuously learn and adapt to new tasks or environments in a collaborative manner without the need for data centralization, thereby greatly enhancing the intelligence and autonomy of IoT systems while reducing privacy risks and communication costs \cite{chen2025knowledge,zhou2024drift}. We depict the application of IoT devices in Fig. \ref{figureiot} in terms of hierarchical learning with the client-edge-cloud model.
    \item \emph{Intelligent Transportation Systems (ITS):} Modern transportation systems generate massive amounts of data from various sources, such as vehicles, roadside units, and traffic cameras \cite{guerrero2018sensor}. The unique challenge lies in the real-time processing system while ensuring data privacy and system scalability. NCCL can be utilized to develop intelligent transportation systems that continuously learn and adapt to changing traffic conditions, user behaviors, and environmental factors while keeping sensitive data local \cite{reddy2024deep,barbieri2022decentralized}. This can lead to more efficient and sustainable ITS. Fig. \ref{figuretransport} depicts the scenario of transportation systems with federated learning.
    \item \emph{Medical Diagnosis:} Medical diagnosis technologies, such as X-ray, CT, and MRI, generate a substantial volume of critical data in disease diagnosis and treatment \cite{kapoor2024fedcl,hussain2022modern}, requiring careful handling to protect patient privacy and data integrity. NCCL enables distributed medical institutions to collaboratively learn and improve diagnostic models without sharing raw patient data \cite{guo2022federated,sun2025federated}. This can accelerate the development of more accurate and robust medical imaging AI while safeguarding patient privacy. Fig. \ref{figuremedical} illustrates the application scenario of medical images in terms of decentralized learning without a central server.
    \item {\emph{Network Application:} Communication technologies such as 5G/6G open radio access networks and digital twin infrastructures generate vast amounts of critical data for network management and anomaly detection. NCCL enables distributed network operators to collaboratively learn and optimize diagnostic and anomaly detection models without sharing raw data \cite{xia2024fcla,he2023federated}. This facilitates the rapid development of more accurate and robust artificial intelligence systems for both 5G network anomaly detection and digital twin applications while simultaneously safeguarding data integrity and privacy.}
\end{itemize}

\begin{table*}[h]
\centering
\caption{NCCL for Real-World Applications (1/2).}
\label{tableapplications1}
\renewcommand{\arraystretch}{1.1}
\begin{tabular}{|m{0.05\textwidth}|m{0.45\textwidth}|m{0.2\textwidth}|m{0.2\textwidth}|}
\hline 
    \textbf{Ref} & \textbf{Key Ideas}  & \textbf{Advantages} & \textbf{Limitations}\\ \hline
    \multicolumn{4}{|c|}{\raisebox{-1.8ex}[0pt][0pt]{\centering \textbf{Application: Internet of Things (IoT)}}} \\[3ex]\hline
    \cite{Haipeng}  & The factory participates in the federated sub-terminal selection with dynamic local data. & Data sharing in a privacy-preserving manner. & Complexity in parameter depth management. \\ \hline
    \cite{liu2019lifelong} & Enhancing robot navigation within cloud robotics through the application of knowledge transfer within federated continual learning. & Efficient knowledge fusion with different robots. & High computational overhead. \\ \hline
    \cite{yu2022towards} & Implementing federated learning with continuous sim-to-real transfer for vision-based obstacle avoidance in IoT robotic systems. & Enhanced model generalization in obstacle avoidance. & Mismatching between simulation and reality. \\ \hline
    \cite{jin2024fl} &  Utilizing class gradient balance and label smoothing loss in conjunction with federated continual learning for dynamic IoT intrusion detection. & Dynamic memory and balanced loss design.  & Limited non-IID adaptability. \\ \hline
    \cite{jin2023federated} &  Employing a discriminative auto-encoder and federated continual learning for incremental intrusion detection in IoT networks.  & Reduced communication overhead. & Dependence on partial historical data.\\ \hline
    \cite{tomas2024novel} &  Achieving decentralized anomaly detection through continual updates leveraging unsupervised machine learning. & Efficient continual model adaptation. & Inconsistent performance across different scenarios. \\ \hline
    \cite{putra2024collaborative} &  Enhancing decentralized learning with iterative model aggregation for continuous deepfake video detection in the entertainment industry. & Resilience against single points of failure. & Slightly lower accuracy with increased devices. \\ \hline
    \cite{lozano2024continual} &  Utilizing satellites as agents in a deep reinforcement learning framework, with federated continual learning enabling adaptive routing in satellite networks. & Adaptability to dynamic network environments. & Complexity in model synchronization across satellites. \\ \hline
    \cite{osifeko2021surveilnet} &  SurveilNet leverages federated learning for collaborative anomaly detection in surveillance systems with streaming data while ensuring data privacy. & Efficient privacy preservation. & Limited scalability under large-scale deployments. \\ \hline
    \cite{alkouz2023failure} & A failure-aware framework for drone delivery incorporates weighted FL for predictive maintenance and a speed-heuristic approach for real-time service.  & Enhanced failure prediction with FL. & Increased computational overhead in large-scale deployments. \\ \hline
    \cite{zhang2023incremental} & Addressing concept drift and privacy in photovoltaic power forecasting through energy-distance-aided drift detection and incremental federated broad learning. & Efficient concept drift detection. & Potential high communication overhead. \\ \hline
    \cite{zhang2023online} &  Enhancing fault prediction in power networks through personalized federated learning and incremental SCNs, dynamically integrating data across substations.  & Incremental SCN for dynamic streaming data. & Sensitivity to extreme data heterogeneity.  \\ \hline
    \cite{lv2023blockchain} & The blockchain-enhanced algorithm ensures the security of digital twin networks through decentralized learning and robust privacy protection.  & Blockchain-enhanced data security and privacy. & High communication and storage overhead. \\ \hline
    \cite{schreyer2022federated} & Continual learning facilitates the detection of accounting anomalies in financial audits by collaboratively updating models across decentralized clients.  & Privacy-preserving federated continual learning framework. & Fail with internal adversarial attacks. \\ \hline
    \multicolumn{4}{|c|}{\raisebox{-1.8ex}[0pt][0pt]{\centering \textbf{Application: Intelligent Transportation System (ITS)}}} \\[3ex]\hline
    \cite{math11081867} & Federated incremental learning integrated with meta-learning and batch-dependent temporal factors for advanced driver distraction detection in 3D space. & Efficient communication cost reduction via gradient filtering. & High computational overhead on client devices. \\  \hline
    \cite{lanza2023urban} &  Point-to-point federated continual learning for accurate traffic condition prediction.  & Enhanced privacy protection through decentralized data processing. & Limited model generalization under heterogeneous sensor data. \\ \hline
    \cite{lei2022federated} & A vehicle diversity selection algorithm that leverages multiple performance indicators for enhanced intelligent connected vehicles systems. & Improved performance while preserving privacy. & Dependence on stable network coordination. \\ \hline
    \cite{reddy2024deep} & A fog-cloud system in FL boosts the accuracy of ITS event detection by processing local vehicle data and sharing global insights among clients.  & Efficient communication via the fog-cloud framework. & Potential communication overhead between fog and cloud. \\ \hline
    \cite{barbieri2022decentralized} & Connected vehicles engage in consensus-driven federated learning to collaboratively refine deep neural networks collaboratively, enabling extended sensing capabilities in 6G environments. & Communication-efficient modular design. & High Dependence on connectivity density. \\  \hline
\end{tabular}
\end{table*}

\subsubsection{IoT} In \cite{Haipeng}, authors propose a federated incremental learning method for data sharing in the industrial IoT. This addresses the common challenges of high new data influx and volume imbalance among factory nodes. This approach integrates new state data seamlessly with existing industry federation models. It begins by distributing an initial local model to plant sub-terminals. To handle dynamic and uneven data distribution, a federated sub-terminal optimization algorithm adjusts participating nodes for fair and efficient training. The core of this is computing incremental weights via a federated incremental learning algorithm. This allows rapid data integration, enhancing system adaptability and responsiveness. The method is tested on the CWRU bearing dataset \cite{neupane2020bearing}, outperforming traditional FedAvg and non-incremental methods in diagnostic accuracy.


Similar to \cite{9821057}, the authors in \cite{9810510} use a federated incremental learning framework combined with knowledge distillation to address the modulation classification challenge in the context of cognitive IoT, aiming to effectively integrate the classification knowledge from the private categories of local devices into the global model. This method regards the private classes of a particular device, which are different from the common classes among all devices, as incremental classes. Incremental learning is adopted during the training of each local model to learn the classification knowledge of these private classes. Moreover, knowledge distillation is utilized in the local model training stage to maintain the classification knowledge from the global model and avoid excessive dispersion between the global model parameters and local model parameters.

In both \cite{9821057} and \cite{9810510}, knowledge distillation is leveraged to adapt tasks effectively. The first study employs models from both the previous client round and the current server as teacher models, encapsulating the necessary knowledge for assessing past tasks. By calculating the distillation loss between these teacher models and the current client's model, the method retains prior knowledge. The second study treats each client's unique class data as incremental, using knowledge distillation during local training to preserve global model classification knowledge. Here, local models are trained to excel on their tasks and mimic the global model's behavior. This dual objective minimizes a combined loss function of local data loss and distillation loss. Both methods integrate classification and distillation losses to mitigate forgetting.

In real-world IoT traffic classification, the authors in \cite{zhu2022attention} introduced Fed-SOINN, a novel attention-based federated incremental learning algorithm. By integrating incremental and federated learning, Fed-SOINN collaboratively trains a comprehensive network classification model while preserving data privacy in non-IID IoT environments. Based on the SOINN algorithm and radial basis function, Fed-SOINN's online traffic classification model adapts to time-varying data. This semi-supervised approach identifies unknown traffic categories and utilizes an attention mechanism to enhance the weights of beneficial client model parameters. By sharing only important gradients, Fed-SOINN reduces communication rounds. Evaluated on Moore \cite{moore2005internet} and ISCX-VPN datasets \cite{wang2017end}, Fed-SOINN improved detection accuracy by 3.1\% and reduced communication rounds by up to 73\%. Additionally, its incremental learning mechanism effectively identifies new traffic categories, ensuring stable overall performance.

In \cite{9973580}, the authors aim to address data freshness in IoT and mobile edge computing. It prioritizes newer data batches through a loss compensation mechanism, ensuring they have a more substantial influence on model updates. Suitable for continuous data flows from IoT devices, the method adjusts the loss function based on data batch arrival times, enhancing the model's responsiveness to recent data pattern changes. Implementing a batch-oriented aggregation method updates federated models more frequently and at a finer granularity, aligning with the dynamic nature of IoT data. Compared to traditional approaches, the method significantly improves convergence speed and learning performance.

\subsubsection{Intelligent Transportation Systems} In \cite{math11081867}, the authors propose ICMFed, an incremental and cost-efficient federated meta-learning mechanism for real-world 3D driver distraction detection in transportation systems. ICMFed tackles four key challenges: data accumulation, communication optimization, data heterogeneity, and device heterogeneity. By integrating incremental learning, meta-learning, and federated learning, ICMFed introduces a novel learning paradigm for 3D tasks. It stabilizes local model training with a temporal factor associated with local batches and retains accumulated knowledge. A gradient filter optimizes client-server interaction and reduces communication costs. Additionally, a normalized weight vector enhances global model aggregation, considering the richness of local updates. ICMFed supports rapid user personalization by adapting the global meta-model to specific contexts. {The framework was validated using the State-Farm-Distracted-Driver-Detection dataset, which contains 22,424 labeled samples from 26 drivers. ICMFed addresses real-world challenges in detecting unsafe behaviors like unfocused eyesight and inattention, achieving 96.86\% quality improvement while protecting user privacy through federated updates.}

In \cite{lanza2023urban}, the authors introduce federated point-to-point continuous learning, a federated peer-to-peer strategy for continuously learning from traffic intensity sensor data in urban areas. It aims to collaboratively develop a robust model capable of predicting traffic conditions in non-static urban environments. Leveraging edge computing, this paradigm processes data locally on IoT devices like traffic sensors. Its peer-to-peer architecture distributes the learning process across devices, fostering incremental model growth through collective contributions. As new sensor data is acquired, this method dynamically updates the traffic model, ensuring accuracy and relevance without needing full retraining. The framework was tested on real traffic data from Madrid's Salamanca district, processing 127 sensors' 15-minute granularity traffic flow measurements. The method successfully predicted short-term traffic patterns while reducing energy consumption by 87\%, demonstrating practical value for dynamic urban mobility management.

The authors in \cite{lei2022federated} proposed a federated learning framework based on incremental weighting and diversity selection to address the challenges of isolated data islands, incremental data, and data diversity. This approach integrates incremental learning and federated learning to provide a novel learning paradigm. The method introduces a vehicle diversity selection algorithm that uses various performance indicators, such as accuracy and loss, to calculate diversity scores. The algorithm effectively reduces homogeneous computing and maintains independent complementary data diversity by employing an improved cosine distance measure with a penalty factor.
Moreover, this method proposes a vehicle federated incremental learning algorithm that uses an improved arc tangent curve as the decay function to realize the rapid fusion of incremental data with existing machine learning models. The incremental parameter depth value is calculated based on the proportion of incremental data and an adjustment variable.
%
{Applied to IoV scenarios with 353 million smart vehicles generating 4000 GB/day, the framework demonstrated 32\% accuracy improvement on vehicle image datasets while handling heterogeneous road conditions (urban vs. suburban, weather variations). This enables real-time adaptation for driving assistance systems without centralized data collection.}

\begin{table*}[h]
\centering
\caption{NCCL for Real-World Applications (2/2).}
\label{tableapplications2}
\renewcommand{\arraystretch}{1.1}
\begin{tabular}{|m{0.05\textwidth}|m{0.45\textwidth}|m{0.2\textwidth}|m{0.2\textwidth}|}
\hline 
    \textbf{Ref} & \textbf{Key Ideas}  & \textbf{Advantages} & \textbf{Limitations}\\ \hline
    \multicolumn{4}{|c|}{\raisebox{-1.8ex}[0pt][0pt]{\centering \textbf{Application: Medical Diagnosis}}} \\[3ex]\hline
    \cite{zheng2023asynchronous} & A replay technique leveraging shared experience replay buffers is employed to refine landmark localization in medical imagery. & Robust decentralized architecture. & High on-device computational load. \\ \hline
    \cite{dong2022federated} & A Forgetting-balanced learning model addresses the challenge of federated incremental semantic segmentation in the context of medical images. & Enhanced performance while privacy-preserving. & High computational overhead. \\ \hline
    \cite{wang2023peer} & A peer-to-peer federated continual learning strategy is introduced to enhance the imaging quality of low-dose CT scans.& Decentralized privacy preservation. & Complex communication overhead. \\ \hline
    \cite{zhang2024intelligent} & Federated continual learning, integrated with the Swin Transformer, is utilized for diagnosing plant leaf diseases.  & High accuracy with robust data augmentation. & Limited real-world generalization. \\ \hline
    \cite{kim2024continual} & A decentralized continual learning framework, leveraging GAN-based synthetic data, is employed for multi-center ECG studies to evaluate arrhythmia detection methods. & Privacy-preserving fake data evaluation. & High computational cost for GAN training. \\ \hline
    \cite{sun2023federated} &  MetaCL harnesses federated continual learning and blockchain technology for secure and adaptable physiological signal classification in the IoT. & Ensuring privacy in federated blockchain integration. & Residual blockchain poses privacy risks. \\ \hline
    \cite{guo2022federated} & Incremental exemplar-based learning is proposed to manage real-time distributed medical data in computer-aided diagnosis for smart healthcare applications. & Efficient resource in federated-exemplar fusion. & Scalability challenges in exemplar management. \\  \hline
    %
    \multicolumn{4}{|c|}{\raisebox{-1.8ex}[0pt][0pt]{\centering \textbf{Application: Network Optimization}}} \\[3ex]\hline
    \cite{muhtasim2023continual} & Using the reservoir sampling buffer to enhance network anomaly detection in 5G Open-RAN by preserving previously learned knowledge. & Replay buffer for knowledge retention & Communication overhead \\
    \hline
    \cite{benzaid2024federated} & Combining federated learning with a replay memory-based continual learning strategy to enhance collaborative network anomaly detection in 5G Open-RAN. & Efficient replay memory strategy & Resource limitation \\
    \hline 
    \cite{xia2024fcllm} & A federated continual learning framework integrated with digital twin technology to enhance bearing fault diagnosis by addressing abnormal sensor data and maintaining data privacy across distributed factories. & Enhanced fault diagnosis accuracy & High computational and memory demands \\
    \hline
    \cite{xia2024fcla} & A federated continual learning framework with group signature authentication to ensure secure, efficient predictive maintenance for digital twins. & Efficient authentication with group signature & Complexity and computational overhead \\
    \hline 
    \cite{he2023federated} & Integrating the stacked broad learning system into a federated continual learning framework to facilitate continuous intrusion detection model training in unmanned aerial vehicle networks. & Efficient learning while privacy-preserving & Complexity in device selection \\ 
    \hline
\end{tabular}
\end{table*}
\subsubsection{Medical Diagnosis} In \cite{zheng2023asynchronous}, the ADFLL framework is introduced, combining asynchronous learning, decentralized communication, and continual learning principles to optimize landmark localization in medical imaging. This approach addresses challenges faced by traditional federated learning systems, particularly with complex, privacy-sensitive medical data. Each agent in the ADFLL framework utilizes experience replay buffers to store tuples generated during training, enabling them to learn from past experiences and mitigate catastrophic forgetting through selective experience replay. The asynchronous nature of ADFLL allows agents to train independently as soon as new data or buffers are available, which is particularly beneficial in medical settings with varying data availability. Collaboration among agents is facilitated through shared buffers, enriching the collective knowledge base while maintaining data privacy. {The framework was evaluated on the BraTS 2017 dataset, involving 285 patients with brain tumors across four MRI sequences (T1, T1ce, T2, FLAIR) and three orientations. Agents trained asynchronously on heterogeneous hardware achieved a mean localization error of 7.81 mm, outperforming traditional models in landmark localization across various imaging sequences.}

In \cite{huang2204continual}, the authors adopted a decentralized learning model for the study of brain metastasis identification, in which no central server is involved and the data is retained in the local center to comply with privacy regulations. Each center sequentially trains the model using locally available data and then passes the model to the next center. The learning process is facilitated through a technique called Synaptic Intelligence (SI), which plays a pivotal role in preserving essential model weights across different training centers. This method facilitates effective collaboration between multiple centers without sharing the actual patient data, thus respecting privacy concerns. The mathematical foundation of SI is based on the concept that the parameters that undergo significant changes during learning phases should be considered more important and, thus, less prone to modification in subsequent learning sessions. 
%
Once the importance weights are established, they are used to penalize changes to critical parameters in subsequent training sessions, thus ensuring that the knowledge acquired from previous tasks is not forgotten.
%
{A collaborative scenario across seven medical centers was simulated for 920 cases of enhanced T1 MRI brain transfer data \cite{van2016effects}. Using a decentralized joint learning framework, the model flows between centers via cyclic weight shifting and incorporates synaptic intelligence (SI) algorithms to prevent catastrophic forgetting. terative training further improved sensitivity to 0.914, matching centralized data pooling performance while reducing false positives to \textless 1 per volume.}


In \cite{wang2023peer}, the authors propose a novel peer-to-peer federated continual learning strategy called icP2P-FL to address improving the imaging performance of low-dose CT from multiple institutions. This approach integrates continual learning with a peer-to-peer federated learning framework to enable collaborative training of a global CT denoising model while preserving data privacy. The icP2P-FL method adopts a cyclic task-incremental continual learning strategy to ensure the model's performance across multiple institutions. In each cycle, the model has continually trained one institution after another via model transferring and inter-institutional parameter sharing. 
Furthermore, an intermediate controller is developed to make the overall training more flexible. It consists of a performance assessment module and an online determination module. The first module evaluates the denoising performance using quantitative measures, which are then fed into the second one to adjust the inter-institutional training order and determine the number of training rounds in each institution. 
{The model was evaluated on the AAPM Grand Challenge dataset \cite{yang2018autosegmentation} and three medical institutional datasets using the icP2P-FL framework for cross-protocol denoising. Dynamically adjust the training order and rounds through an intermediate controller. The denoising results on 100 cases of real low-dose CT data show that the structural similarity error is reduced to 3.2\% that meets the clinical diagnosis requirements.}

{\subsubsection{Network Application} In modern 5G open Radio Access Networks (RAN), anomaly detection is crucial due to increasing cyber threats such as distributed denial of service attacks. Traditional machine learning techniques, especially FL, face significant challenges, with one key issue being catastrophic forgetting, which is exacerbated in 5G networks owing to their disaggregated architecture and heterogeneous traffic. To address this problem, the authors in \cite{muhtasim2023continual} integrate the concept of replay buffers into FL, effectively mitigating catastrophic forgetting while also ensuring the privacy of data. Within this framework, anomaly detectors are deployed as applications in the near real-time RAN intelligent controller, and the FL aggregation server is located in the non-real-time RAN intelligent controller. This setup enables distributed anomaly detection across the network while maintaining local data privacy. The replay buffer stores a fixed number of past data samples using a method called reservoir sampling. This ensures that a representative dataset is available for continual learning, allowing the model to retain important information from prior attack patterns while adapting to new traffic types. As a result, the proposed solution improves the accuracy and stability of anomaly detection systems within RAN, addressing the evolving nature of cyber threats in 5G environments. Additionally, it supports scalable, privacy-preserving security measures that are essential for safeguarding the dynamic and rapidly evolving 5G networks.}

{In traditional Internet of Things scenarios that employ digital twins, sensor failures can lead to significant data loss, disrupting continuous data streams. Conventional diagnostic models, on the other hand, may struggle with erroneous sensor readings and often pose risks of privacy breaches. To address these challenges, the authors in \cite{xia2024fcllm} suggest an innovative solution that combines the concept of digital twins with retrieval-augmented generation-assisted large language models and federated continual learning. In the proposed framework, each factory deploys a digital twin to model the physical behavior of machine components, such as bearings, allowing it to refine and correct incoming sensor data to mitigate anomalies. When sensor malfunctions occur, the retrieval-augmented generation-assisted large language model generates virtual datasets to restore data continuity. Subsequently, local fault diagnosis models are trained on this enhanced data using an adaptive loss function, and periodic updates to these models are aggregated through a federated continual learning process, which employs weighted averaging. This dynamic aggregation approach ensures the protection of data privacy across distributed facilities while enabling both historical and new data to effectively contribute to the global model.}


{In addition to bearing fault diagnosis tasks, which require knowledge retention on time-series data, the application of digital twin technology in intrusion detection systems for Unmanned Aerial Vehicle (UAV) networks also faces numerous challenges, including fragmented and heterogeneous data sources, limited computational resources, and the phenomenon of catastrophic forgetting. To overcome these issues, the authors of \cite{he2023federated} introduce a federated continual learning framework that integrates an incremental learning approach based on a stacked broad learning system with a digital twin network. Each UAV locally collects network intrusion data and incrementally trains an intrusion detection model using the stacked broad learning system, an architecture that efficiently updates model parameters by stacking multiple learning blocks, thereby avoiding complete retraining with new data. Moreover, to ensure that only UAVs with high-quality data and adequate computing power participate in the global learning process, an intelligent selection strategy is implemented using a deep deterministic policy gradient algorithm. Selected UAVs then transmit their locally updated model parameters to a central server, where the global model is asynchronously aggregated, ensuring data privacy and robust adaptation to dynamic network conditions.}

\begin{table*}[h]
\centering
\caption{Resource Constraints in Real-World Applications of NCCL}
\label{resource}
\renewcommand{\arraystretch}{1.1}
\begin{tabular}{|m{0.05\textwidth}|m{0.45\textwidth}|m{0.20\textwidth}|m{0.20\textwidth}|} 
\hline 
    \textbf{Ref} & \textbf{Key Ideas} & {\textbf{Advantages}} & {\textbf{Limitations}} \\ \hline
        \multicolumn{4}{|c|}{\raisebox{-1.8ex}[0pt][0pt]{\centering \textbf{Resource: Memory Buffer}}} \\[3ex]\hline
        \cite{li2024towards} & Integrating global and local information to quantify sample importance for efficient replay.  & Private modular design and efficient sample caching. & Parameter sensitivity and Static client assumptions.\\ \hline
        \cite{li_tpds} & Calculating both domain-representative score and cross-client collaborative score for sample selection. & Secure synergistic caching mechanism. &  High computational overhead and hyperparameter sensitivity. \\ \hline
        \cite{dong2022federated} & Combining class-aware gradient compensation loss and class-semantic relationship distillation loss with limited buffers.  & Privacy-preserving prototype gradient communication.  &  Computational overhead from proxy server operations. \\ \hline
        \cite{dong2023no} & Combining category-balanced gradient adaptive compensation loss and category gradient-induced semantic distillation loss with an improved proxy server. & Adaptive gradient compensation loss. & Security reliance on proxy trust.  \\ \hline
        \cite{zhu2025dmaf} & Combining auxiliary client-based synthetic data generation and multi-teacher knowledge distillation to mitigate catastrophic forgetting without local storage of old data.  & Reduced dependence on memory buffer. & Increased computational and communication overhead due to synthetic data generation.  \\ \hline
        \multicolumn{4}{|c|}{\raisebox{-1.8ex}[0pt][0pt]{\centering \textbf{Resource: Computational Budget}}} \\[3ex]\hline
        \cite{le2021federated} & Implementing broad learning into the federated continual learning and updating the local model by the number of tasks and cached samples.  & Local training for fast updates and batch-asynchronous aggregation for efficiency. &  Communication overhead in batch processing and limited scalability for large client networks.\\  \hline
        \cite{li2024rehearsal} & Exploring the regularization technique and improving the synaptic intelligence with balanced global and local knowledge.  & Low computational overhead. & Dependence on SI algorithm. \\ \hline
        \cite{yuan2023peer} &  Using an iteration-based learning rate decay strategy and gossip protocol to efficiently learn new tasks.  & Low communication or computational overhead. &  Resource dependence on client devices. \\ \hline
        \cite{10779516} & Dynamically adjusting parameter importance via attention-weighted aggregation of cross-task/client models with balanced global-local knowledge.  & Balanced global-local learning and parameter-efficient. & Hyperparameter sensitivity. \\ \hline
        \multicolumn{4}{|c|}{\raisebox{-1.8ex}[0pt][0pt]{\centering \textbf{Resource: Public Dataset}}} \\[3ex]\hline
        \cite{wuerkaixi2024accurate} & Utilizing the normalizing flow model to construct the distribution and avoid introducing noisy data information. & Selective forgetting via probabilistic correlation estimation. & Computational overhead from generative model training. \\ \hline
        \cite{ma2022continual} &  Using public datasets as surrogate datasets to perform knowledge distillation at the client and server levels respectively.  & Client division mechanism for balanced learning. & Dependence on public surrogate dataset quality. \\  \hline
        \multicolumn{4}{|c|}{\raisebox{-1.8ex}[0pt][0pt]{\centering \textbf{Resource: Device Deployment}}} \\[3ex]\hline
        \cite{luopan2023fedknow} & Integrated signature task knowledge and gradient optimization to preserve task-specific information while minimizing negative knowledge transfer from non-IID data. & Lightweight and scalable client-side solution. & Dependence on hyperparameter tuning. \\ \hline
        \cite{zuo2024fedvit} & Leveraging sample-based signature task knowledge and gradient optimization, enabling lightweight, accurate ViT training on edge devices while minimizing communication overhead.  & Lightweight client-side solution with sample-based knowledge extraction. &  Impact of the hyperparameter selection. \\ \hline
        \cite{10779516} & Using an energy-efficient federated continual learning framework to optimize model training on edge devices with minimal energy usage. & Energy-efficient and suitable for small devices. & Limited scalability.\\ \hline
        \cite{liu2025sparse} & Mitigating catastrophic forgetting and system overhead by utilizing expandable models and dynamic sparse training. & Reduced system overhead and suitable for small devices. & Model complexity in large tasks.\\ \hline
\end{tabular}
\end{table*}

\subsection{Resource Constraint}
In the previous section, we have already discussed the applications of NCCL in three classic real-world scenarios. While NCCL has shown promising results in enabling collaborative learning among distributed devices without compromising data privacy, its practical deployment is often hindered by the limited availability of memory, computational resources, and access to public datasets. These constraints are particularly prevalent in edge devices, such as smartphones and IoTs, which typically have restricted storage capacity, processing power, and bandwidth \cite{hamdan2020edge}. Moreover, the increasing complexity and heterogeneity of real-world tasks further exacerbate the resource demands on NCCL systems. Thus, it is crucial to develop resource-efficient NCCL methods that can effectively operate under these constraints while maintaining high performance and adaptability. We consider the following three key factors to explore the resource limitations of NCCL, and Table \ref{resource} shows a summary of these resource-constrained scenarios of existing NCCL research.
\begin{itemize}
    \item \emph{Memory Buffer:} In NCCL, memory buffers are an important component for storing and retrieving previously learned knowledge \cite{li2024towards,li_tpds}. Existing methods usually rely on data rehearsal strategies to prevent the degradation of past knowledge, including caching samples of previous tasks or generating synthetic samples for replay so that the model can adapt to new tasks without catastrophic forgetting. However, in many real-world scenarios, such as IoT devices or mobile phones, buffers are usually limited \cite{javed2018internet}. Therefore, it is crucial to develop rehearsal-free NCCL methods that can effectively retain past knowledge while minimizing the dependency on buffers. 
    \item \emph{Computational Budget:} The computational budget is another critical factor in NCCL. As the number and complexity of tasks increase, the computational cost of NCCL can quickly become prohibitive, particularly for edge devices with limited processing power. This can lead to slow convergence, high energy consumption, and reduced responsiveness of the NCCL system \cite{li2024rehearsal}. To address this issue, it is necessary to design computing-efficient NCCL algorithms that can achieve good performance with minimal computational overhead.
    \item \emph{{Public Dataset:}} Access to large-scale, clean, and labeled datasets is crucial for the effective training and evaluation of NCCL methods. However, in many real-world applications, such datasets may not be readily available due to privacy concerns, data scarcity, or the cost of annotation \cite{wang2024comprehensive}. This can pose significant challenges for NCCL methods that rely on supervised learning directly from local datasets \cite{ma2022continual}. To overcome this limitation, it is important to explore an unsupervised paradigm or improve the model's robustness against noisy data that can effectively promote real-world deployment.
    \item {\emph{Device Deployment:} Deploying models on resource-constrained devices presents a significant challenge in NCCL. With the increasing number of devices, managing communication overhead and ensuring efficient model updates across all participants becomes complex \cite{zuo2024fedvit}. Additionally, small devices with limited processing power and memory exacerbate the problem, making it difficult to deploy large-scale models. To address these challenges, it is essential to develop lightweight models that can adapt to resource constraints while maintaining performance \cite{liu2025sparse}. Moreover, optimizing communication strategies and reducing computational requirements through techniques such as model pruning will be crucial for successful deployment in real-world settings.}
    
\end{itemize}

\subsubsection{Memory Buffer} In \cite{li2024towards}, the authors considered the data heterogeneity and the limited storage space of the client memory buffer and proposed a simple and effective framework to solve the catastrophic forgetting with synergistic replay. The core idea of Re-Fed is to coordinate clients to cache important samples of previous tasks for replay under the limited memory of the client before the arrival of the new task. Specifically, when a new task arrives at the client, samples from previous tasks are cached according to their global and local importance. Then, the client trains the local model using both the cached samples and the samples from the new task. To quantify the importance of samples, the authors introduce an additional personalized informative model that integrates knowledge from the global model and the local model. Then, after the client receives the global model, the PIM is updated iteratively using a momentum-based method.
In the process of updating the model PIM, the gradient norm of each sample is recorded to calculate the sample importance score. 
Based on the importance scores, each client caches the most informative samples within its memory buffer limit. The cached samples are then used for replay during the training of new tasks to alleviate catastrophic forgetting.

Similar to \cite{dong2022federated}, the authors in \cite{dong2023no} proposed a novel Local-Global Anti-forgetting (LGA) model. This method is also designed to solve the problem of alleviating the catastrophic forgetting of the global model on old categories when the local client has limited memory. Compared with GLFC, both consider the class imbalance of new and old categories of the local client and the imbalance of non-IID categories between clients, and both use a proxy server to select the best old global model for the local client to perform knowledge distillation to solve the global forgetting caused by the imbalance of non-IID categories across clients. The difference is that LGA introduces a category-balanced gradient adaptive compensation loss and a category gradient-induced semantic distillation loss. For global forgetting, LGA improves the proxy server by introducing a self-supervised prototype augmentation strategy to augment the reconstructed perturbed prototype images, thus enhancing the robustness in selecting the best old global model, while GLFC does not include this enhancement strategy. Extensive experiments on representative datasets demonstrate that LGA outperforms state-of-the-art methods by 4.8\%$\sim$27.3\% in terms of average accuracy. Moreover, the ablation studies validate the effectiveness of each module in the proposed model, especially the category-balanced gradient-adaptive compensation loss and category gradient-induced semantic distillation loss.

\subsubsection{Computational Budget}
In \cite{le2021federated}, the authors introduced a federated continual learning scheme with a broad learning framework that effectively addresses computational constraints in real-world scenarios. Central to their approach is the utilization of the broad learning paradigm, which enables efficient incremental learning without requiring full retraining of local models by each client upon encountering new data, thereby reducing computational overhead. This framework operates on the principle that local training initiates when the influx of new samples exceeds a predefined threshold. During local updates, the update of local weights for client $i$ in update $j$ is computed using a specific equation that adjusts the prior weights $w$ based on the difference between new labels and model predictions, incorporating updated feature and enhancement nodes of the incremental network.

In \cite{yuan2023peer}, the authors proposed a novel peer-to-peer federated learning framework to address the resource constraints and privacy protection issues in driver behavior recognition tasks for the internet of vehicles. From the perspective of computational budget, the method employs a continual learning paradigm to propagate knowledge directly among clients, avoiding the computational overhead caused by central server aggregation. Specifically, when a new task arrives, each client first receives the model parameters from the previous client as initialization and then fine-tunes the model on its local data. To balance personalization and generalization, it introduces a loss function that consists of two parts: a negative log-likelihood loss and a proximal term. Moreover, the authors adopt an iteration-based learning rate decay strategy to accelerate convergence and prevent catastrophic forgetting. In terms of communication, the method utilizes a gossip protocol to randomly propagate models among clients, overcoming the highly dynamic nature and distance limitations of vehicle connections. Experiments on two real-world driver behavior recognition datasets (StateFarm \cite{statefarm} and AICity \cite{wang20248th}) demonstrate that FedPC is highly competitive compared to traditional FL in terms of performance, knowledge dissemination rate, and compatibility with new clients.


\subsubsection{{Public Dataset}}
In \cite{ma2022continual}, the authors introduce a novel continual federated learning framework known as CFeD, which ingeniously leverages public datasets as surrogate data to perform knowledge distillation both at the client and server levels, thus effectively strengthening the local datasets. Each client and the server maintain an independent surrogate dataset, which is not directly derived from the client's private data, thus addressing privacy concerns effectively. When introduced to a new task, each client utilizes its surrogate dataset to transfer knowledge from the previous tasks' model to the new model, thereby alleviating inter-task forgetting. 

To mitigate the impact of noisy data in streaming tasks, \cite{wuerkaixi2024accurate} proposes a novel method called accurate forgetting federated continual learning. The core of this method lies in the application of a normalizing flow model to quantify the credibility of previous knowledge and achieve accurate forgetting. The model is trained in the feature space of the classifier to generate features that represent the previous tasks. These generated features are then used to augment the learning process of the classifier, preventing complete forgetting of previous tasks. The correlation probability of the generated features is used to re-weight the loss objective, enabling the classifier to selectively forget biased or irrelevant features. Specifically, the loss objective is composed of three terms: the classification loss on real data, the classification loss on generated features, and a regularization term that penalizes large deviations of the generated features from the true data distribution. The weights of these terms are adjusted based on the correlation probability, allowing the classifier to adaptively forget erroneous information and learn unbiased features. 

\subsubsection{{Device Deployment}}

{In \cite{luopan2023fedknow}, the authors proposed a client-side federated continual learning framework from the perspective of handling device number expansion. To handle the increase in device numbers, the method extracts and integrates signature task knowledge. Each client has a knowledge extractor that retains partial model weights as task knowledge, reducing computational load. A gradient integrator combines gradients from the current and relevant past tasks to prevent catastrophic forgetting and negative knowledge transfer.}

{Similarly, in \cite{zuo2024fedvit}, the FedViT framework is proposed for ViT models in federated continual learning. It also adopts a client-side strategy. FedViT extracts knowledge as a subset of training samples (e.g., 10\% per task) to reduce device-side burden. Leveraging ViT's decomposability, it trains the model in stages. FedViT consists of three key components: knowledge extractor, gradient restorer, and gradient integrator. The knowledge extractor stores the knowledge of each task by retaining a subset of training samples that are well-adapted to the trained ViT model and maintains the class distribution of the task's complete samples.} 
%
{Extensive experiments on various datasets demonstrate that FedViT significantly improves model accuracy without increasing training time and reduces communication costs. Moreover, FedViT achieves more improvements under difficult scenarios such as large numbers of tasks or clients and training different complex ViT models.}

{In \cite{10779516}, the author introduces an energy-efficient framework for personalized federated continual learning, specifically designed to address the challenges of continual learning on edge devices with limited computational resources. To address this issue, the model is divided into two parts: one part retains old knowledge, while the other learns new tasks, minimizing computational costs. A sparse pruning strategy is adopted to preserve critical model components without sacrificing performance.}
{This sparse optimization approach reduces both memory and computation requirements, making it particularly well-suited for edge devices. The experimental results demonstrate that the proposed method significantly reduces energy consumption by training only part of the network. In terms of scalability, as the number of devices increases from 20 to 40, the proposed method continues to maintain high performance.}

In \cite{liu2025sparse}, the author suggests a sparse personalized federated class-incremental learning framework to tackle key challenges in continual learning on edge devices with limited resources. The framework targets issues such as catastrophic forgetting and system overhead and proposes two main solutions: expandable class-incremental models and dynamic sparse training. To mitigate catastrophic forgetting, the framework replaces single-task models with expandable class-incremental models. This approach allows for task-specific blocks to be added as new tasks are learned, preventing the model from forgetting old knowledge while enhancing its capacity for new tasks. Additionally, dynamic sparse training customizes sparse local models for each client, reducing system overhead by pruning unnecessary parameters, thus mitigating data heterogeneity and over-parameterization without increasing computational costs. Personalized sparse masks optimize the model by adjusting to task requirements, ensuring efficient resource use and updating only critical components. This approach does not require additional data for knowledge distillation, making it ideal for edge devices with privacy concerns. 

\vspace{3pt}
\noindent {\textbf{Conclusion.} This section examines three downstream task scenarios of NCCL and four resource constraint issues in real-world deployment contexts. NCCL has been extensively investigated across numerous practical applications, demonstrating its critical role in economic development. Unfortunately, resource constraint issues remain under-investigated. Future research should prioritize lightweight NCCL algorithms and extend their applicability to broader real-world scenarios.}


\section{Benchmark Experiments}\label{VII}
In this section, we will establish benchmarks to empirically evaluate the performance of current NCCL methods. We are the first to construct such benchmarks for testing. Here, considering the diversity of NCCL scenarios, we only adopt task scenarios that combine FL and CL, also commonly referred to as federated continual learning, which is a mainstream research direction with substantial existing studies. Below, we will introduce the datasets, baselines, experimental configurations, and the analysis of experimental results \footnote{We will continually organize the relevant code and datasets, and all related methods will be integrated into a unified code framework to facilitate comparison and deployment by researchers. The code will be available at: \url{https://github.com/YichenLi-Hust/NCCL/}}.

\subsection{Dataset}
For benchmark datasets, we considered two of the most common CL scenarios: class-incremental learning (Class-IL) and domain-incremental learning (Domain-IL). For each scenario, we utilized three datasets to conduct experiments. The specific details of the datasets are as follows:

\noindent\textbf{Class-Incremental Task:} Datasets in the Class-IL scenario progressively introduce new classes over time, starting with a subset of classes and expanding in subsequent stages. This enables models to adapt to an increasing number of classes.
\begin{itemize}
    \item \textit{CIFAR-10} \cite{krizhevsky2009learning}: A dataset includes 10 different object categories, each sample represented as a 32$\times$32-pixel color image. It comprises 50,000 training samples and 10,000 test samples.
    \item \textit{CIFAR-100} \cite{krizhevsky2009learning}: A dataset with 100 fine-grained classes, including 50,000 training images and 10,000 test images, provides a challenging benchmark for image recognition.
     \item \textit{Tiny-ImageNet} \cite{le2015tiny}: A subset of ImageNet with 200 classes and 120,000 images (100,000 training, 10,000 validation, and 10,000 tests).   
\end{itemize}

\noindent\textbf{Domain-Incremental Task:} Initially comprising instances from a particular domain, Domain-IL progressively incorporates novel domains over time. This setup effectively facilitates adaptation and generalization to previously unseen domains.

\begin{itemize}
    \item \textit{Digit-10}: A dataset encompasses ten categories of digits across four domains: MNIST \cite{lecun2010mnist}, EMNIST \cite{cohen2017emnist}, USPS \cite{hull1994database}, and SVHN \cite{netzer2011reading}. Each dataset represents ten classes of digit images within a specific domain, such as handwriting, printed text, and others.
    \item \textit{Office-31} \cite{saenko2010adapting}: A dataset consists of images gathered from three varying domains: Amazon, Webcam, and DSLR. It encompasses 31 categories of objects, with approximately 4,100 images belonging to each domain.
    \item \textit{Office-Caltech-10} \cite{zhang2020impact}: A dataset featuring images sourced from four distinct domains: Amazon, Caltech, Webcam, and DSLR. It encompasses 10 categories of objects, with approximately 2,500 images per domain.
\end{itemize}

\begin{table*}[h]
    \renewcommand\arraystretch{1.1}
    \centering
    \caption{Experimental Details. Settings of different datasets in the experiments section.}
    \label{conf}
    \resizebox{0.85\linewidth}{!}{
    \begin{tabular}{l | c c c | c c c}
     \toprule[1pt]
      \textbf{Attributes} & \textbf{CIFAR-10} & \textbf{CIFAR-100} & \textbf{Tiny-ImageNet} & \textbf{Digit-10} & \textbf{Office-31} & \textbf{Office-Caltech-10} \\  \hline
      Task size & 178MB & 178MB & 435MB & 480M & 88M & 58M\\
      Image number & 60K & 60K & 120K & 110K & 4.6k & 2.5k\\
      Image Size & 3$\times$32$\times$32 & 3$\times${32}$\times$32 & 3$\times$64$\times$64 & 1$\times$28$\times$28 & 3$\times$300$\times$300 & 3$\times$300$\times$300\\
      Task number & $n$ = 5 & $n$ = 10 & $n$ = 10 & $n$ = 4 & $n$ = 3 & $n$ = 4\\
      Task Scenario & Class-IL & Class-IL & Class-IL & Domain-IL & Domain-IL & Domain-IL\\ \hline
      Batch Size & $s$ = 64 & $s$ = 64 & $s$ =128 & $s$ = 64 & $s$ = 32 & $s$ = 32\\
      ACC metrics & Top-1 & Top-1 & Top-10 & Top-1 & Top-1 & Top-1\\
      Learning Rate & $l$ = 0.01 & $l$ = 0.01 & $l$ = 0.001 & $l$ = 0.001 & $l$ = 0.01 & $l$ = 0.01\\
      Data heterogeneity & $\alpha$ = 0.1 & $\alpha$ = 1.0 & $\alpha$ = 10.0 & $\alpha$ = 0.1 & $\alpha$ = 1.0 & $\alpha$ = 1.0\\
      Client numbers & $C$ = 20 & $C$ = 20 & $C$ = 20 & $C$ = 15 & $C$ = 10 & $C$ = 8\\
      Local training epoch & $E$ = 20 & $E$ = 20 & $E$ = 20 & $E$ = 20 & $E$ = 20 & $E$ = 15\\
      Client selection ratio & $k$ = 0.4 & $k$ = 0.5 & $k$ = 0.6 & $k$ = 0.4 & $k$ = 0.4 & $k$ = 0.5\\
      Communication Round  & $T$ = 80 & $T$ = 100 & $T$ = 100 & $T$ = 60 & $T$ = 60 & $T$ = 40\\ 
    \bottomrule[1pt]
    \end{tabular}}
\end{table*}

\begin{table*}[t]
    \renewcommand\arraystretch{1.1}
    \centering
    \caption{Performance comparison of various methods in two incremental scenarios.}
    \vspace{-2pt}
    \label{RESULT}
    \resizebox{\linewidth}{!}{
    \begin{tabular}{c  c c c c c c |c c c c c c}
     \toprule
     \multirow{2}{*}{Method} & \multicolumn{2}{c}{CIFAR-10} &  \multicolumn{2}{c}{CIFAR-100} & \multicolumn{2}{c|}{Tiny-ImageNet}& \multicolumn{2}{c}{Digit-10} &  \multicolumn{2}{c}{Office-31} & \multicolumn{2}{c}{Office-Caltech-10}\\ \cline{2-13}
     & $A(f)$ & $\bar{A}$ & $A(f)$ & $\bar{A}$ & $A(f)$ & $\bar{A}$ & $A(f)$ & $\bar{A}$ & $A(f)$ & $\bar{A}$ & $A(f)$ & $\bar{A}$\\ \hline
      FedAvg & 36.68\scriptsize{±1.32}  & 59.17\scriptsize{±0.08}
       & 27.15\scriptsize{±0.87}
      & 41.36\scriptsize{±0.24}
     & 30.16\scriptsize{±0.19} & 50.65\scriptsize{±0.11}& 68.12\scriptsize{±0.04} & 80.34\scriptsize{±0.02} & 48.97\scriptsize{±0.74} & 56.29\scriptsize{±1.15} & 55.41\scriptsize{±0.52} &57.61\scriptsize{±0.93}\\
     %
      %
      Re-Fed & 38.08\scriptsize{±0.46} & 59.02\scriptsize{±0.31}  & 32.95\scriptsize{±0.31} & 42.50\scriptsize{±0.18}  & 33.43\scriptsize{±0.54} & 51.98\scriptsize{±0.32}  & 67.85\scriptsize{±0.37} & 79.85\scriptsize{±0.25}  & 50.11\scriptsize{±0.29} & 57.46\scriptsize{±0.34}  & 59.16\scriptsize{±0.40} & 60.01\scriptsize{±0.33} \\
      FedCIL & 37.96\scriptsize{±1.68} & 58.30\scriptsize{±1.22}  & 30.88\scriptsize{±1.04} & 42.16\scriptsize{±0.97}  & 31.35\scriptsize{±1.27} & 50.93\scriptsize{±0.84}  & 68.17\scriptsize{±0.85} & 80.02\scriptsize{±0.63}  & 49.15\scriptsize{±0.92} & 56.78\scriptsize{±0.79}  & 57.80\scriptsize{±0.74} & 59.13\scriptsize{±0.52} \\
      GLFC & 38.43\scriptsize{±1.43} & 60.03\scriptsize{±1.16}  & 33.17\scriptsize{±0.62} & 43.28\scriptsize{±0.81}  & 32.11\scriptsize{±0.40} & 51.79\scriptsize{±0.54}  & 68.12\scriptsize{±0.04} & 80.34\scriptsize{±0.02} & 48.97\scriptsize{±0.74} & 56.29\scriptsize{±1.15} & 55.41\scriptsize{±0.52} &57.61\scriptsize{±0.93} \\
      FOT & 40.18\scriptsize{±1.26} & 61.41\scriptsize{±0.66}  & 36.15\scriptsize{±0.40} & 43.14\scriptsize{±0.51}  & 37.23\scriptsize{±0.14} & 54.87\scriptsize{±0.22}  & 68.54\scriptsize{±0.38} & 79.70\scriptsize{±0.11}  & 49.12\scriptsize{±0.83} & 56.17\scriptsize{±0.44}  & 60.30\scriptsize{±0.13} & 60.76\scriptsize{±0.98} \\
      FedWeIT & 37.96\scriptsize{±0.52} & 59.89\scriptsize{±0.12} & 35.84\scriptsize{±1.05} & 44.20\scriptsize{±0.45}   & 34.98\scriptsize{±0.93} & 53.04\scriptsize{±0.71}   & 69.71\scriptsize{±0.20} & 80.91\scriptsize{±0.09}   & 51.49\scriptsize{±0.64} & 57.83\scriptsize{±0.89}   & 58.53\scriptsize{±0.49} & 59.72\scriptsize{±0.70} \\
     CFeD & 38.19\scriptsize{±1.44} & 60.42\scriptsize{±1.32} & 32.69\scriptsize{±0.94} & 40.30\scriptsize{±1.28}   & 32.22\scriptsize{±0.86} & 50.10\scriptsize{±0.92}   & 66.89\scriptsize{±0.52} & 77.58\scriptsize{±0.76}   & 46.44\scriptsize{±0.85} & 53.29\scriptsize{±0.71}   & 54.11\scriptsize{±0.59} & 55.75\scriptsize{±0.88} \\
     Target & 37.68\scriptsize{±1.24} & 58.12\scriptsize{±1.63} & 33.15\scriptsize{±1.86} & 41.05\scriptsize{±1.51} & 29.65\scriptsize{±1.41} & 48.37\scriptsize{±1.09} & 67.19\scriptsize{±1.13} & 76.32\scriptsize{±0.85} & 47.51\scriptsize{±1.46} & 54.82\scriptsize{±0.97} & 55.72\scriptsize{±0.65} & 58.23\scriptsize{±0.55} \\
     AF-FCL & 39.23\scriptsize{±1.33} & 59.78\scriptsize{±1.67} & 34.16\scriptsize{±1.15} & 41.94\scriptsize{±0.91} & 30.95\scriptsize{±1.26} & 50.88\scriptsize{±1.10} & 68.08\scriptsize{±0.95} & 78.65\scriptsize{±0.77} & 48.22\scriptsize{±0.86} & 57.13\scriptsize{±0.92} & 58.81\scriptsize{±1.24} & 60.36\scriptsize{±1.05} \\
     MFCL & 41.26\scriptsize{±0.86} & 62.04\scriptsize{±0.77} & 34.32\scriptsize{±1.35} & 42.79\scriptsize{±1.28} & 31.46\scriptsize{±0.84} & 51.19\scriptsize{±0.92} & 66.33\scriptsize{±0.70} & 75.79\scriptsize{±0.76} & 47.69\scriptsize{±0.67} & 55.24\scriptsize{±0.89} & 56.67\scriptsize{±1.05} & 55.12\scriptsize{±0.83} \\
     pFedDIL& 37.52\scriptsize{±0.82} & 57.29\scriptsize{±0.59} & 34.30\scriptsize{±0.61} & 41.96\scriptsize{±0.83} & 32.87\scriptsize{±0.39} & 50.05\scriptsize{±0.42} & 71.40\scriptsize{±0.59} & 83.80\scriptsize{±0.77} & 53.73\scriptsize{±0.35} & 58.49\scriptsize{±0.39} & 63.13\scriptsize{±0.81} & 63.71\scriptsize{±0.72} \\
     SR-FDIL& 37.98\scriptsize{±0.61} & 58.84\scriptsize{±0.80} & 32.78\scriptsize{±0.48} & 41.55\scriptsize{±0.36} & 31.74\scriptsize{±0.78} & 50.28\scriptsize{±0.66} & 69.86\scriptsize{±0.35} & 81.97\scriptsize{±0.24} & 52.50\scriptsize{±0.49} & 59.21\scriptsize{±0.53} & 60.92\scriptsize{±0.70} & 63.29\scriptsize{±0.44} \\
    \bottomrule
    \end{tabular}}
\end{table*}

\begin{table*}[t]
    \renewcommand\arraystretch{1.1}
    \centering
    \caption{{Evaluation based on the number of communication rounds to reach the best accuracy. We use "$\Delta$" to represent the difference between the percentage increase in accuracy and the percentage increase in rounds for other baselines compared to FedAvg.}}
    \label{round}
    \vspace{-2pt}
    \resizebox{\linewidth}{!}{\begin{tabular}{l  c c c c c c| c c c c c c}
     \toprule[1pt]
     
     \multirow{2}{*}{Method} & \multicolumn{2}{c}{CIFAR-10} &  \multicolumn{2}{c}{CIFAR-100} & \multicolumn{2}{c|}{Tiny-ImageNet}& \multicolumn{2}{c}{Digit-10} &  \multicolumn{2}{c}{Office-31} & \multicolumn{2}{c}{Office-Caltech-10}\\
     \cline{2-13}
     & Rounds & $\Delta$ & Rounds & $\Delta$ & Rounds & $\Delta$ & Rounds & $\Delta$ & Rounds & $\Delta$ & Rounds & $\Delta$ \\
     \hline

      FedAvg& 
      304\scriptsize{±2.31}&/&
      839\scriptsize{±1.67}&/&
      893\scriptsize{±2.93}&/&
      208\scriptsize{±1.58}&/&  
      165\scriptsize{±0.94}&/&  
      132\scriptsize{±1.25}&/ \\

  Re-Fed&
  319\scriptsize{±1.23}&1.03\%$\downarrow$&
844\scriptsize{±2.15}&17.01\%$\uparrow$&
894\scriptsize{±1.92}&9.67\%$\uparrow$&
222\scriptsize{±1.48}&6.70\%$\downarrow$&
168\scriptsize{±0.98}&0.49\%$\uparrow$&
145\scriptsize{±0.76}&2.63\%$\downarrow$ \\
      FedCIL&
      323\scriptsize{±2.78}&2.51\%$\downarrow$&
866\scriptsize{±2.25}&8.96\%$\uparrow$&
903\scriptsize{±1.94}&2.69\%$\uparrow$&
220\scriptsize{±1.42}&5.38\%$\downarrow$&  
168\scriptsize{±1.07}&1.42\%$\downarrow$&  
147\scriptsize{±0.91}&6.07\%$\downarrow$ \\
      GLFC&
312\scriptsize{±1.47}&1.99\%$\uparrow$&
830\scriptsize{±2.03}&19.23\%$\uparrow$&
901\scriptsize{±2.12}&5.18\%$\uparrow$&
220\scriptsize{±0.96}&6.54\%$\downarrow$&  
166\scriptsize{±1.14}&1.99\%$\downarrow$&  
158\scriptsize{±1.31}&11.60\%$\downarrow$  \\
            FOT&
      306\scriptsize{±2.26}&8.06\%$\uparrow$&
840\scriptsize{±1.35}&24.78\%$\uparrow$&
903\scriptsize{±1.09}&17.88\%$\uparrow$&
220\scriptsize{±0.76}&4.84\%$\downarrow$&  
170\scriptsize{±0.98}&2.64\%$\downarrow$&  
158\scriptsize{±1.04}&8.35\%$\downarrow$ \\

FedWeIT&      312\scriptsize{±1.95}&0.81\%$\uparrow$&
870\scriptsize{±1.58}&20.68\%$\uparrow$&
880\scriptsize{±2.26}&15.26\%$\uparrow$&
217\scriptsize{±1.31}&1.87\%$\downarrow$&
165\scriptsize{±1.79}&4.89\%$\uparrow$&  
141\scriptsize{±1.52}&1.05\%$\downarrow$\\

CFed&      
317\scriptsize{±2.47}&0.15\%$\downarrow$&
845\scriptsize{±2.68}&16.24\%$\uparrow$&
907\scriptsize{±1.54}&4.85\%$\uparrow$&
225\scriptsize{±1.25}&9.39\%$\downarrow$&
164\scriptsize{±0.82}&4.84\%$\downarrow$&  
151\scriptsize{±0.58}&14.99\%$\downarrow$\\

Target &
310\scriptsize{±2.78}&0.72\%$\uparrow$&
851\scriptsize{±3.09}&16.69\%$\uparrow$&
897\scriptsize{±2.31}&2.17\%$\downarrow$&
220\scriptsize{±1.48}&6.84\%$\downarrow$&
166\scriptsize{±1.09}&3.68\%$\downarrow$&
147\scriptsize{±1.70}&9.65\%$\downarrow$\\

AF-FCL & 
321\scriptsize{±2.53}&1.20\%$\uparrow$&
862\scriptsize{±3.21}&17.85\%$\uparrow$&
911\scriptsize{±2.18}&0.58\%$\uparrow$&
217\scriptsize{±1.75}&4.21\%$\downarrow$&
166\scriptsize{±1.31}&2.16\%$\downarrow$&
145\scriptsize{±1.15}&3.18\%$\downarrow$\\

MFCL &
320\scriptsize{±1.63}&6.10\%$\uparrow$&
855\scriptsize{±2.17}&19.02\%$\uparrow$&
900\scriptsize{±1.93}&3.35\%$\uparrow$&
223\scriptsize{±1.65}&9.43\%$\downarrow$&
167\scriptsize{±0.94}&3.88\%$\downarrow$&
142\scriptsize{±0.87}&4.82\%$\downarrow$\\

pFedDIL&
312\scriptsize{±1.94}&0.33\%$\downarrow$&
868\scriptsize{±1.88}&17.50\%$\uparrow$&
908\scriptsize{±1.12}&6.59\%$\uparrow$&
202\scriptsize{±0.73}&7.56\%$\uparrow$&
162\scriptsize{±1.25}&10.71\%$\uparrow$&
140\scriptsize{±1.06}&6.51\%$\uparrow$\\

SR-FDIL& 
314\scriptsize{±1.85}&0.24\%$\uparrow$&
847\scriptsize{±2.24}&16.23\%$\uparrow$&
905\scriptsize{±1.36}&3.65\%$\uparrow$&
205\scriptsize{±1.27}&3.95\%$\uparrow$&
166\scriptsize{±1.10}&6.12\%$\uparrow$&
138\scriptsize{±1.34}&4.70\%$\uparrow$\\
    \bottomrule[1pt]
    \end{tabular}}
\end{table*}

\subsection{Baseline}
For the baseline, we chose relevant work published in top conferences and journals (such as ICLR, NeurIPS, ICML, CVPR, ECCV, ICCV, etc.) in recent years to ensure the superiority and advancement of baselines. To ensure a fair comparison among baselines, we follow the protocols established in \cite{fedavg} and \cite{li2024towards} to set up tasks: FedAvg \cite{fedavg}, Re-Fed \cite{li2024towards}, FedCIL \cite{qi2023better}, GLFC \cite{dong2022federated}, FOT \cite{bakman2023federated}, FedWeIT \cite{yoon2021federated}, CFeD \cite{CFED}, Target \cite{zhang2023target}, AF-FCL \cite{wuerkaixi2024accurate}, MFCL \cite{MFCL}, pFedDIL \cite{li2024pfeddil}, SR-FDIL \cite{li_tpds}. As these baselines have been extensively discussed in previous sections, we omit detailed descriptions here to avoid redundancy.

\subsection{Configuration}
We assign different numbers of tasks to various datasets. For class-incremental tasks, using CIFAR-10 as an example, we set five tasks, with each task covering two classes without data overlap. For domain-incremental tasks, each domain represents one task. Dirichlet distribution $Dir(\alpha)$ \cite{minka2000estimating} is used to allocate local samples, ensuring samples of each class are split across clients in different proportions to induce data heterogeneity. A smaller value of $\alpha$ indicates higher data heterogeneity. To ensure the elimination of variances between baseline methods, unless otherwise specified, we uniformly employ ResNet18 \cite{he2016deep} as the backbone model. The implementation of FedCIL and AF-FCL relies on the backbone model itself. To ensure fair comparisons, we set a memory buffer of 300 for each client model to restrict data replay or synthetic samples (e.g., FedAvg can cache 300 samples for each client, and FedCIL also can only generate 300 synthetic samples for local training). Each experimental set is repeated twice, with the average and standard deviation of accuracy computed over the last 10 rounds. We report the final accuracy $A(f)$ after completing the last task and the average accuracy $\bar{A}$ for all tasks. Detailed configurations are presented in Table \ref{conf}.

\subsection{Performance Overview}
\subsubsection{{Test Accuracy}}
Table \ref{RESULT} compares the test accuracy of various methods on six datasets. While most methods outperform FedAvg, their performance gains remain limited compared to original studies due to constrained memory buffers for fair comparison. In Class-IL tasks, FOT achieves state-of-the-art performance on Tiny-ImageNet by eliminating data storage requirements through orthogonal projection techniques for parameter isolation. While generator-based methods (e.g., MFCL/AF-FCL) perform well on CIFAR datasets, their effectiveness diminishes significantly on complex datasets like Tiny-ImageNet, indicating limitations in handling large-scale incremental learning. For Domain-IL tasks, regularization methods (e.g., FOT) maintain effectiveness, while data-replay methods struggle under limited memory buffers. Notably, pFedDIL and SR-FDIL excel in domain adaptation but show limited efficacy in Class-IL scenarios. Our analysis reveals no existing method universally addresses both Class-IL and Domain-IL challenges in federated continual learning, highlighting the need for more versatile approaches to handle diverse incremental learning paradigms.

\subsubsection{{Communication Efficiency}}
Table \ref{round} compares communication rounds required by each method to achieve optimal test accuracy, using FedAvg as the baseline for accuracy-efficiency trade-off evaluation. In Class-IL tasks, all methods show positive gains over FedAvg, with accuracy improvements (avg. +10.36\%) exceeding added communication costs (avg. +1.97\%). For example, MFCL on CIFAR-10 achieves the highest precision despite moderate communication increases. GLFC, FOT, and MFCL exhibit optimal accuracy-efficiency balance. For Domain-IL tasks, Class-IL-oriented methods yield negative $\Delta$ values as communication costs surpass accuracy gains. In contrast, Domain-IL-specific methods (e.g., pFedDIL, SR-FDIL) maintain low overhead while improving accuracy, achieving positive gains. This highlights task-specific design requirements in federated continual learning.

\noindent\textbf{Limitations.} In this paper, we simply set a memory buffer for each baseline. However, it should be noted that there are numerous factors that can affect model performance, such as bandwidth, computing overhead, model size, and so on. Achieving absolute uniformity in hardware environments for all methods in large-scale experiments is extremely challenging. We chose to unify the memory buffer based on its significance as a resource factor in NCCL, which has a substantial impact on model performance. In future work, we will gradually expand the benchmark framework to enable the adjustment of different resources.

\section{Discussion about future research directions in Non-centralized CL}\label{VIII}
Although there is a lot of existing research, there are still challenging new research directions in the deployment of NCCL to be discussed as follows.

\begin{itemize}
    \item \textit{Computing Saving:} Some existing findings \cite{li2025resource} have proved that computational overhead constitutes a pivotal factor in the majority of NCCL methods. However, the acquisition of computational resources, particularly on distributed platforms, frequently incurs substantial costs. Consequently, there is a pressing need to explore more computationally parsimonious algorithms. Furthermore, an alternative perspective revolves around accelerating the convergence of the model, which can significantly reduce the computational cost. 
    \item \textit{Unsupervised Learning:} The existing methods presume full labeling for each task's data, overlooking the reality that real-world situations frequently include both labeled and unlabeled data. Considering the substantial cost of manual annotation, achieving full supervision is unfeasible. Therefore, research into the use of unsupervised or semi-supervised learning within the NCCL framework is crucial, as it can greatly facilitate the practical deployment of NCCL.
    \item \textit{Resource Allocation:} Given limited training resources, optimizing their allocation to improve model performance is a practical solution. This may involve allocating suitable memory buffers to various data and assigning computational resources to specific processes. One approach to consider is using combinatorial optimization techniques, as suggested by studies such as \cite{woo2023convnext}. In practical applications or method designs, dynamically allocating resources for each technique according to availability can further enhance performance.
    \item \textit{Online Learning:} Current NCCL research mainly focuses on offline settings, where task data is collected completely before training and remains unchanged. However, in real-world situations, data from ongoing tasks typically arrives in small, incremental batches rather than all at once. Furthermore, the widespread use of resource-limited smart and edge devices requires frequent model updates with incoming data to optimize memory usage and communication efficiency. This presents a difficulty for generative-based methods, as they face challenges in training effective generators with limited data samples.
    \item \textit{LLM Foundation:} Recently, LLM has attracted considerable attention because of its remarkable abilities, prompting researchers to investigate its integration into NCL frameworks \cite{huang2024mllm}. However, integrating LLM into FIL environments poses a significant challenge. In such settings, distributed devices need to communicate frequently to share knowledge from ongoing tasks, while the server must effectively derive new insights from the LLM. Issues related to communication efficiency and training resources can impede model convergence. Future research should explore innovative approaches for LLM-based NCCL to overcome these obstacles.
    \item {\textit{Generative AI:} Generative AI has emerged as a prominent research area in recent years, as it leverages the generalization capabilities of LLMs to significantly enhance task performance and drive productivity improvements \cite{yang2025recent}. However, even in centralized learning paradigms, continual learning for Generative AI remains hindered by multiple challenges. Unlike conventional backbone models that primarily face catastrophic forgetting, Generative AI systems more frequently encounter difficulties in acquiring new knowledge \cite{zhao2024sapt}. Additionally, the current deployment of Generative AI models through quantization compression on small edge devices complicates continual learning implementation under resource-constrained conditions. While the NCCL has not yet published studies specifically addressing Generative AI, the growing deployment of edge devices equipped with Generative AI capabilities and their potential to collect real-world data from novel scenarios underscores the urgent need for such research.}
    %
    \item \textit{Convergence Analysis:} Although NCCL research has conducted experiments on various datasets using different methods, empirically demonstrating the effectiveness of the approaches, theoretical research in this area is still lacking. One of the key considerations in NCCL is the convergence of the algorithm. For example, FL aims to find weights that minimize the global model aggregation, which essentially constitutes a distributed optimization problem where convergence is not always guaranteed. Theoretical analysis and evaluations of the convergence bounds of gradient descent-based NCCL for both convex and non-convex loss functions represent important research directions.
    \item \textit{Generation Task:} Existing NCCL research has primarily focused on classification tasks and achieved considerable success. However, generation tasks are gradually becoming mainstream, such as conversation recommendation \cite{jannach2021survey}, chat-bots \cite{dam2024complete}, AI image synthesis \cite{zhang2023text}, and more. Future research needs to expand generation tasks as downstream applications and explore the potential new training challenges they may pose. This will significantly enhance the impact and effectiveness of NCCL, facilitating its practical deployment.
    \item \textit{Concept Drift:} Concept drift often arises in practical applications, especially in NCCL environments where distributed devices can encounter it. Despite extensive research on traditional concept learning, like macro-financial analysis and epidemic modeling, addressing concept drift in NCCL and contemporary real-world scenarios offers a fascinating and valuable research direction.
    \item \textit{Asynchronous Learning:} Regarding the current setting of NCCL scenarios, most current NCCL methods focus on synchronous tasks, which assume that incremental tasks arrive at the same time. Asynchronous learning enables each device to collect new training data and initiate training at different timestamps, thereby efficiently utilizing device resources and minimizing unnecessary waiting time. Additionally, asynchronous learning may pose challenges such as global knowledge forgetting and cross-task knowledge integration. However, this more accurately reflects practical NCCL settings and can be a crucial factor in ensuring the scalability of NCCL, given that the NCCL training process can span a longer duration compared to traditional NCCL.
\end{itemize}

\section{Conclusion}\label{IX}
This paper has presented a tutorial on NCCL and a comprehensive survey of the issues related to NCCL implementation. Firstly, we begin with an introduction to the motivation for integrating CL algorithms into the NCL paradigm and how NCCL can serve as a potential technology for better collaborative model training in practical deployment. Then, we describe the fundamentals of NCL paradigms and different CL tasks. Afterward, we provide detailed reviews, analyses, and comparisons of approaches to alleviate catastrophic forgetting and distribution shifts in NCCL from three levels. Furthermore, we also discuss issues that include heterogeneity, security \& privacy, and real-world applications. Finally, we discuss challenges and future research directions.

\section*{Acknowledgments}
This work is supported by the National Key Research and Development Program of China under grant 2024YFC3307900; the National Natural Science Foundation of China under grants 62376103, 62302184, 62436003 and 62206102; Major Science and Technology Project of Hubei Province under grant 2024BAA008; Hubei Science and Technology Talent Service Project under grant 2024DJC078; and Ant Group through CCF-Ant Research Fund. The computation is completed in the HPC Platform of Huazhong University of Science and Technology.  

{\small
\bibliographystyle{IEEEtran}
\bibliography{reference}
}

\begin{IEEEbiography}[{\includegraphics[width=1in,height=1.25in,clip,keepaspectratio]{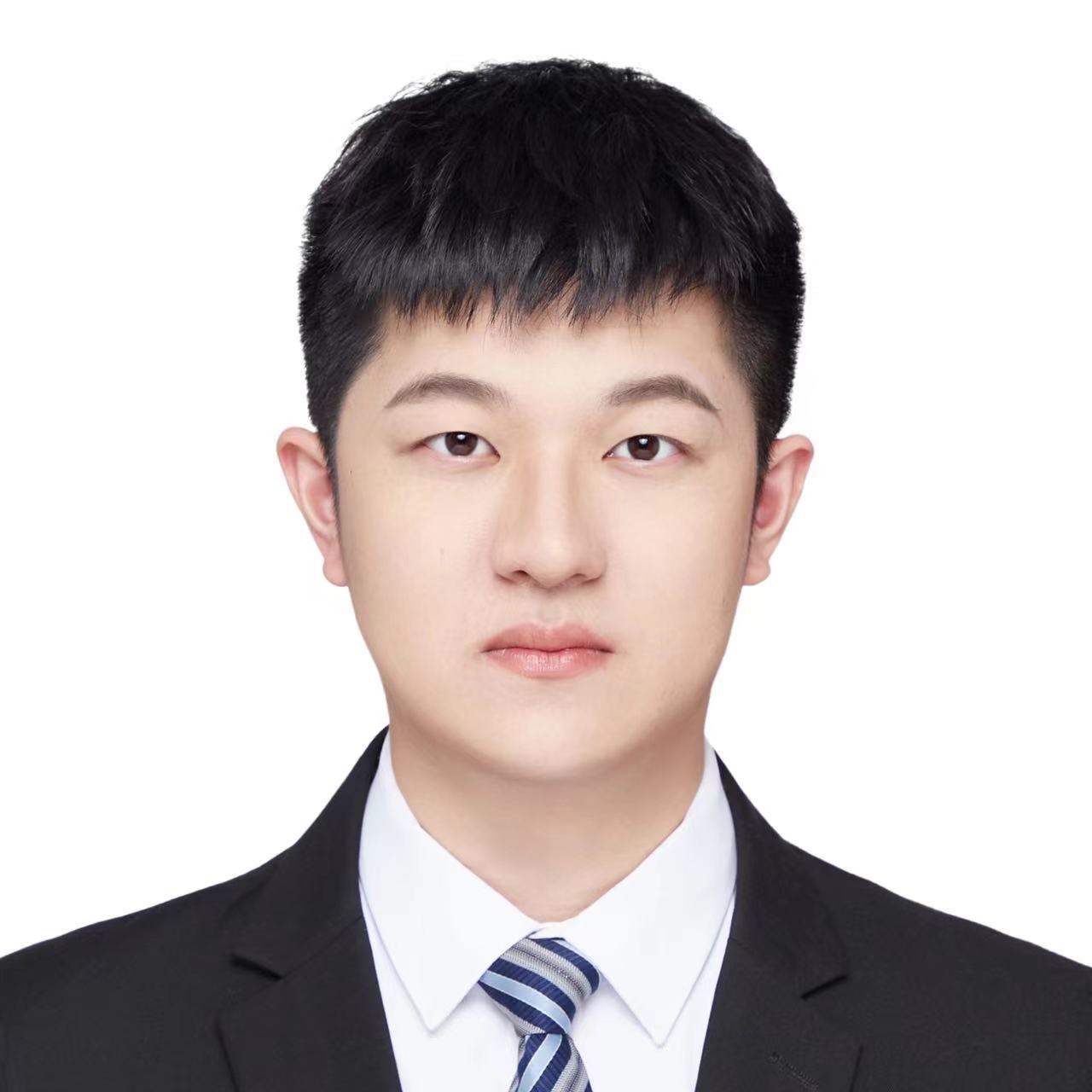}}]{Yichen Li} is currently a Ph.D. candidate in the School of Computer Science and Technology at Huazhong University of Science and Technology. Prior to this, he obtained his bachelor's degree from the School of Computer Science and Technology at Soochow University. His research interests include Federated Learning, Recommendation Systems, and Software Engineering. As the first author, he has published several papers at top-tier machine learning and computer vision conferences and journals such as CVPR, ECCV, ICML, ICLR, and IEEE TPAMI.
\end{IEEEbiography}

\begin{IEEEbiography}[{\includegraphics[width=1in,height=1.25in,clip,keepaspectratio]{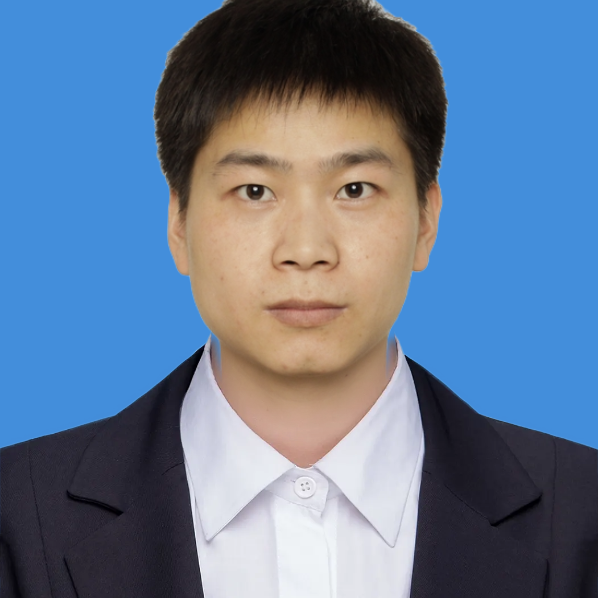}}]{Haozhao Wang} is currently an assistant professor in the School of Computer Science and Technology at Huazhong University of Science and Technology. Before that, he worked as a Postdoc and received a Ph.D. degree at the same department and school. He received the B.S. degree in computer science at the University of Electronic Science and Technology in 2016. He also worked as as a research fellow in SLab at Nanyang Technical University. His research interests include Distributed Machine Learning and Multi-modal Learning.
\end{IEEEbiography}

\begin{IEEEbiography}
[{\includegraphics[width=1in,height=1.25in,clip,keepaspectratio]{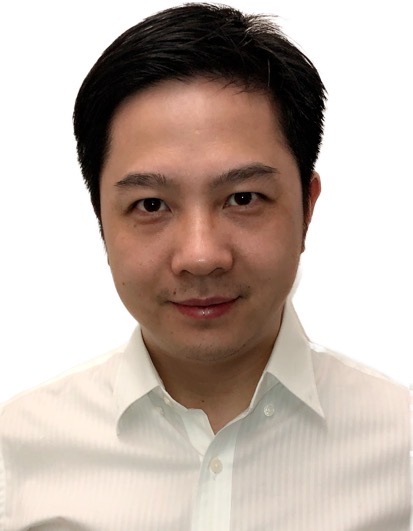}}]{Wenchao~Xu} is a research assistant professor at The Hong Kong Polytechnic University. He received his Ph.D. degree from University of Waterloo, Canada, in 2018. Before that he received the B.E. and M.E. degrees from Zhejiang University, Hangzhou, China, in 2008 and 2011, respectively. In 2011, he joined Alcatel Lucent Shanghai Bell Co. Ltd., where he was a Software Engineer for telecom virtualization. He has also been an Assistant Professor at School of Computing and Information Sciences in Caritas Institute of Higher Education, Hong Kong. His research interests includes wireless communication, Internet of things, distributed computing and AI enabled networking. 
\end{IEEEbiography}

\begin{IEEEbiography}
[{\includegraphics[width=1in,height=1.25in,clip,keepaspectratio]{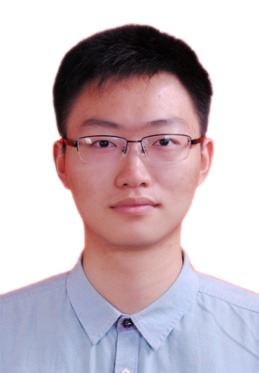}}]{Tianzhe Xiao} received the B.E. degree from Wuhan University of Science and Technology in 2021. He is currently working toward his PhD degree with the School of Computer Science and Technology, Huazhong University of Science and Technology. His research interests include federated learning and machine learning security.
\end{IEEEbiography}

\begin{IEEEbiography}
[{\includegraphics[width=1in,height=1.25in,clip,keepaspectratio]{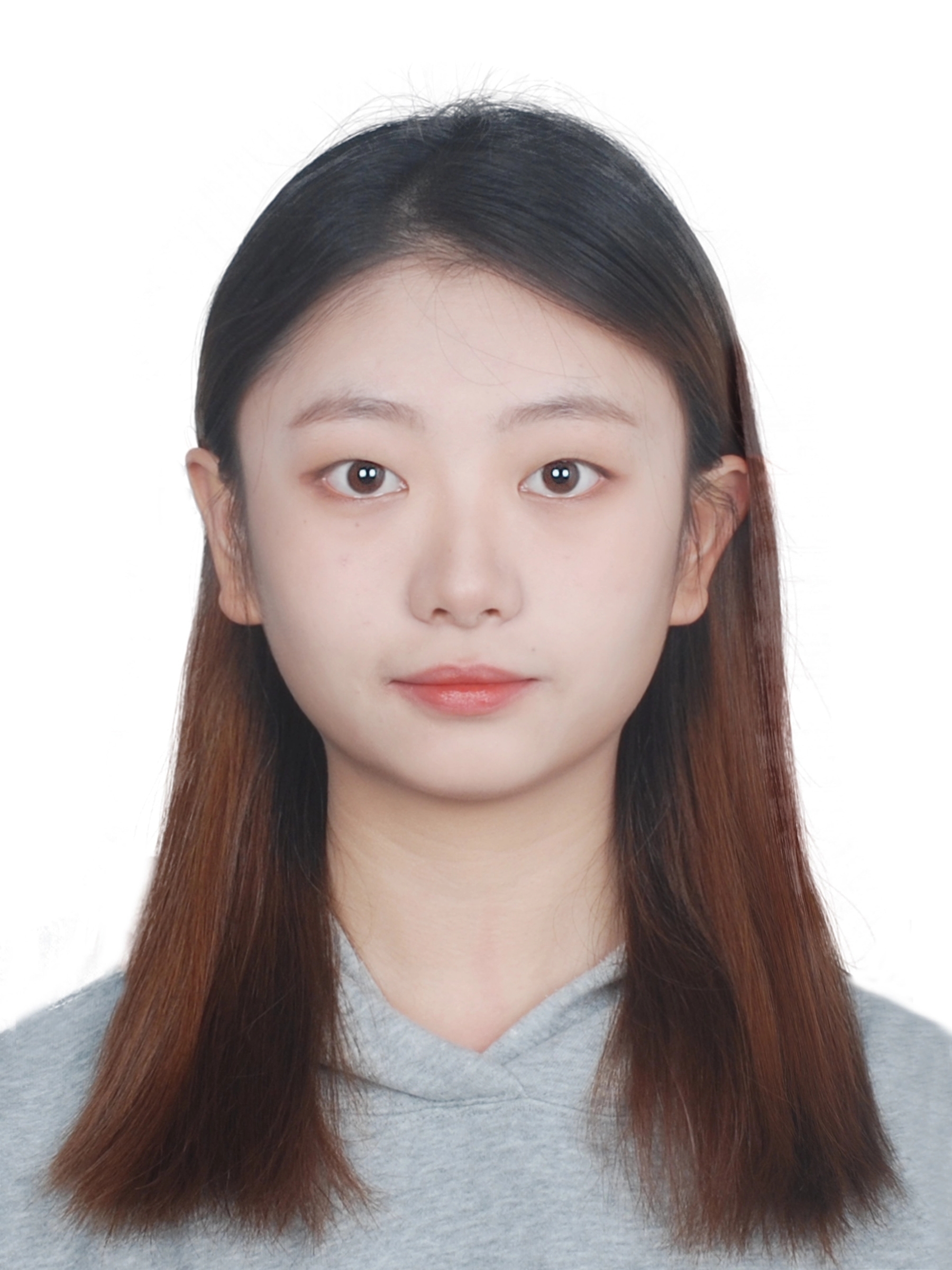}}]
{Hong Liu} received her B.E. degree from Huazhong University of Science and Technology in 2022. She is currently working toward her master's degree in School of Computer Science and Technology, Huazhong University of Science and Technology. Her research interests include federated learning and continual learning.
\end{IEEEbiography}

\begin{IEEEbiography}
[{\includegraphics[width=1in,height=1.25in,clip,keepaspectratio]{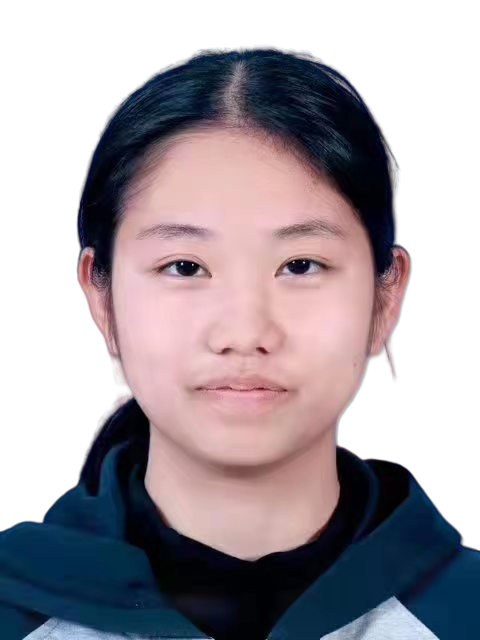}}]
{Minzhu Tu} is currently working toward her B.E. degree with the School of Computer Science and Technology, Beijing University of Posts and Telecommunications. Her research interests include Federated Learning and Software Engineering.
\end{IEEEbiography}

\begin{IEEEbiography}
[{\includegraphics[width=1in,height=1.25in,clip,keepaspectratio]{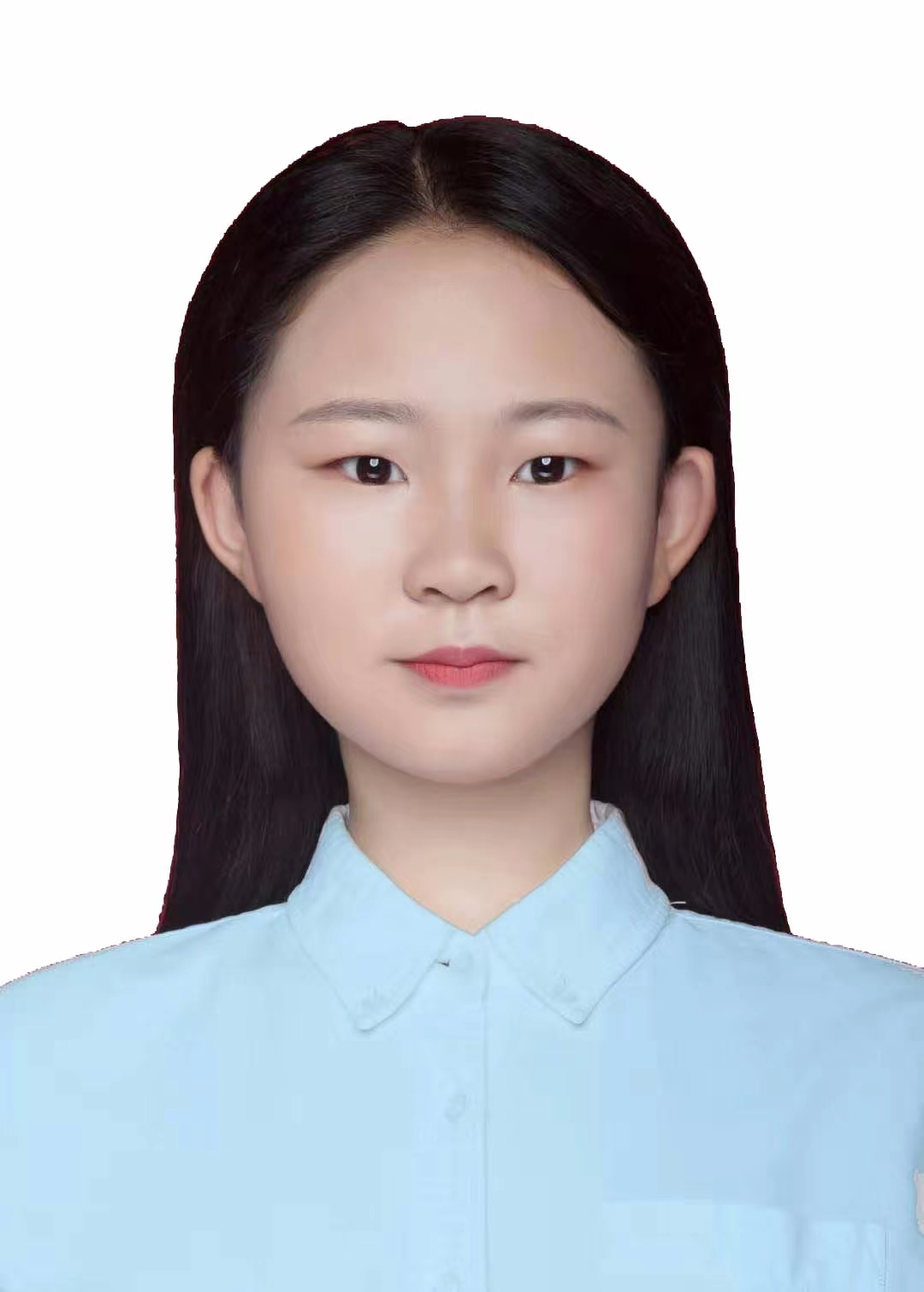}}]{Yuying Wang} received the B.E. degree from Shandong Normal University in 2023. She is currently working toward her master's degree with the School of Computer Science and Technology, Soochow University. Her research interests include Federated Learning and Software Engineering.
\end{IEEEbiography}

\begin{IEEEbiography}
[{\includegraphics[width=1in,height=1.25in,clip,keepaspectratio]{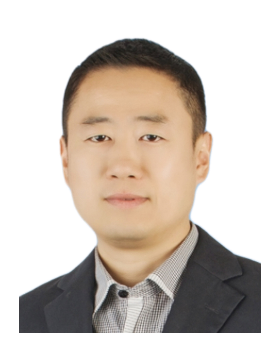}}]{Xin Yang} received an M.S. degree in electronic engineering from Sichuan University, China, in 2010, and the Ph.D. degree in computer science from Southwest Jiaotong University, Chengdu, in 2019. He is currently a Professor at the School of Computing and Artificial Intelligence, Southwestern University of Finance and Economics. He has authored more than 70 research papers in refereed journals and conferences. His research interests include machine learning, continual learning and multi-granularity learning.
\end{IEEEbiography}

\begin{IEEEbiography}[{\includegraphics[width=1in,height=1.25in,clip,keepaspectratio]{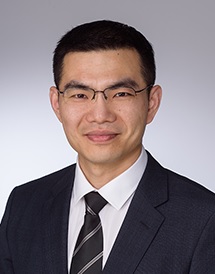}}]{Rui Zhang} is a professor in the School of Computer Science and Technology at Huazhong University of Science and Technology. He is a Visiting Professor at Tsinghua University and was a Professor at the School of Computing and Information Systems of the University of Melbourne. He has won several awards including the prestigious Future Fellowship by the Australian Research Council in 2012, Chris Wallace Award for Outstanding Research by the Computing Research and Education Association of Australasia (CORE) in 2015, and Google Faculty Research Award in 2017. His inventions have been adopted by major IT companies such as Microsoft, Amazon and AT\&T. He obtained his Bachelor's degree from Tsinghua University in 2001, PhD from National University of Singapore in 2006, and has then started as a faculty member in The University of Melbourne since 2007. Before joining the University of Melbourne, he has been a visiting research scientist at AT\&T labs-research in New Jersey and at Microsoft Research in Redmond, Washington. He has also been a regular visiting researcher at Microsoft Research Asia in Beijing. His research interests include big data and AI, particularly in areas of recommendation systems, knowledge bases, chatbot, spatial and temporal data analytics, moving object management and data streams.
\end{IEEEbiography}

\begin{IEEEbiography}[{\includegraphics[width=1in,height=1.25in,clip,keepaspectratio]{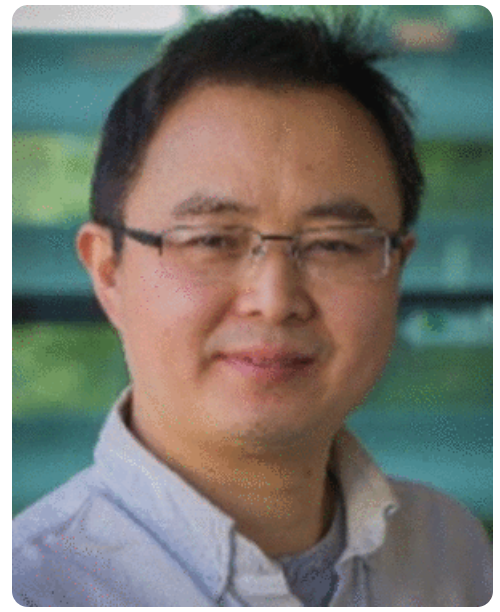}}]{Shui Yu} is Professor of the School of Computer Science in the Faculty of Engineering and Information Technology at UTS, the Deputy Chair of the UTS Research Committee, and is a researcher of cybersecurity, privacy and the networking, communication aspects of Big Data, and applied mathematics for computer science. Many of his research outputs have been adopted by industry, for example, the auto scale strategy of Amazon Cloud against distributed denial-of-service attacks. Shui serves his research communities in various roles, including serving on the editorial boards of IEEE Communications Surveys and Tutorials, IEEE Communications Magazine and the IEEE Internet of Things Journal, among others. He has been a member of organizing committees for many international conferences, such as the publication chair for IEEE Globecom 2015, IEEE INFOCOM 2016 and 2017, TPC chair for IEEE Big Data Service 2015, and general chair for ACSW 2017. He served as a Distinguished Lecturer of IEEE Communications Society (2018-2021). He is a Distinguished Visitor of IEEE Computer Society (2022-2024), a voting member of IEEE ComSoc Educational Services board, and an elected member of Board of Governors of IEEE Communications Society and IEEE Vehicular Technology Society, respectively. He is a Fellow of IEEE.

\end{IEEEbiography}

\begin{IEEEbiography}[{\includegraphics[width=1in,height=1.25in,clip,keepaspectratio]{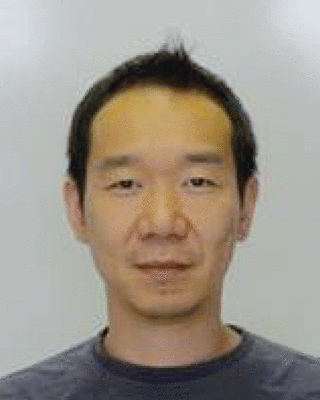}}]{Song Guo} is a chair professor in the Department of Computer Science and Engineering at the Hong Kong University of Science and Technology. Before joining HKUST in 2023, he was a Professor at The Hong Kong Polytechnic University. He also holds a Changjiang Chair Professorship awarded by the Ministry of Education of China. Prof. Guo is a Fellow of the Canadian Academy of Engineering and a Fellow of the IEEE (Computer Society). His research interests are mainly in big data, edge AI, mobile computing, and distributed systems. He published many papers in top venues with wide impact in these areas and was recognized as a Highly Cited Researcher (Clarivate Web of Science). He is the recipient of over a dozen Best Paper Awards from IEEE/ACM conferences, journals, and technical committees. Prof. Guo is the Editor-in-Chief of IEEE Open Journal of the Computer Society and the Chair of IEEE Communications Society (ComSoc) Space and Satellite Communications Technical Committee. He was an IEEE ComSoc Distinguished Lecturer and a member of IEEE ComSoc Board of Governors. He has served the IEEE Computer Society on the Fellow Evaluation Committee and has been named on the editorial board of a number of prestigious international journals like IEEE TPDS, IEEE TCC, IEEE TETC, etc. He has also served as chair of organizing and technical committees of many international conferences.
\end{IEEEbiography}

\begin{IEEEbiography}[{\includegraphics[width=1in,height=1.25in,clip,keepaspectratio]{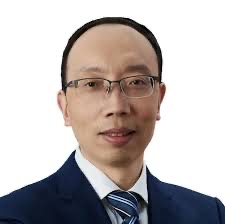}}]{Ruixuan Li} is a professor in the School of Computer Science and Technology at Huazhong University of Science and Technology. He received the B.S., M.S. and Ph.D. in computer science from Huazhong University of Science and Technology, China in 1997, 2000 and 2004 respectively.He was the visiting scholar of the University of Western Sydney (UWS), Australia, in 2005. He was the visiting researcher of the University of Toronto (UofT), Canada, from June 2009 to October 2010. He also holds an Adjunct Professor position at Concordia University, an Adjunct Senior Fellow position at University of Western Sydney (UWS), and a Guest Researcher position at Institute of Information Engineering, Chinese Academy of Sciences. His research interests include cloud and edge computing, big data management, and distributed system security. He is a member of IEEE and ACM; is a distinguished member of China Computer Federation (CCF); is the Vice Director of Technical Committee of Distributed Computing and System of CCF, and Technical Committee of Information System of CCF.
\end{IEEEbiography}

\end{document}